\documentclass[10pt]{article} %
\usepackage[preprint]{tmlr}

\definecolor{citecolor}{HTML}{0071bc}
\usepackage[breaklinks=true,bookmarks=false,colorlinks,bookmarks=false, citecolor=citecolor]{hyperref} 
\usepackage{url}

\PassOptionsToPackage{numbers}{natbib}
\usepackage{array, multirow, tablefootnote}
\PassOptionsToPackage{dvipsnames,table}{xcolor}
\usepackage[bottom]{footmisc}
\usepackage{adjustbox}
\usepackage{amsmath, amsfonts, bm}
\usepackage{caption}
\usepackage{subcaption}
\usepackage{pifont}
\usepackage{booktabs}
\usepackage{xspace}
\usepackage{wrapfig}
\usepackage{placeins}
\usepackage[normalem]{ulem}

\let\oldcite\cite
\renewcommand{\cite}[1]{\mbox{\oldcite{#1}}}

\def \ours {{DA\textsubscript{IC-GAN}}\xspace}
\def \ccicgan {{CC-IC-GAN}\xspace}
\def \icgan {{IC-GAN}\xspace}
\def \allicgan {{(CC-)IC-GAN}\xspace}

\newcommand{\cmark}{\ding{51}}%
\newcommand{\cmarkg}{\textcolor{gray}{\ding{51}}}%
\newcommand{\xmark}{\ding{55}}%
\newcommand{\xmarkg}{\textcolor{gray}{\ding{55}}}%

\newcommand*{\x}{\mathbf{x}}
\newcommand*{\y}{\mathbf{y}}
\newcommand*{\z}{\mathbf{z}}
\newcommand*{\h}{\mathbf{h}}

\newcommand*{\D}{\mathcal{D}}
\newcommand*{\R}{\mathbb{R}}
\newcommand*{\Rimg}{\R^{3 \times H \times W}}

\title{Instance-Conditioned GAN Data Augmentation\\for Representation Learning}

\author{\name Pietro Astolfi\footnotemark[1] \footnotemark[2] \textsuperscript{1},
Arantxa Casanova\footnotemark[1] \textsuperscript{1,2,5},
Jakob Verbeek\textsuperscript{1},
Pascal Vincent\textsuperscript{1,2,3,4},\\
Adriana Romero-Soriano\textsuperscript{1,2,6},
Michal Drozdzal\textsuperscript{1}\\
\addr 
\textsuperscript{1}Meta AI, 
\textsuperscript{2}Mila, Quebec AI Institute,
\textsuperscript{3}Université de Montréal,
\textsuperscript{4}CIFAR,\\
\textsuperscript{5}École Polytechnique de Montréal,
\textsuperscript{6}McGill University
\\
\email \footnotemark[2] pietroastolfi@meta.com \\
\footnotemark[1] Contributed equally
      }

\def\month{MM}  %
\def\year{YYYY} %
\def\openreview{\url{https://openreview.net/forum?id=XXXX}} %

\begin{document}

\maketitle

\begin{abstract}

Data augmentation has become a crucial component to train state-of-the-art visual representation models. However, handcrafting combinations of transformations that lead to improved performances is a laborious task, which can result in visually unrealistic samples. To overcome these limitations, recent works have explored the use of generative models as learnable data augmentation tools, showing promising results in narrow application domains, e.g., few-shot learning and low-data medical imaging. In this paper, we introduce a data augmentation module, called \ours, which leverages instance-conditioned GAN generations and can be used off-the-shelf in conjunction with most state-of-the-art training recipes. We showcase the benefits of \ours by plugging it out-of-the-box into the supervised training of ResNets and DeiT models on the ImageNet dataset, and achieving accuracy boosts up to between 1\%p and 2\%p with the highest capacity models. Moreover, the learnt representations are shown to be more robust than the baselines when transferred to a handful of out-of-distribution datasets, and exhibit increased invariance to variations of instance and viewpoints. We additionally couple \ours with a self-supervised training recipe and show that we can also achieve an improvement of 1\%p in accuracy in some settings. %
With this work, we strengthen the evidence on the potential of learnable data augmentations to improve visual representation learning, paving the road towards non-handcrafted augmentations in model training.

\end{abstract}

\section{Introduction}
\label{sec:intro}

\begin{figure}
    \centering

    \begin{subfigure}[b]{0.16\textwidth}
         \centering
         {Original sample}
     \end{subfigure}
    \begin{subfigure}[b]{0.16\textwidth}
         \centering
         {RandomCrop}
     \end{subfigure}
    \begin{subfigure}[b]{0.16\textwidth}
         \centering
         {RandAugment}
    \end{subfigure}
    \begin{subfigure}[b]{0.16\textwidth}
         \centering
         {MixUp}
     \end{subfigure}
    \begin{subfigure}[b]{0.16\textwidth}
         \centering
         {CutMix}
     \end{subfigure}
    \begin{subfigure}[b]{0.16\textwidth}
         \centering
         {\icgan}
    \end{subfigure}
    
    \begin{subfigure}[b]{0.16\textwidth}
       \centering
        \includegraphics[width=\textwidth]{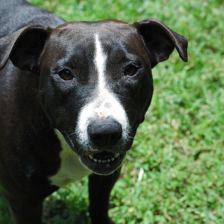}
    \end{subfigure}
    \begin{subfigure}[b]{0.16\textwidth}
       \centering
        \includegraphics[width=\textwidth]{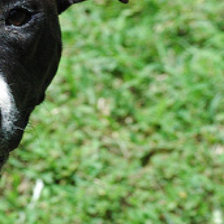}
    \end{subfigure}
    \begin{subfigure}[b]{0.16\textwidth}
       \centering
        \includegraphics[width=\textwidth]{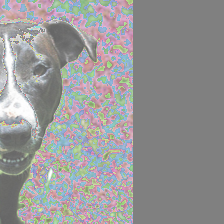}
    \end{subfigure}
    \begin{subfigure}[b]{0.16\textwidth}
       \centering
        \includegraphics[width=\textwidth]{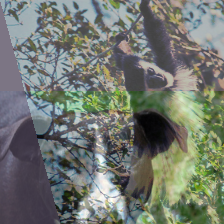}
    \end{subfigure}
    \begin{subfigure}[b]{0.16\textwidth}
       \centering
        \includegraphics[width=\textwidth]{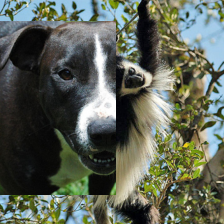}
    \end{subfigure}
    \begin{subfigure}[b]{0.16\textwidth}
       \centering
        \includegraphics[width=\textwidth]{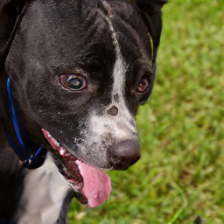}
    \end{subfigure}
    
    \caption{Visual comparison of various hand-crafted data augmentations and one \icgan generation, using a sample from ImageNet as input.}
    \label{fig:da_comparison}
\end{figure}

Recently, deep learning models have been shown to achieve astonishing results across a plethora of computer vision tasks when trained on \emph{very large} datasets of hundreds of millions datapoints \citep{alayrac_flamingo_2022, gafni22arxiv, goyal_vision_2022, radford_learning_2021, ramesh_hierarchical_2022, saharia_photorealistic_2022, zhang_opt_2022}. 
Oftentimes, however,  large datasets are not available, limiting the performance of deep learning models. To overcome this limitation, researchers explored ways of artificially increasing the size of the training data by transforming the input images via \emph{handcrafted} data augmentations.
These augmentation techniques consist of heuristics involving different types of image distortion~\citep{shorten_survey_2019}, including random erasing \citep{devries_improved_2017, zhong_random_2020}, and image mixing \citep{yun_cutmix_2019, zhang_mixup_2017}. 
It is important to note that all current state-of-the-art representation learning models seem to benefit from such complex data augmentation recipes as they help regularizing models -- e.g., vision transformers trained with supervision~\citep{dosovitskiy_image_2021, touvron_training_2021, steiner_how_2021, touvron_deit_2022} and models trained with self-supervision~\citep{chen_simple_2020, he_momentum_2020, caron_unsupervised_2020, caron_emerging_2021, grill_bootstrap_2020-1}. However, coming up with data augmentation recipes is laborious and the augmented images, despite being helpful, often look unrealistic, see first five images in Figure~\ref{fig:da_comparison}. Such a lack of realism is a sub-optimal effect of these heuristic data augmentation strategies, which turns out to be even detrimental when larger training datasets are available~\citep{steiner_how_2021} and no strong regularization is needed.\looseness-1

To this end, researchers have tried to move away from training exclusively on real dataset samples and their corresponding hand-crafted augmentations, and have instead explored increasing the dataset sizes with generative model samples~\citep{frid-adar_gan-based_2018, bowles_gan_2018,ravuri_classification_2019,zhang_datasetgan_2021-2,li_semantic_2021}. 
A generative model can  potentially provide infinitely many synthetic image samples; however, the quality and diversity of the generated samples is usually a limiting factor, resulting in moderate gains for specific tasks like image segmentation~\citep{zhang_datasetgan_2021-2,li_semantic_2021} and in poor performance in standard image classification benchmarks~\citep{ravuri_classification_2019}. With the advent of photorealistic image samples obtained with generative adversarial networks (GAN)~\citep{goodfellow_generative_2014}, researchers have explored the use of GAN-based data augmentation~\citep{antoniou_data_2018, tritrong_repurposing_2021, mao_generative_2021, wang_regularizing_2021}. However, none of these approaches has shown improvement when applied to large-scale datasets, such as ImageNet~\citep{deng09cvpr}, in most cases due to a lossy and computationally intensive GAN-inversion step.

In this paper, we study the use of Instance-Conditioned GAN (IC-GAN)~\citep{casanova_instance-conditioned_2021}, a generative model that, conditioned on an image, generates samples that are semantically similar to the conditioning image.
Thus, we propose to leverage \icgan to generate plausible augmentations of each available datapoint and design a module, called \ours, that can be coupled off-the-shelf with most supervised and self-supervised data augmentation strategies and training procedures.
We validate the proposed approach by training supervised image classification models of increasing capacity on the ImageNet dataset and evaluating them in distribution and out-of-distribution. Our results highlight the benefits of leveraging \ours, by outperforming strong baselines when considering high-capacity models, and by achieving robust representations exhibiting increased invariance to viewpoint and instance. We further couple \ours with a self-supervised learning model and show that we can also boost its performance in some settings.

Overall, the contributions of this work can be summarized as follows:
\begin{itemize}

    \item We introduce \ours, a data augmentation module that combines IC-GAN with handcrafted data augmentation techniques and that can be plugged off-the-shelf into most supervised and self-supervised training procedures.\looseness-1
    
    \item We find that using \ours in the supervised training scenario is beneficial for high-capacity networks, e.g., ResNet-152, ResNet-50W2, and DeIT-B, boosting in-distribution performance and robustness to out-of-distribution when combined with traditional data augmentations like random crops and RandAugment. %

    \item We extensively explore \ours’s impact on the learned representations, we discover an interesting correlation between per-class FID and classification accuracy, and report promising results in the self-supervised training of SwAV when not used in combination with multi-crop.\looseness-1

\end{itemize}

\section{Related Work}
\label{sec:relw}

\paragraph{Image distortion.} %
Over the past decades, the research community has explored a plethora of simple hand-designed image distortions such as zoom, reflection, rotation, shear, color jittering, solarization, and blurring --- see \cite{shorten_survey_2019} and \cite{perez_effectiveness_2017} for an extensive survey. %
Although all these distortions induce the model to be robust to small perturbations of the input, they might lead to unrealistic images and provide only limited image augmentations. To design more powerful image distortions, the research community has started to combine multiple simple image distortions into a more powerful data augmentation schemes such as Neural Augmentation~\citep{perez_effectiveness_2017}, SmartAugment~\citep{lemley_smart_2017}, AutoAugment~\citep{cubuk_autoaugment_2019}, and RandAugment~\citep{cubuk_randaugment_2020-2}. Although these augmentation schemes oftentimes significantly improve model performance, the resulting distortions are limited by the initial set of simple distortions. Moreover, finding a good combination of simple image distortions is computationally intense since it requires numerous network trainings. 
\paragraph{Image mixing.} An alternative way to increase the diversity of augmented images is to consider multiple images and their labels. For example, CutMix~\citep{yun_cutmix_2019} creates collages of pairs of images while MixUp~\citep{zhang_mixup_2017} interpolates them pixel-wise. In both cases, the mixing factor is regulated by a hyper-parameter, which is also used for label interpolation. However, these augmentation techniques directly target the improvement of class boundaries, at the cost of producing
unrealistic  %
images. 
We argue %
that unrealistic augmentations are a sub-optimal heuristic currently adopted as a strong regularizer, which is no longer needed when larger datasets are available, as shown in~\cite{steiner_how_2021}.%

\paragraph{Data augmentation with autoencoders.} To improve the realism of augmented images, some researchers have explored applying the image augmentations in the latent space of an autoencoder (AE). \cite{devries_dataset_2017} and 
\cite{liu_data_2018} proposed to interpolate/extrapolate neighborhoods in latent space to generate new images. %
Alternatively, \cite{schwartz_delta-encoder_2018} introduced a novel way of training AE to synthesize images from a handful of samples and use them as augmentations to enhance few-shot learning.
Finally, \cite{pesteie_adaptive_2019} used a variational AE trained to synthesize clinical images for data augmentation purposes. However, most of above-mentioned approaches are limited by the quality of the reconstructed images which are oftentimes blurry.

\paragraph{Data augmentation with generative models.} To improve the visual quality of augmented images, the community has studied the use of generative models in the context of both data augmentation and dataset augmentation. \cite{tritrong_repurposing_2021, mao_generative_2021} explored the use of instance-specific augmentations obtained via GAN inversion~\citep{xia_gan_2022, huh_transforming_2020, zhu_generative_2016}, which map original images into latent vectors that can be subsequently transformed to generate augmented images~\citep{jahanian_steerability_2020, harkonen_ganspace_2020}. %
However, GAN inversion is a computationally intense operation and latent space transformations are difficult to control~\citep{wang_implicit_2019, wang_regularizing_2021}. \cite{antoniou_data_2018} proposed a specific GAN model to generate a realistic image starting from an original image combined with a noise vector. However, this work was only validated on low-shot benchmarks. Researchers have also explored learning representations using samples from a trained generative model exclusively~\citep{shrivastava_learning_2017, zhang_datasetgan_2021-2, li_bigdatasetgan_2022, besnier_this_2020, li_semantic_2021, zhao_synthesizing_2022, jahanian_generative_2022} as well as combining real dataset images with samples from a pre-trained generative model~\citep{frid-adar_gan-based_2018, bowles_gan_2018, ravuri_classification_2019}, with the drawback of drastically shifting the training distribution.
Finally, the use of unpaired image-to-image translation methods to augment small datasets was explored in \cite{sandfort_data_2019, huang_auggan_2018, gao_low-shot_2018, choi_self-ensembling_2019}.
However, such approaches are designed to translate source images into target images and thus are limited by the source and target image distributions.

\paragraph{Data augmentation with latent neighbor images.} Another promising data augmentation technique uses neighbor images to create semantically-similar image pairs that can be exploited for multi-view representation learning typical of SSL. This technique was promoted in NNCLR~\citep{dwibedi_little_2021-1}, an extension of the SSL model SimCLR~\citep{chen_simple_2020} to use neighbors, with some limitations due to the restricted and dynamic subset used for neighbor retrieval. Alternatively, \cite{jahanian_generative_2022} explored the generation of neighbor pairs by using latent space transformations in conjunction with a pre-trained generative model. However, this model only uses generated samples to learn the representations and reports poor performance on a simplified ImageNet setup (training on $128\times128$ resolution images for only 20 epochs).

\section{Methodology}
\label{sec:method}

\subsection{Review of Instance-Conditioned GAN (IC-GAN)}
\label{ssec:met:bg}

Instance-conditioned GAN (IC-GAN)~\citep{casanova_instance-conditioned_2021} is a  conditional generative model that synthesizes high quality and diverse images that resemble an input image used to condition the model. The key idea of \icgan is to model the data distribution as a mixture of overlapping and fine-grained data clusters, defined by each datapoint -- or ``instance'' -- and the set of its nearest neighbors. Training \icgan requires access to a dataset $\mathcal{D} = \{\x_i\}_{i=1}^N$ with $N$ datapoints and a pre-trained feature extractor $E_\phi$ parameterized by $\phi$. The pre-trained feature extractor is used to extract embedded representations $\mathbf{h}_i = E_\phi(\x_i)$. Next, a set of nearest neighbors $\mathcal{A}_i$, with cardinality $k$, is computed using the cosine similarity in the embedded representation space. The \icgan generator network, $G_\psi$, parameterized by $\psi$, takes as input an embedded representation $\mathbf{h}$ together with a Gaussian noise vector $\z \sim \mathcal{N}(0,I)$, and generates a synthetic image $\tilde\x = G_\psi(\z, \mathbf{h})$. IC-GAN is trained using a standard adversarial game between a generator $G_\psi$ and a discriminator $D_\omega$, parameterized by $\omega$, as follows:
\begin{equation}
\min_{\psi}\max_{\omega} \;  
\mathbb{E}_{\x_i\sim p(\x), \x_j\sim \mathcal{A}_i} \left[\ln D_\omega(\x_j, \mathbf{h}_i)\right] \; +
\; \mathbb{E}_{ \x_i\sim p(\x),\z \sim \mathcal{N}(0,I) } \left[\ln(1-D_\omega(G_\psi(\z, \mathbf{h}_i), \mathbf{h}_i))\right].
\end{equation}
The discriminator $D_\omega$ attempts to distinguish between real samples in $\mathcal{A}_i$ and the generated samples, while the generator $G_\psi$ tries to fool the discriminator by generating realistic images following the distribution of the nearest neighbor samples in $\mathcal{A}_i$. In the class-conditional version of \icgan, referred to as \ccicgan, a class label $y$ is used as an extra input conditioning for the generator, such that $\tilde\x = G_\psi(\z, \mathbf{h}, y)$; this enables control over the generations given both a class label and an input image.

\subsection{Data augmentation with \icgan}
\label{ssec:met:da_icg}

\begin{figure}
    \centering
    \includegraphics[width=1\textwidth]{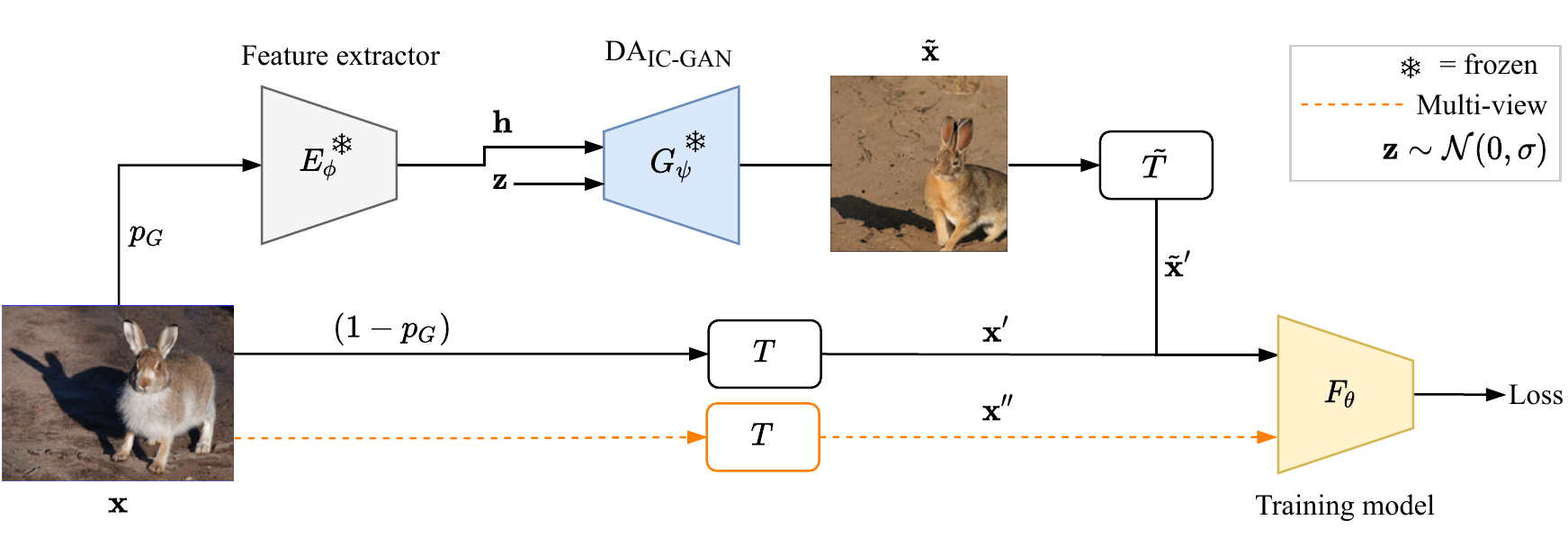}
    \caption{\ours integration scheme to train a model $F_\theta$. For each image $\x \in \D$ we apply \ours, with probability $p_G$. When an image is \icgan-augmented, the  representation $\h$ of the image is used as input to the generator, together with the Gaussian noise $\z$. 
    The generated image $\tilde\x$ undergoes an additional sequence of handcrafted data augmentations, $\tilde T$. When \ours is not applied, the standard handcrafted data augmentation, $T$, is applied to $\x$ to produce $\x'$. In the case of multi-view SSL training (orange branch), an additional view, $\x''$, is obtained by independently applying $T$ to the original image.
    }
    \label{fig:method}
\end{figure}

\paragraph{Data augmentation notation.} We define a data augmentation recipe as a transformation, $T$, of a datapoint, $\x_i \in \D$, to produce a perturbed version  $\x_i' = T(\x_i)$ of it. 
The data augmentation mapping $T : \Rimg \rightarrow \Rimg$ is usually composed of multiple single transformation functions defined in the same domain, $\tau : \Rimg \rightarrow \Rimg$, $T = \tau_1 \circ \tau_2 \circ ... \circ \tau_t$. Each function $\tau$ corresponds to a specific augmentation of the input such as zooming or color jittering, and is applied with a probability  $p_\tau$. Moreover, $\tau$ can be modified with other hyper-parameters $\lambda_\tau$ that are augmentation-specific -- e.g. magnitude of  zooming, or intensity of color distortion.

\paragraph{\ours.} We introduce a new data augmentation, \ours, that leverages a pre-trained \icgan generator model and can be used in conjunction with other data augmentation techniques to train a neural network. \ours is applied before any other data augmentation technique and is regulated by a hyper-parameter $p_G$ controlling a percentage of datapoints to be augmented. When a datapoint $\x_i$ is \icgan-augmented, it is substituted by the model sample $\tilde\x_i = G_\psi(\z, E_\phi(\x_i))$, with $\z$ a Gaussian noise vector. $\tilde\x_i$ may then be further transformed with a sequence of subsequent transformations $\tilde T = \tau_1 \circ \tau_2 \circ ... \circ \tau_{\tilde t}$. 
Note that $\tilde T$ might differ from the sequence of transformations $T$ applied when $\x_i$ is not \icgan-augmented. We depict this scenario in Figure \ref{fig:method}. Moreover, we use the {\em truncation trick} \citep{marchesi_megapixel_2017} and introduce a second hyper-parameter, $\sigma$, to control the variance of the latent variable $\z$. During \icgan training truncation is not applied, and  $\z$ is sampled from the unit Gaussian distribution. \ours augmentations can be applied to both supervised and self-supervised representation learning off-the-shelf, see section \ref{sec:exps} for details.\looseness-1 %

\section{Experimental Setup}
\label{sec:exps}

In our empirical analysis, we investigate the effectiveness of \ours in supervised and self-supervised representation learning. In the following subsections, we describe the experimental details of both scenarios.  

\subsection{Models, metrics, and datasets}

\paragraph{Models.} For supervised learning, we train ResNets~\citep{he_deep_2016} with different depths: 50, 101 and 152 layers, and widths: ResNet-50 twice wider (ResNet-50W2)~\citep{zagoruyko_wide_2016-1}, and DeiT-B~\citep{touvron_training_2021}. For self-supervised learning, we train the SwAV~\citep{caron_unsupervised_2020} model with a ResNet-50 backbone. For \ours, we employ two pre-trained generative models on ImageNet: \icgan and \ccicgan, both using the BigGAN~\citep{brock_large_2019} backbone architecture. \icgan conditions the generation process on instance feature representations, obtained with a pre-trained SwAV model\footnote{\url{https://dl.fbaipublicfiles.com/deepcluster/swav_800ep_pretrain.pth.tar}}, while \ccicgan conditions the generation process on both the instance representation obtained with a ResNet-50 trained for classification\footnote{\url{https://download.pytorch.org/models/resnet50-19c8e357.pth}} and a class label. %
Unless specified otherwise, our models use the default \icgan and \ccicgan configuration from \cite{casanova_instance-conditioned_2021}: neighborhood size of $k$=50 and $256\times256$ image resolution, trained using only horizontal flips as data augmentation\footnote{\url{https://github.com/facebookresearch/ic_gan}}. To guarantee a better quality of generations we set truncation $\sigma = 0.8$ and 1.0 for \icgan and \ccicgan respectively.
For simplicity, we will use the term \allicgan to refer to both pre-trained models hereinafter.

\paragraph{Datasets.} We train all the considered models from scratch on ImageNet \underline{(IN)}~\citep{deng09cvpr} and test them on the IN validation set. Additionally, in the supervised learning case, models are tested for robustness on a plethora of datasets, including \underline{Fake-IN}: containing 50K generated images obtained by conditioning the \icgan model on the IN validation set; \underline{Fake-IN\textsubscript{CC}}: containing 50K images generated with the \ccicgan conditioned on the IN validation set\footnote{To avoid as much as possible unrealistic generations in creating Fake-IN and Fake-IN\textsubscript{CC}, for each IN image we generate a set of 20 samples, from which we chose the one most similar (cosine similarity) to the conditioning image in the feature space.}; IN-Adversarial \underline{(IN-A)}~\citep{hendrycks_natural_2021}: composed of ResNet's adversarial examples present in IN\footnote{Although IN-A contains samples from only 200 out of the 1000 classes of IN, we compute the results without restricting the predictions to those 200 classes.}; IN-Rendition \underline{(IN-R)}~\citep{hendrycks_many_2021}: containing stylized images such as cartoons and paintings belonging to IN classes; \underline{IN-ReaL}~\citep{beyer_are_2020}: a relabeled version of the IN validation with multiple labels per image; and \underline{ObjectNet}~\citep{barbu_objectnet_2019}: containing object-centric images specifically selected to increase variance in viewpoint, background and rotation w.r.t. IN\footnote{The class mapping from ObjectNet to IN is one-to-multi -- i.e., one class is mapped to one or more classes of IN. We consider predictions pointing to any of the mapped classes as correct.}. We also consider the following datasets to study invariances in the learned representations: \underline{IN validation set} to analyze {\em instance+viewpoint} invariances; Pascal-3D+ \underline{(P3D)}~\citep{xiang_beyond_2014}, composed of $\sim$36K images from 12 categories to measure {\em instance}, and {\em instance+viewpoint} invariances; \underline{GOT}~\citep{huang_got-10k_2019}, 10K video clips with a single moving object to measure invariance to object {\em occlusion}; and \underline{ALOI}~\citep{geusebroek_amsterdam_2005}, single-object images from 1K object categories with plain dark background to measure invariance w.r.t.\ {\em viewpoint} (72 viewpoints per object), {\em illumination color} (12 color variations per object), and {\em illumination direction} (24 directions per object).

\paragraph{Metrics.} We quantify performance for classification tasks as the top-1 accuracy on a given dataset. Moreover, we analyze invariances of the learned representations by using the top-25 Representation Invariance Score (RIS) proposed by \cite{purushwalkam_demystifying_2020-1}. In particular, given a class $y$, we sample a set of object images $\mathcal{T}$ by applying a transformation $\tau$ with different parameters $\lambda_\tau$ such that $\mathcal{T} = \{\tau(x, \lambda_\tau) | \forall \lambda_\tau\}$. We then compute the mean invariance on the transformation $\tau$ of all the objects belonging to $y$ as the average firing rate of the (top-25) most frequently activating neurons/features in the learned representations. We follow the recipe suggested in \cite{purushwalkam_demystifying_2020-1} and compute the top-25 RIS only for ResNets models, extracting the learned representations from the last ResNet block ($2048$-$d$ vectors).

\paragraph{Per-class metrics.} We further stratify the results by providing class-wise accuracies and correlating them with the quality of the generated images obtained with \allicgan. We quantify the quality and diversity of generations using the Fréchet Inception Distance (FID)~\citep{heusel17nips}. We compute per-class FID by using the training samples of each class both as the reference and as the conditioning to generate the same number of synthetic images. We also measure a particular characteristic of the \icgan model: the \textit{NN corruption}, which measures the percentage of images in each datapoint's neighborhood that has a different class than the datapoint itself; this metric is averaged for all datapoints in a given class to obtain per-class NN corruption.

\subsection{Training recipes}
In this subsection, we define the training recipe for each model in both supervised and self-supervised learning; we describe which data augmentation techniques are used, how \ours is integrated and the hyper-parameters used to train the models.

\paragraph{Model selection.} In all settings, hyper-parameter search was performed with a restricted grid-search for the learning rate, weight decay, number of epochs, and \ours probability $p_G$, selecting the model with the best accuracy on IN validation.

\subsubsection{Supervised learning} 

For the ResNet models, we follow the standard procedure in Torchvision\footnote{See the \texttt{IMAGENET1K\_V1} recipe at \url{https://github.com/pytorch/vision/tree/main/references/classification}.} and apply random horizontal flips \underline{(Hflip)} with 50\% probability as well as random resized crops \underline{(RRCrop)}~\citep{krizhevsky_imagenet_2012}. We train ResNet models for 105 epochs, following the setup in VISSL~\citep{goyal2021vissl}. For DeiT-B we follow the experimental setup from~\cite{touvron_training_2021}, whose data augmentation recipe is composed of Hflip, RRCrop, RandAugment~\citep{cubuk_randaugment_2020-2}, as well as color jittering (CJ) and random erasing (RE)~\citep{zhong_random_2020} ---we refer to RandAugment + CJ + RE as \underline{RAug}---, and typical combinations of CutMix~\citep{yun_cutmix_2019} and MixUp~\citep{zhang_mixup_2017}, namely \underline{CutMixUp}. 
DeiT models are trained for the standard 300 epochs~\citep{touvron_training_2021} except when using only Hflip or RRCrop as data augmentation, where we reduce the training time to 100 epochs to mitigate overfitting. Both ResNets and DeiT-B are trained with default hyper-parameters; despite performing a small grid search, better hyper-parameters were not found. Additional details can be found in Appendix~\ref{suppl:details}.

\paragraph{Soft labels.} We note that the classes in the neighborhoods used to train the
\allicgan models are not homogeneous: a neighborhood, computed via cosine similarity between embedded images in a feature space, might contain images depicting different classes. Therefore, \allicgan samples are likely to follow the class distribution in the conditioning instance neighborhood, generating images that may mismatch with the class label from the conditioning image. To account for this mismatch when using \allicgan samples for training, we employ \textit{soft labels}, which are soft class membership distributions corresponding to each instance-specific neighborhood class distribution. More formally, considering the $i$-th datapoint, its $k$-size neighborhood in the feature space, $\mathcal{A}_i$, and its class label $y_i \in C$ one-hot encoded with the vector $\y_i$, we compute its soft label as:
\begin{equation}
    \y^\mathrm{soft}_i = \frac{1}{k} \sum_{j \in \mathcal{A}_i}{\y_j}, \quad \textrm{with} \quad \y_j \in \{0,1\}^C  \quad \textrm{and} \quad \sum_{c} {\y_{j,c} = 1}.
\end{equation}

\subsubsection{Self-supervised learning} 

We devise a straightforward use of \ours for \textit{multi-view} SSL approaches. Although we chose SwAV~\citep{caron_unsupervised_2020} to perform our experiments, \ours could also be applied to other state-of-the-art methods for \textit{multi-view} SSL like MoCov3~\citep{chen_empirical_2021}, SimCLRv2~\citep{chen_big_2020}, DINO~\citep{caron_emerging_2021} or BYOL~\citep{grill_bootstrap_2020-1}. In this family of approaches, two or more views of the same instance are needed in order to learn meaningful representations. 
These methods construct multi-view positive pairs $(\x_i', \x_i'')^+$ from an image $\x_i$ by applying two independently sampled transformations to obtain $\x'_i=T(\x_i)$ and similarly for $\x_i''$ (see orange branch in Figure~\ref{fig:method}).
To integrate \ours in such pipelines as  an alternative form of  data augmentation, we replace $\x_i'$ with a generated image $\tilde \x_i'$ with probability $p_G$. To this end, we sample an image $\tilde \x_i$ from \icgan conditioned on $\x_i$, and apply further hand-crafted data augmentations $\tilde T$ to obtain $\tilde \x_i' = \tilde T (\tilde\x_i)$.

\paragraph{SwAV pre-training and evaluation.} We follow the SwAV pre-training recipe proposed in \cite{caron_unsupervised_2020}. This recipe comprises the use of random horizontal flipping, random crops, color distortion, and Gaussian blurring for the creation of each image view. In particular, we investigate two augmentation recipes, differing in the use of the {\em multi-crop} augmentation~\citep{caron_unsupervised_2020} or the absence thereof. 
The multi-crop technique augments positive pairs $(\x_i', \x_i'')^+$ with multiple other views obtained from smaller crops of the original image: $(\x_i', \x_i'', \x_i^{\mathrm{small}'''}, \x_i^{\mathrm{small}''''}, ...)^+$. In all experiments, we pre-train SwAV for 200 epochs using the hyper-parameter settings of \cite{caron_unsupervised_2020}. 
To evaluate the learned representation we freeze the ResNet-50 SwAV-backbone and substitute the SSL SwAV head with a linear classification head, which we train supervised on IN validation set for 28 epochs with Momentum SGD and step learning rate scheduler --following the VISSL setup~\citep{goyal2021vissl}.          

\paragraph{Neighborhood augmented SwAV.} To further evaluate the impact of \ours in SSL, we devise an additional baseline, denoted as SwAV-NN, that uses real image neighbors as augmented samples instead of \icgan generations: $(\x_j', \x_i'')^+,\, \x_j \in \mathcal{A}_i$. SwAV-NN is inspired by NNCLR~\citep{dwibedi_little_2021-1}, with the main difference that neighbor images are computed off-line on the whole dataset rather than online using a subset (queue) of the dataset. The nearest neighbors are computed using cosine similarity in the same representation space used for \icgan training. With a probability $p_G$, each image in a batch is paired with a uniformly sampled neighbor in each corresponding neighborhood.

\section{Experimental Evaluation}
\label{sec:results}

In this section, we first present the results obtained in the supervised setting using ResNets and DeiT: in-distribution evaluation on IN (Section~\ref{sssec:ind}); classification results on robustness benchmarks (Section~\ref{sssec:ood}); invariance of learned representations (Section~\ref{sssec:inv}); stratified per-class analysis (Section~\ref{sssec:class}); and sensitivity and ablation studies (Section~\ref{sssec:ablation}). Secondly, we show the SSL results of SwAV on IN (Section~\ref{ssec:ssl_res}).

\subsection{Supervised ImageNet training}
\label{ssec:sup_res}

\subsubsection{In-distribution evaluation}
\label{sssec:ind}

\begin{figure}
\centering
\begin{subfigure}[b]{\textwidth}
    \centering
    \includegraphics[width=.35\textwidth]{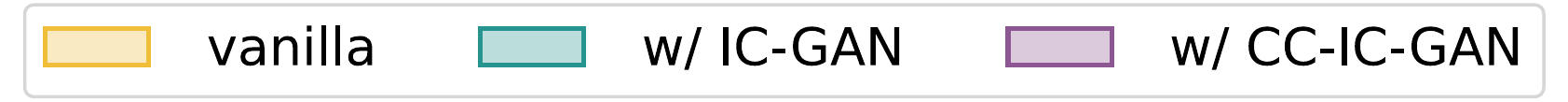}
\end{subfigure}
\begin{subfigure}[b]{.17\textwidth}
    \centering
    \includegraphics[height=4.5cm]{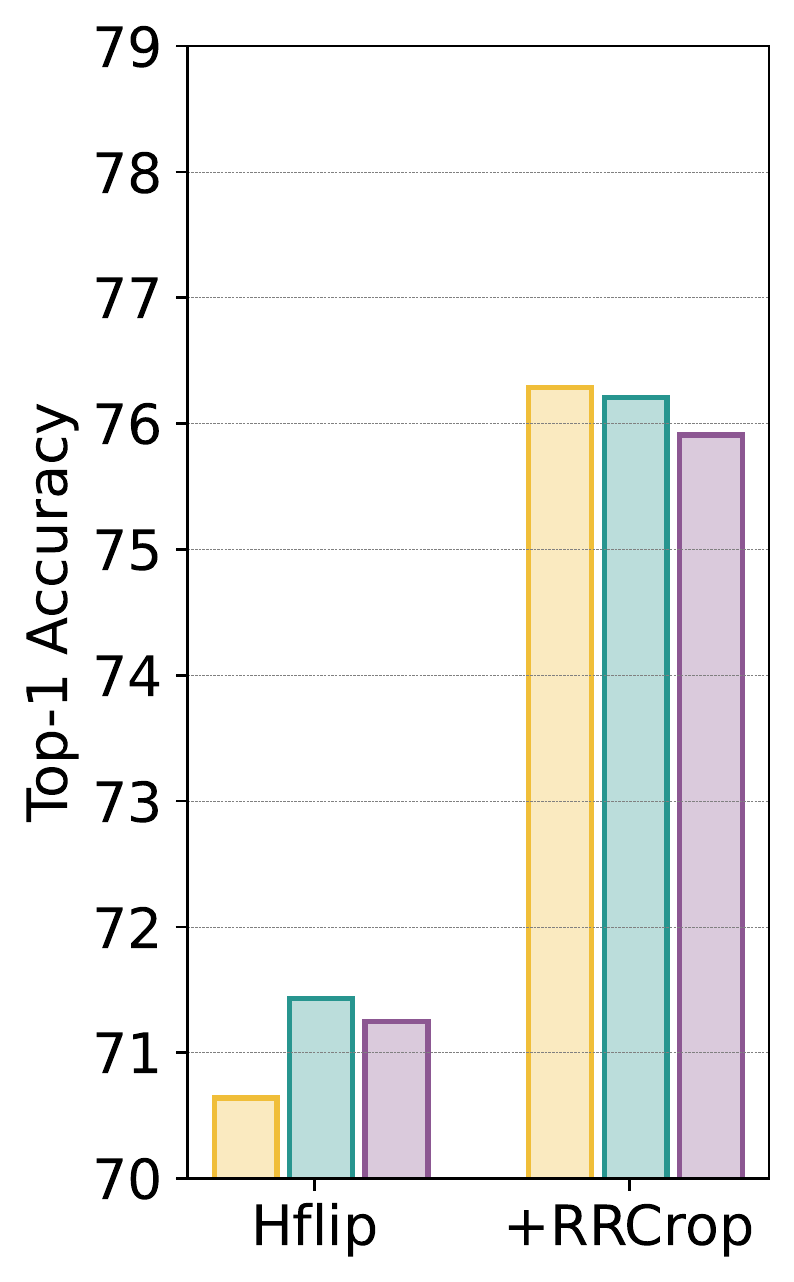}
    \caption{ResNet-50}
    \label{fig:rn50_top1_in1k}
\end{subfigure}
\hfill
\begin{subfigure}[b]{.17\textwidth}
    \centering
    \includegraphics[height=4.5cm]{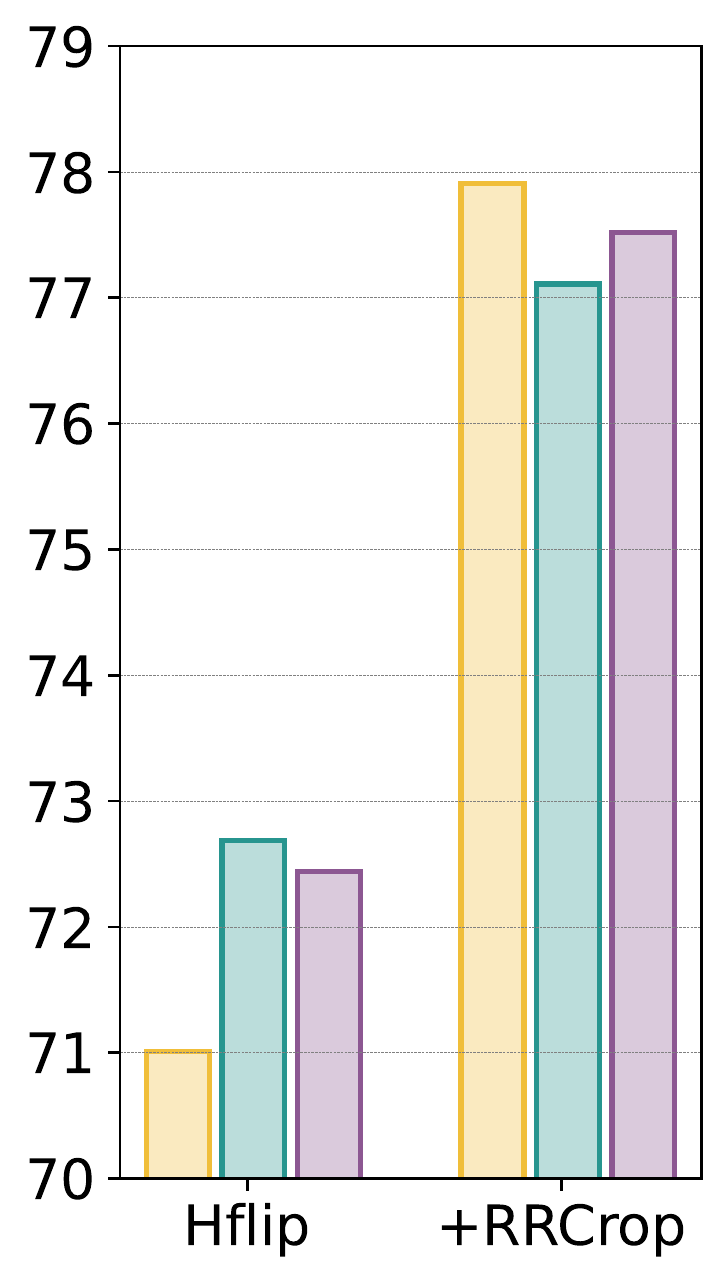}
    \caption{ResNet-101}
    \label{fig:rn101_top1_in1k}
\end{subfigure}
\hfill
\begin{subfigure}[b]{.17\textwidth}
    \centering
    \includegraphics[height=4.5cm]{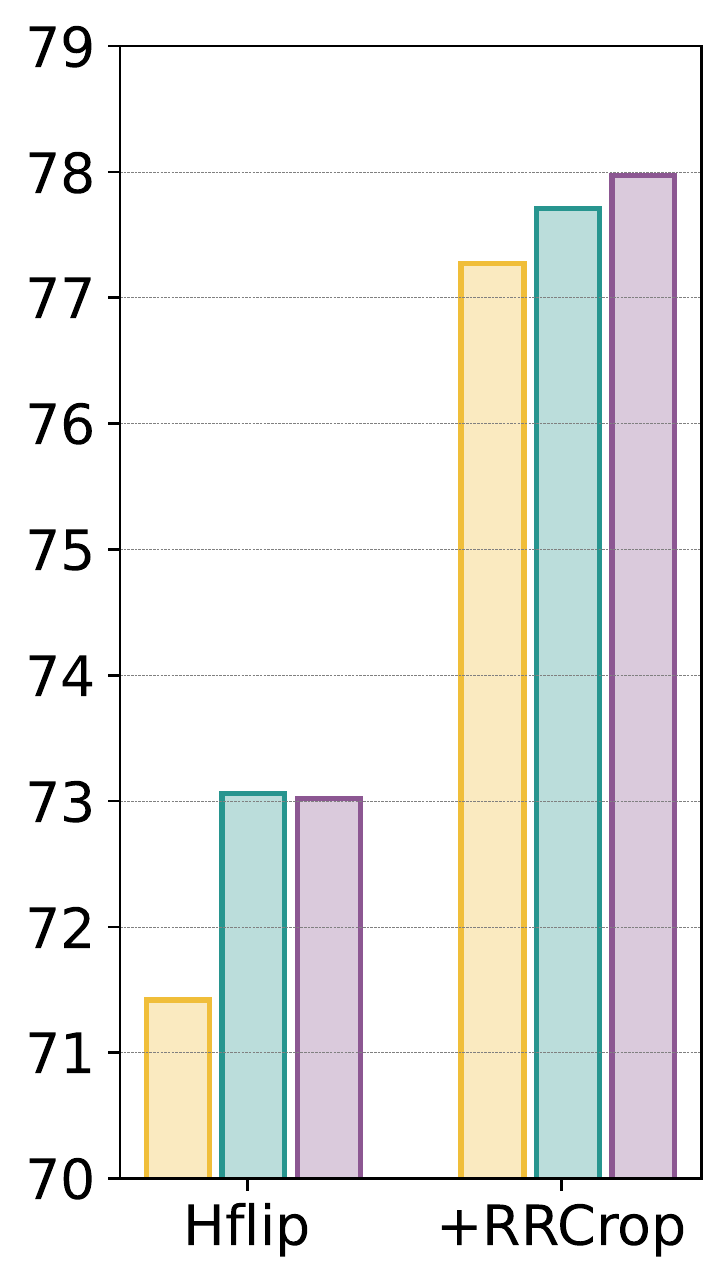}
    \caption{ResNet-152}
    \label{fig:rn152_top1_in1k}
\end{subfigure}
\hfill
\begin{subfigure}[b]{.17\textwidth}
    \centering
    \includegraphics[height=4.5cm]{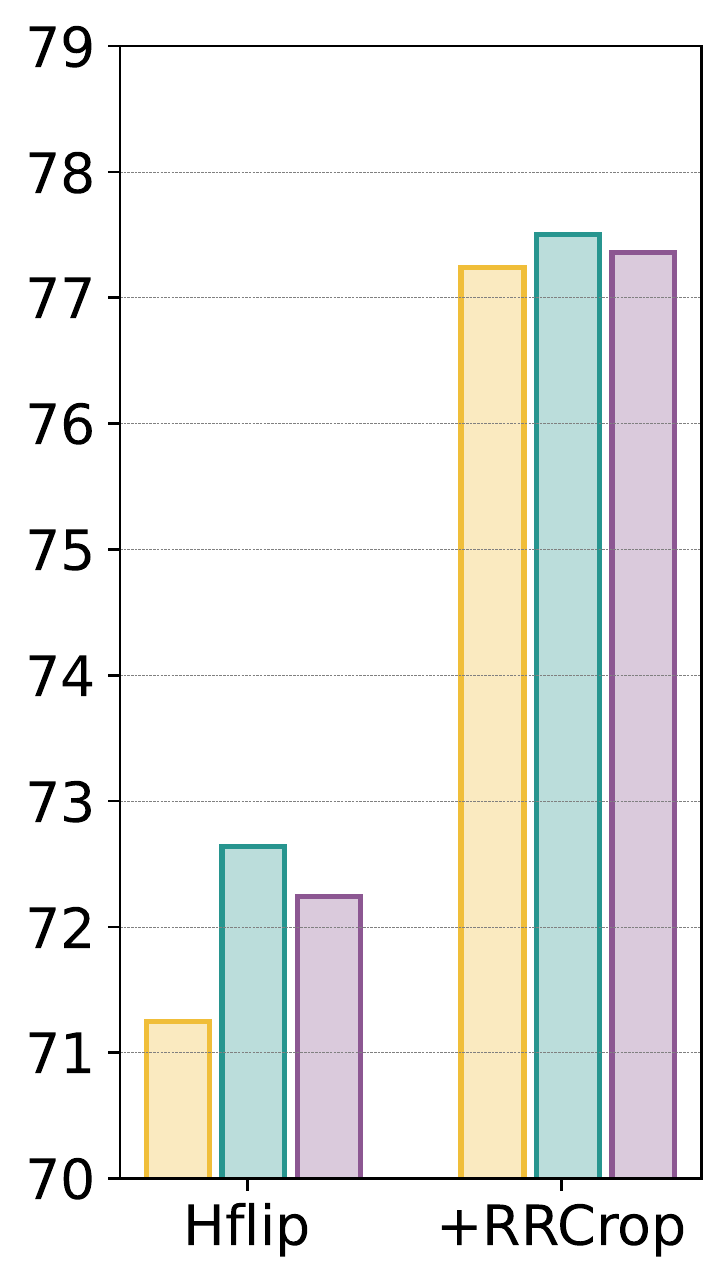}
    \caption{ResNet-50W2}
    \label{fig:rnw2_top1_in1k}
\end{subfigure}
\hfill
\begin{subfigure}[b]{.29\textwidth}
    \centering
    \includegraphics[height=4.5cm]{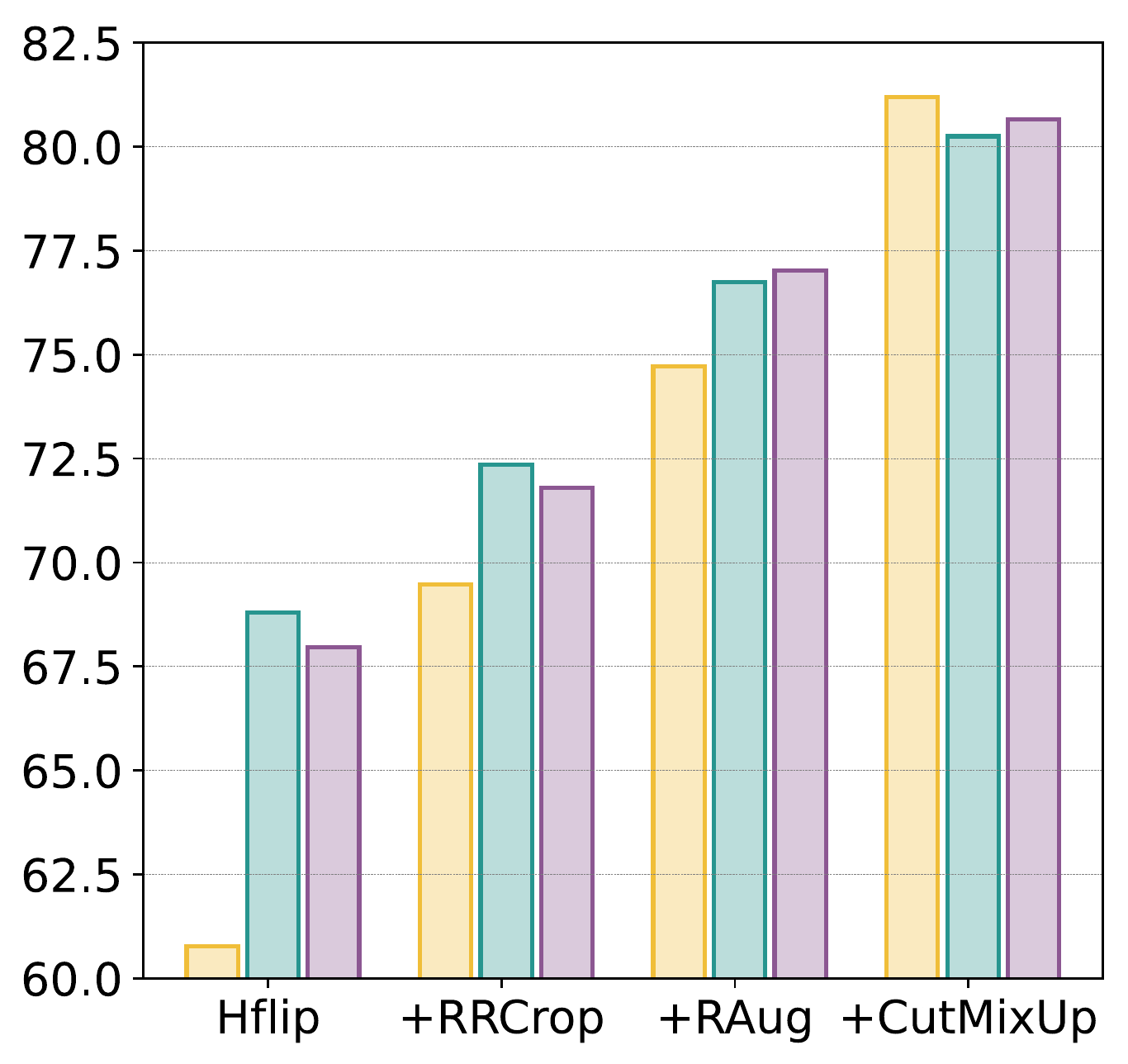}
    \caption{DeiT-B}
    \label{fig:deit_top1_in1k}
\end{subfigure}
    \caption{Impact of \ours when coupled with different data augmentation (DA) recipes for training on IN. Hand-crafted DA techniques are: Hflip --random horizontal flips--, RRCrop --random resized crops--, RAug --RandAugment\citep{cubuk_randaugment_2020-2}--, and CutMixUp~\citep{yun_cutmix_2019, zhang_mixup_2017}. DA techniques are added from left to right, with the right-most column combining all possible DA strategies; i.e, +RRCrop applies RRCrop on top of Hflip.}

    \label{fig:top1_in1k}    
\end{figure}

We start by analyzing the impact of \ours when used in addition to several hand-crafted data augmentation recipes for ResNet-50, ResNet-101, ResNet-152, ResNet-50W2, and DeiT-B. In Figure~\ref{fig:top1_in1k}, we report the top-1 accuracy on the IN validation set for the models under study. 

When training ResNets (see Figures \ref{fig:rn50_top1_in1k}-\ref{fig:rnw2_top1_in1k}), coupling \ours with random horizontal flips (Hflip) results in an overall accuracy boost of 0.5-1.7\%p when using \icgan and 0.3-1.6\%p when using \ccicgan. However, when pairing \ours with random horizontal flips and crops (+RRCrop), we observe an accuracy decrease of 0.1-0.4\%p for ResNet-50 and 0.4-0.8\%p for ResNet-101, while for the bigger capacity ResNets the accuracy is boosted by 0.4-0.7\%p for ResNet-152 and 0.2\%p for ResNet-50W2 with either \icgan or \ccicgan. These results show that \ours is beneficial for higher capacity networks, which might be able to capture the higher image diversity induced by more aggressive DA recipes. These observations align with those of~\cite{kolesnikov_big_2020}, who showed that extremely large dataset sizes may be detrimental to low-capacity networks such as ResNet-50 and ResNet-101, and those of~\cite{steiner_how_2021}, who showed that hand-crafted DA strategies applied on large-scale datasets can also result in performance drops.\looseness-1
 
When training DeiT-B (see Figure \ref{fig:deit_top1_in1k}), the largest model considered --with $\sim4\times$ as many parameters as the ResNet-50-- and which usually needs aggressive regularization strategies~\citep{steiner_how_2021, touvron_deit_2022}, we observe that \ours paired with random horizontal flips (Hflip), random crops (+RRCrop) and RandAugment (+RAug) provides a remarkable accuracy boost of 8.0/7.2\%p, 2.9/2.3\%p, 2.0/2.3\%p for \icgan and \ccicgan, respectively, when compared to only using the hand-crafted DA recipes. However, when extending the recipe by adding more aggressive DA such as CutMixUp, the combination with \ours results in a slight decrease of -0.5\%p and -1.0\%p in accuracy for \ccicgan and \icgan respectively.\looseness-1

Overall, \ours boosts the top-1 accuracy when paired with most of the hand-crafted DA recipes studied and with larger ResNet models, showcasing the promising application of \ours as a DA tool. We hypothesize that \ours acts as an implicit regularizer and as such, when paired with the most aggressive DA recipes for smaller ResNet models and DeiT-B, does not lead to an accuracy improvement, possibly due to an over-regularization of the models. Moreover, we argue that state-of-the-art training recipes with hand-crafted DA strategies have been carefully tuned and, therefore, simply adding \ours into the mix without careful tuning of training and hand-crafted DA hyper-parameters or the optimization strategy might explain the decrease in accuracy for these recipes.

\subsubsection{Robustness evaluation}
\label{sssec:ood}
We present results on six additional datasets: Fake-IN, Fake-IN\textsubscript{CC}, IN-A, IN-R, IN-Real and ObjectNet, to test the robustness of our models. We consider the ResNet-50 model for its ubiquitous use in the literature, as well as the high capacity models ResNet-152 and DeiT-B for their high performance. Results are reported in Table~\ref{tab:ood}.\looseness-1

\begin{table}
\centering
\caption{Robustness evaluation. Top-1 accuracy for ResNet-50, ResNet-152 and DeiT-B, trained on IN and evaluated on: IN-Real (IN-ReaL), Fake-IN, FAKE-IN\textsubscript{CC}, IN-A(IN-A), IN-R (In-R) and ObjectNet. IN (in distribution) results are reported for reference.}
\label{tab:ood}
\resizebox{.8\textwidth}{!}{%
\begin{tabular}{@{}cllccccccc@{}}
\toprule
& & & & \multicolumn{6}{c}{robustness benchmarks}\\
\cmidrule(lr){5-10}
Model & DA base & \ours & IN & IN-ReaL & Fake-IN & Fake-IN\textsubscript{CC} & IN-A & IN-R & ObjectNet \\ \midrule
\multirow{6}{*}{ResNet-50} & \multirow{3}{*}{Hflip} & / & 70.75 & 74.18 & 33.11 & 55.00 & 2.07 & \textbf{23.07}  & \textbf{33.33} \\
 &  & w/ \icgan & \textbf{71.43} & 74.38 & \textbf{39.53} & \textbf{58.06} & 0.97 & 21.46 & 31.93 \\
 &  & w/ \ccicgan & 71.25 & \textbf{74.63} & 33.38 & 57.70 & \textbf{2.23} & 22.47  & \textbf{33.32} \\ \cmidrule(l){2-10} 
 & \multirow{3}{*}{+ RRCrop} & / & \textbf{76.29} & \textbf{77.52} & 37.55 & 61.87 & \textbf{0.61} & \textbf{23.28}  & \textbf{34.67} \\
 &  & w/  \icgan & 76.21 & 77.23 & \textbf{40.65} & 63.14 & 0.45 & 22.99 & 34.45 \\
 &  & w/  \ccicgan & 75.91 & 77.21 & 38.55 & \textbf{65.06} & \textbf{0.60} & 22.70 & 33.52 \\ \midrule
 \multirow{6}{*}{ResNet-152} & \multirow{3}{*}{Hflip} & / & 71.42 & 73.90 & 34.85 & 58.61 & 1.68 & 23.02  & 33.44 \\
 &  & w/ \icgan & \textbf{73.06} & \textbf{75.29} & \textbf{38.28} & 60.52 & \textbf{2.31} & 24.24  & \textbf{35.44} \\
 &  & w/ \ccicgan & 73.02 & 75.09 & 34.39 & \textbf{65.02} & 2.08 & \textbf{24.93}  & 35.28 \\ \cmidrule(l){2-10} 
 & \multirow{3}{*}{+ RRCrop} & / & 77.27 & 78.17 & 37.88 & 63.64 & 1.56 & 24.68 & 35.51 \\
 &  & w/  \icgan & 77.71 & \textbf{78.90} & \textbf{40.56} & 64.11 & 2.32 & \textbf{26.03}  & 38.16 \\
 &  & w/  \ccicgan & \textbf{77.97} & 78.87 & 37.90 & \textbf{66.50} & \textbf{2.57} & 25.96  & \textbf{38.27} \\ \midrule
\multirow{9}{*}{DeiT-B} & \multirow{3}{*}{Hflip + RRCrop} & / & 69.47 & 70.46 & 35.78 & 59.38 & 1.85 & 15.82  & 20.36 \\
 &  & w/ \icgan & \textbf{72.35}  & \textbf{73.59} & \textbf{43.79} & 62.63 & 1.92 & 18.36 & \textbf{24.49} \\
 &  & w/ \ccicgan & 71.79  & 72.74 & 36.41 & \textbf{75.35} & \textbf{2.12} & \textbf{18.74} & 23.78 \\ \cmidrule(l){2-10} 
 & \multirow{3}{*}{+ RAug} & / & 75.28 & 75.56 & 34.53 & 59.75 & 4.36 & 24.18 & 27.31 \\
 &  & w/ \icgan & 76.74 & 77.37 & \textbf{42.80} & 64.49 & 4.34 & 25.11  & \textbf{31.85} \\
 &  & w/ \ccicgan & \textbf{77.02} & \textbf{77.53} & 37.07 & \textbf{75.71} & \textbf{4.86} & \textbf{26.37}  & 31.49 \\ \cmidrule(l){2-10} 
 & \multirow{3}{*}{+ CutMixUp} & / & \textbf{81.19} & \textbf{81.23} & 37.87 & 66.90 & \textbf{11.95} & 31.63  & \textbf{40.56} \\
 &  & w/ \icgan & 80.16 & 80.84 & \textbf{41.78} & 67.33 & 11.14 & 30.85 & 38.59 \\
 &  & w/ \ccicgan & 80.65 & 80.97 & 38.23 & \textbf{76.41} & 11.70 & \textbf{32.23} & 38.58 \\ \bottomrule
\end{tabular}%
}
\end{table}

On Fake-IN and Fake-IN\textsubscript{CC}, datasets composed of generated images obtained with \icgan and \ccicgan respectively, we make two observations. First, the decrease in accuracy of the vanilla ResNets and DeiT-B on these datasets with respect to their IN accuracy highlights a considerable data distribution shift between IN and both Fake-IN and Fake-IN\textsubscript{CC}. Moreover, the accuracy on Fake-IN is significantly lower than on Fake-IN\textsubscript{CC}, as one may expect given the higher generation quality and label preservation of \ccicgan. Secondly, the use of \icgan and \ccicgan provides significant boosts on the respective Fake-IN and Fake-IN\textsubscript{CC} datasets, highlighting the increased robustness of the models trained with \ours while remaining competitive on IN.\looseness-1

On IN-A and IN-R, \ours outperforms the vanilla baselines in most of the settings explored, while especially increasing robustness for larger models -- i.e., ResNet-152 and DeiT-B. However, we notice a better impact of \ccicgan compared to \icgan in 6/7 cases for IN-A and in 5/7 for IN-R, which overturns the results on IN validation where \icgan is better in most cases. This might be explained by the fact that despite being less diverse, \ccicgan generations are more likely to depict the correct class; during training the lower sample diversity reduces the regularization effect providing lower in-distribution gains.  

Finally, on ObjectNet and IN-ReaL, we observe similar trends to those in IN: ResNet-152 and DeiT-B with horizontal flips, random crops and random augment benefit from \ours, leading to an increase in accuracy. This evidences that the improvements that \ours provides in-distribution to high capacity models transfer well when considering a more correct IN labeling, such as the one of IN-ReaL, and more importantly when classifying different objects with several viewpoints and backgrounds, such as those in ObjectNet.\looseness-1

Overall, this robustness evaluation confirms a positive impact of \ours for high-capacity models --already benefiting on in-distribution data--, suggesting that they learn more robust representations which may transfer to unseen datasets. In particular, the generations of \allicgan appear to increase the robustness of the trained models by presenting them with slightly different characteristics from in-distribution images.\looseness-1 %

\subsubsection{Feature invariances}
\label{sssec:inv}

We study the invariance of the learned representations of the ResNet-152 model--our best-performing ResNet model--, to assess whether the \ours's performance boosts could be attributed to more robust learned representations. In particular, we evaluate representation invariances to {\em instance},  {\em viewpoint}, {\em occlusion}, and {\em illumination}, in terms of the top-25 RIS scores. Results are reported in Table~\ref{tab:invariances}.\looseness-1

\begin{table}
\caption{Top-25 Representation Invariance Score (RIS) of the learned representations, evaluated on ImageNet (IN), Pascal3D (P3D), GOT-10K (GOT), and ALOI datasets. $\uparrow$ top-25 RIS means $\uparrow$ invariance.\looseness-1}
\label{tab:invariances}
\centering
\resizebox{\linewidth}{!}{%
\begin{tabular}{@{}cllcccccc@{}}
\toprule

\multirow{2}{*}{Model} & \multirow{2}{*}{DA base} & \multirow{2}{*}{\ours} %
& P3D & P3D & GOT & ALOI & ALOI & ALOI \\
 &  & %
 & Instance & Inst. + View. & Occlusion & Viewpoint & IllumColor & IllumDir \\ \midrule
\multirow{6}{*}{ResNet-152} & \multirow{3}{*}{Hflip} & / %
& 57.05 & 61.15 & \textbf{73.12} & \textbf{81.23} & \textbf{98.91} & \textbf{90.11}  \\
 &  & w/  \icgan %
 & \textbf{58.64} & 61.63 & 70.63 & 78.19 & \textbf{98.90} & 87.27 \\
 &  & w/ \ccicgan %
 & 58.58 & \textbf{62.02} & 71.09 & 78.68 & 98.65 & 87.53 \\ \cmidrule(l){2-9} 
 & \multirow{3}{*}{+ RRCrop} & / %
 & 59.74 & 62.86 & 74.06 & 83.53 & \textbf{99.67} & 90.00 \\
 &  & w/  \icgan %
 & \textbf{62.22} & \textbf{65.87} & \textbf{74.37} & 83.89 & 99.63 & \textbf{91.31} \\
 &  & w/  \ccicgan %
 & 62.01 & 64.93 & 74.36 & \textbf{84.06} & 99.65 & 91.23 \\ 

\bottomrule
\end{tabular}
}
\end{table}

We measure the invariance to {\em instance} on P3D. We observe that \ours always induces a higher RIS. We argue that this result might be expected by considering the ability of \allicgan of populating the neighborhood of each datapoint, i.e., instance. 

Next, we quantify the invariance to {\em viewpoint} using both P3D -- {\em instance + viewpoint} -- and ALOI. In this case, we also notice that \ours generally induces more consistent representations, except when combined with horizontal flips and evaluated on ALOI. Our explanation for the generally higher viewpoint invariance is that \allicgan samples depict slightly different viewpoints of the object present in the conditioning image -- see visual examples in Appendix~\ref{suppl:visual}.

Finally, by looking at the {\em occlusion} invariance on GOT, and the {\em illumination color} and {\em direction} invariance on ALOI, we observe mixed results: in some cases \ours slightly increases the RIS while in some other cases \ours slightly decreases it. This result is perhaps unsurprising as none of these invariances is directly targeted by \ours; and the slight increases observed in some cases could be a side-effect of the larger diversity given by \ours -- e.g., higher occlusion invariance might be due to erroneous generations not containing the object class.\looseness-1

Overall, the invariance analysis highlights that \ours, by leveraging the diversity of the  neighborhood, can be useful not only to regularize the model and achieve better classification accuracy, but also to provide more consistent feature representations across variations of {\em instance} and {\em viewpoint}. Guaranteeing such invariances is likely to lead to a better transferability/robustness of the representations -- as shown in Section~\ref{sssec:ood}.

\subsubsection{Per-class analysis}
\label{sssec:class}

\begin{figure}
\centering
\begin{subfigure}[b]{.49\textwidth}
    \centering
    \includegraphics[width=\textwidth]{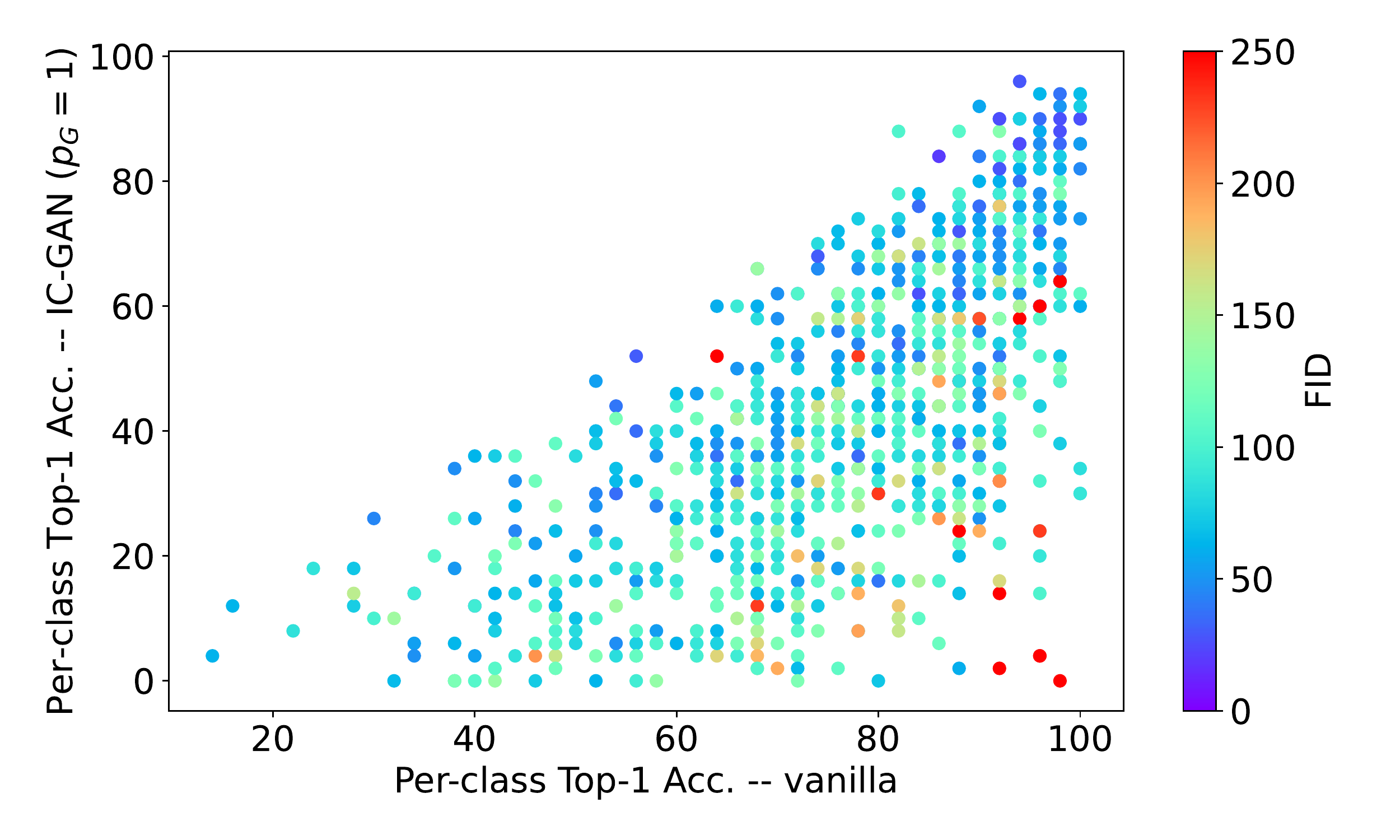}
    \caption{FID -- ResNet-152, \icgan}
    \label{fig:fid_icgan_rn152}
\end{subfigure}
\centering
\begin{subfigure}[b]{.49\textwidth}
    \centering
    \includegraphics[width=\textwidth]{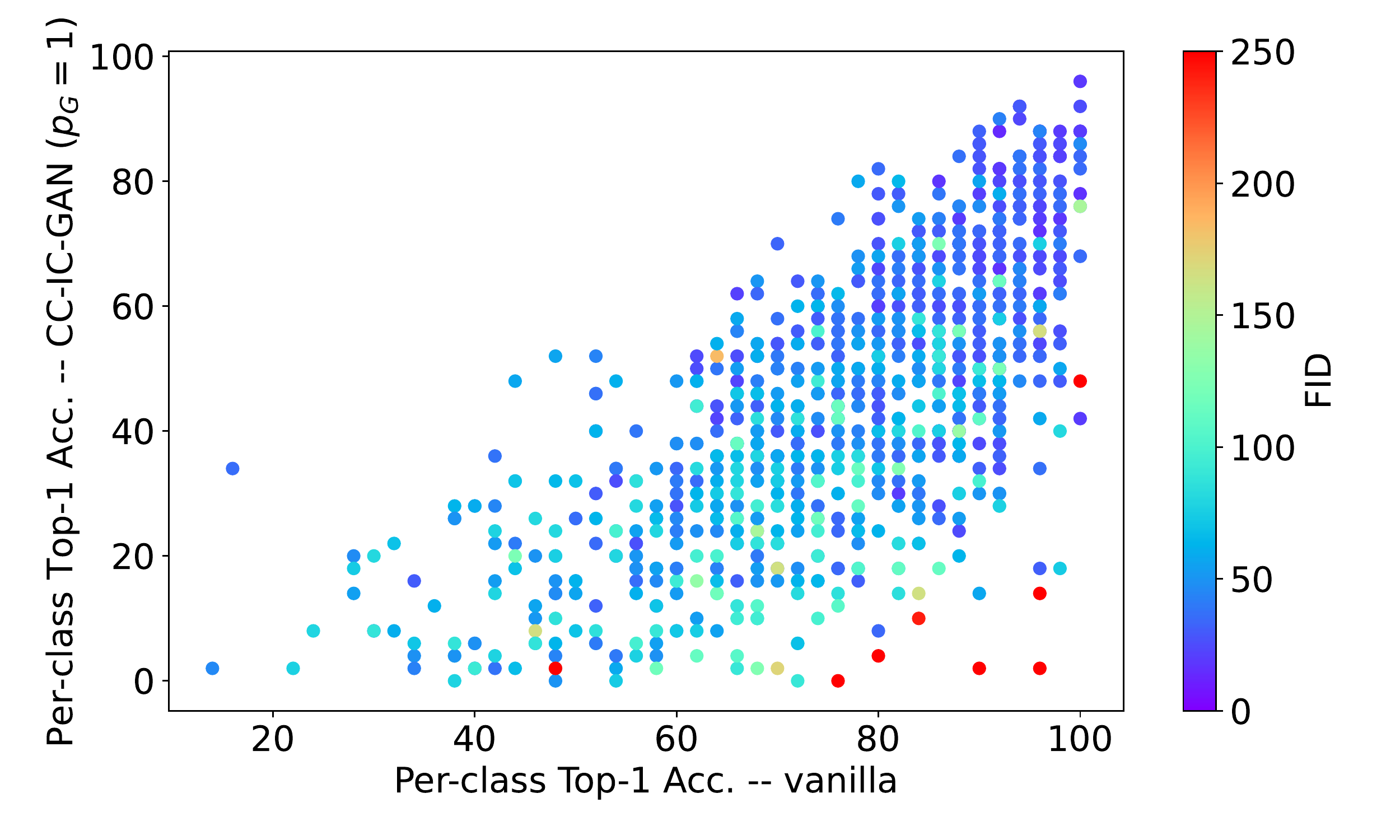}
    \caption{FID -- ResNet-152, \ccicgan}
    \label{fig:fid_ccicgan_rn152}
\end{subfigure}
\hfill
\begin{subfigure}[b]{.49\textwidth}
    \centering
    \includegraphics[width=\textwidth]{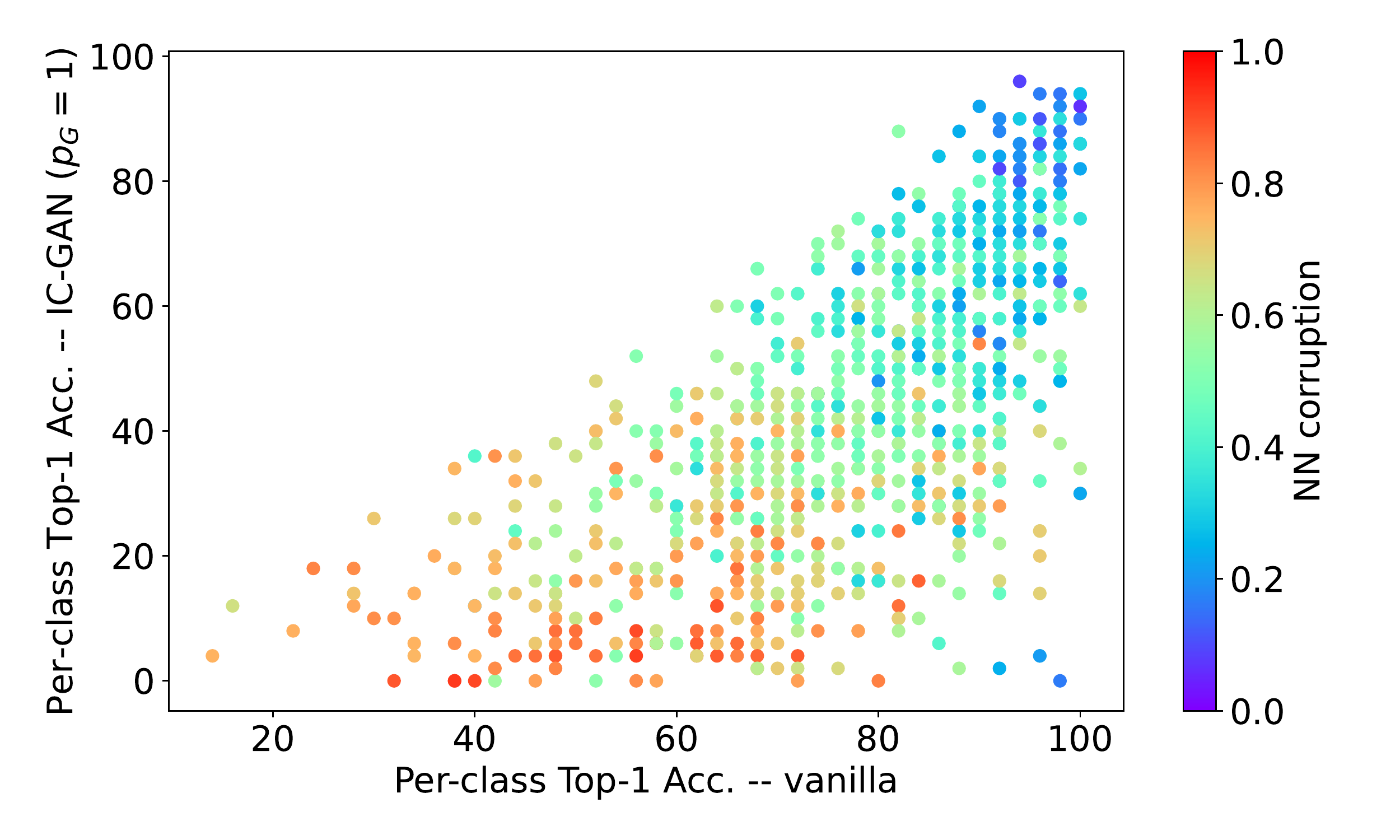}
    \caption{NN corruption -- ResNet-152, \icgan}
    \label{fig:nn_icgan_rn152}
\end{subfigure}
\begin{subfigure}[b]{.49\textwidth}
    \centering
    \includegraphics[width=\textwidth]{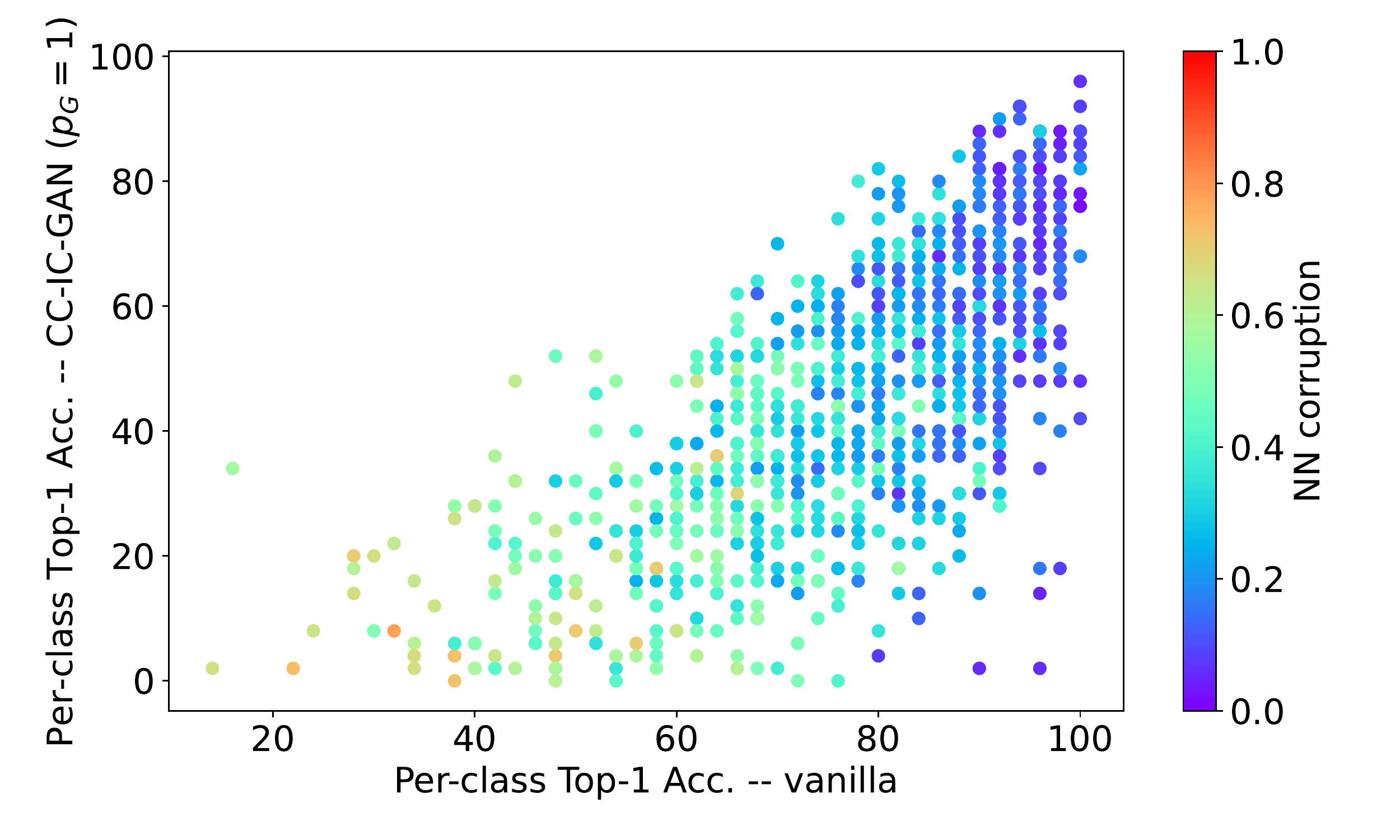}
    \caption{NN corruption -- ResNet-152, \ccicgan}
    \label{fig:nn_ccicgan_rn152}
\end{subfigure}

\caption{Impact of \allicgan's generation quality on per-class performance. (a-b) Per-class FID as a function of per-class top-1 accuracy of the vanilla and \ours models. We observe that higher quality \allicgan generations tend to result in improved performances. (c-d) Per-class NN corruption as a function of per-class top-1 accuarcy of the vanilla and \ours models. We observe that less corrupted classes tend to result in improved performances. ImageNet validation results shown for the ResNet-152 model trained with horizontal flips and random crops.
We limited FID colormap interval to 250 to aid interpretability, while we observed FID values up to 500 for certain classes.}
\label{fig:correlation_fid}
\end{figure}

To further characterize the impact of \ours, we perform a more in-depth analysis by stratifying the ResNet-152 results per class. We compare the per-class FID of \icgan and \ccicgan, as well as their NN corruption, with the top-1 accuracy per class of a vanilla model -- trained without \ours --  and a model trained only with generated samples -- i.e., using \ours with $p_G = 1.0$ -- with the goal of better understanding the impact of \allicgan's generations on the model's performance. Results are reported in Figure~\ref{fig:correlation_fid}. Note that the exclusive use of generated samples leads to rather low top-1 accuracy: $\sim$43\% and $\sim$46\% for ResNet-152 %
when using \icgan and \ccicgan respectively.\looseness-1

\begin{figure}[t]
     \centering

        \begin{subfigure}[b]{1.0\textwidth}
         \centering
         {Class "ringlet butterfly", \icgan model, FID\textsubscript{class} = 310}
     \end{subfigure}
    \begin{subfigure}[b]{0.03\textwidth}
        \raisebox{\dimexpr 2.3cm-\height}{\rotatebox[origin=c]{90}{{Conditioning}}}
     \end{subfigure}
    \begin{subfigure}[b]{0.150\textwidth}
       \centering
             \includegraphics[width=\textwidth]{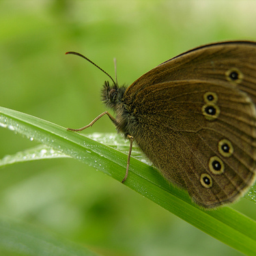}
    \end{subfigure}
        \begin{subfigure}[b]{0.150\textwidth}
           \centering
         \includegraphics[width=\textwidth]{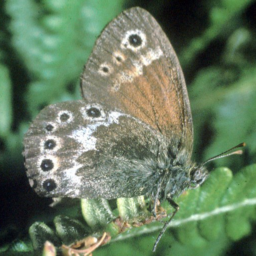}
     \end{subfigure}
             \begin{subfigure}[b]{0.15\textwidth}
                \centering
         \includegraphics[width=\textwidth]{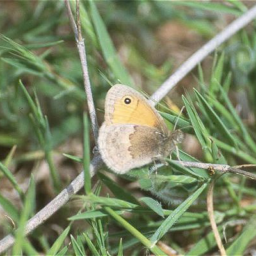}
     \end{subfigure}
             \begin{subfigure}[b]{0.15\textwidth}
                \centering
         \includegraphics[width=\textwidth]{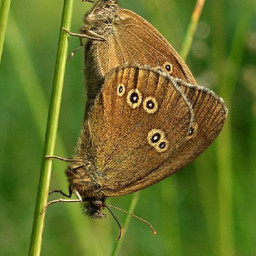}
     \end{subfigure}
             \begin{subfigure}[b]{0.15\textwidth}
                \centering
         \includegraphics[width=\textwidth]{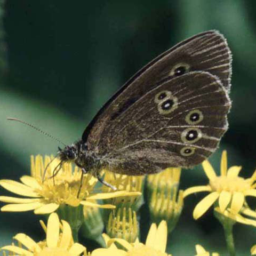}
     \end{subfigure}
             \begin{subfigure}[b]{0.15\textwidth}
                \centering
         \includegraphics[width=\textwidth]{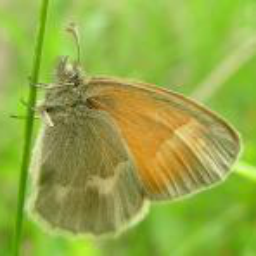}
     \end{subfigure}
     
       \begin{subfigure}[b]{0.03\textwidth}
        \raisebox{\dimexpr2.3cm-\height}{\rotatebox[origin=c]{90}{{Gen. samples}}}
     \end{subfigure}
    \begin{subfigure}[b]{0.150\textwidth}
       \centering
             \includegraphics[width=\textwidth]{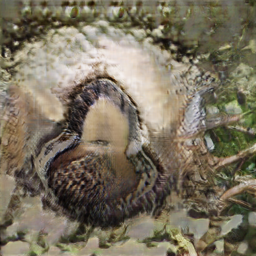}
    \end{subfigure}
        \begin{subfigure}[b]{0.150\textwidth}
           \centering
         \includegraphics[width=\textwidth]{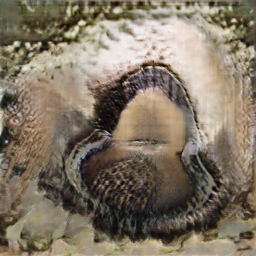}
     \end{subfigure}
             \begin{subfigure}[b]{0.15\textwidth}
                \centering
         \includegraphics[width=\textwidth]{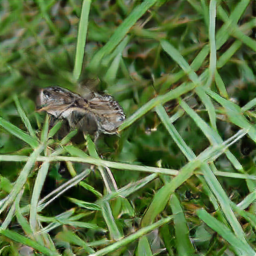}
     \end{subfigure}
             \begin{subfigure}[b]{0.15\textwidth}
                \centering
         \includegraphics[width=\textwidth]{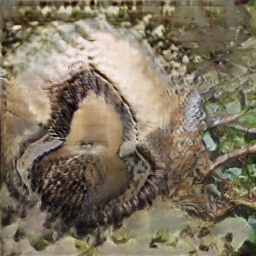}
     \end{subfigure}
             \begin{subfigure}[b]{0.15\textwidth}
                \centering
         \includegraphics[width=\textwidth]{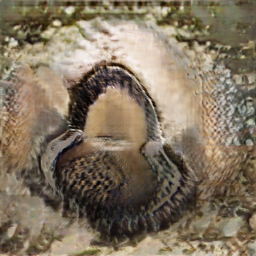}
     \end{subfigure}
             \begin{subfigure}[b]{0.15\textwidth}
                \centering
         \includegraphics[width=\textwidth]{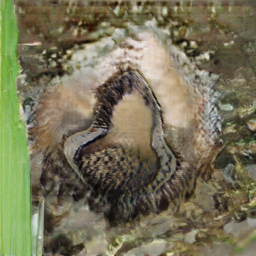}
     \end{subfigure}

      \begin{subfigure}[b]{1.0\textwidth}
         \centering
         {Class "typewriter", \ccicgan model, FID\textsubscript{class} = 385}
     \end{subfigure}
    \begin{subfigure}[b]{0.03\textwidth}
        \raisebox{\dimexpr 2.3cm-\height}{\rotatebox[origin=c]{90}{{Conditioning}}}
     \end{subfigure}
    \begin{subfigure}[b]{0.150\textwidth}
       \centering
             \includegraphics[width=\textwidth]{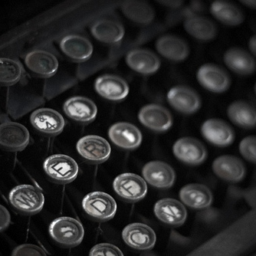}
    \end{subfigure}
        \begin{subfigure}[b]{0.150\textwidth}
           \centering
         \includegraphics[width=\textwidth]{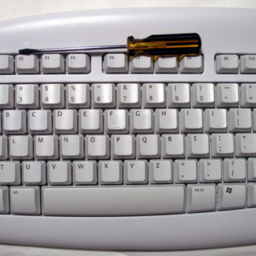}
     \end{subfigure}
             \begin{subfigure}[b]{0.15\textwidth}
                \centering
         \includegraphics[width=\textwidth]{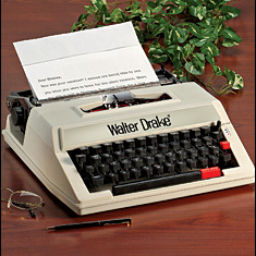}
     \end{subfigure}
             \begin{subfigure}[b]{0.15\textwidth}
                \centering
         \includegraphics[width=\textwidth]{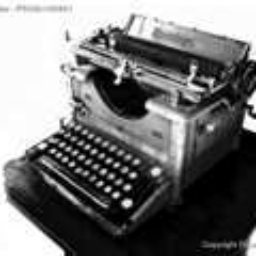}
     \end{subfigure}
             \begin{subfigure}[b]{0.15\textwidth}
                \centering
         \includegraphics[width=\textwidth]{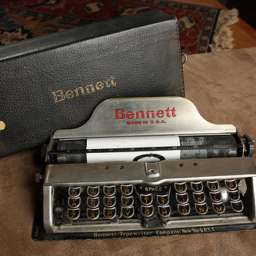}
     \end{subfigure}
             \begin{subfigure}[b]{0.15\textwidth}
                \centering
         \includegraphics[width=\textwidth]{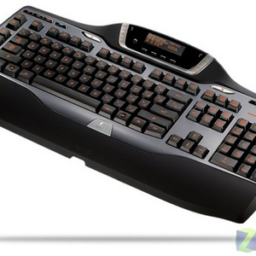}
     \end{subfigure}
     
       \begin{subfigure}[b]{0.03\textwidth}
        \raisebox{\dimexpr2.3cm-\height}{\rotatebox[origin=c]{90}{{Gen. samples}}}
     \end{subfigure}
    \begin{subfigure}[b]{0.150\textwidth}
       \centering
             \includegraphics[width=\textwidth]{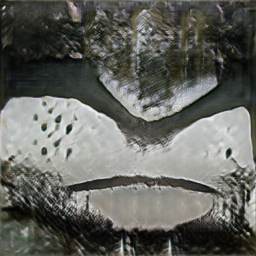}
    \end{subfigure}
        \begin{subfigure}[b]{0.150\textwidth}
           \centering
         \includegraphics[width=\textwidth]{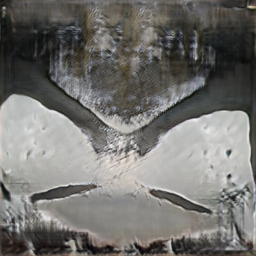}
     \end{subfigure}
             \begin{subfigure}[b]{0.15\textwidth}
                \centering
         \includegraphics[width=\textwidth]{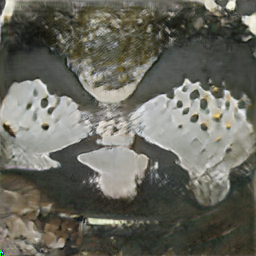}
     \end{subfigure}
             \begin{subfigure}[b]{0.15\textwidth}
                \centering
         \includegraphics[width=\textwidth]{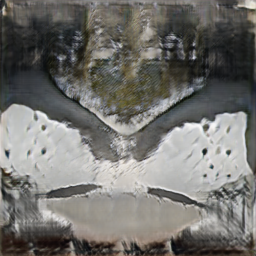}
     \end{subfigure}
             \begin{subfigure}[b]{0.15\textwidth}
                \centering
         \includegraphics[width=\textwidth]{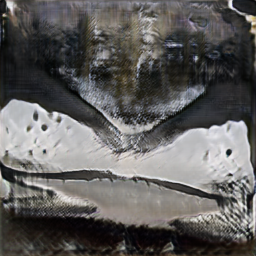}
     \end{subfigure}
             \begin{subfigure}[b]{0.15\textwidth}
                \centering
         \includegraphics[width=\textwidth]{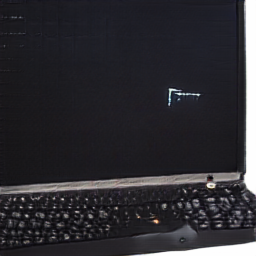}
     \end{subfigure}
           
        \caption{Conditioning sample and one of its generated samples with \icgan or \ccicgan, illustrating the mode collapse for some classes in ImageNet. Note that the mode collapse is evidenced in different classes for \icgan and \ccicgan.}
        \label{fig:icg_ccicg_mode_coll}
\label{fig:mode_collapse}
\end{figure}

Figures~\ref{fig:fid_icgan_rn152} and~\ref{fig:fid_ccicgan_rn152} present the per-class FID of \allicgan as a function of per-class top-1 accuracy of the vanilla baseline and the \ours models. We observe that \ours tends to exhibit higher accuracy for classes with lower FID values, and lower accuracy for classes with higher FID values overall. In particular, classes for which generated images have good quality and diversity (e.g., $\sim$ 50 FID or lower) tend to achieve high top-1 accuracy for both the vanilla model and the one trained with only generated data. Conversely, when the FID of a class is high, its per-class accuracy oftentimes drops for the \allicgan-trained models, whereas the vanilla model remains performant. Perhaps unsurprisingly, this evidences that leveraging image generations of poorly modeled classes to train the ResNet-152 model hurts the performance. Moreover, we note that there are more classes with very high FID ($\sim$ 200 or higher) for \icgan than \ccicgan. Intuitively, this could be explained by the fact that \ccicgan uses labels to condition the model and appears to be less prone to mode collapse (see Figure~\ref{fig:mode_collapse}). %

We additionally observe in Figures~\ref{fig:nn_icgan_rn152} and~\ref{fig:nn_ccicgan_rn152} that the low accuracies of the model trained with generated data can be partially explained by the NN corruption: classes with less corrupted neighborhoods tend to exhibit higher top-1 accuracies than the more corrupted ones. However, we observe some specific cases of classes with low corruption which result in very low accuracy when considering the model trained with all generated samples (see the bottom-right corner of the plots). This could be explained by the mode collapse that \allicgan experience, as we see that those same classes generally have very high FID ($>$ 200) in Figures~\ref{fig:fid_icgan_rn152} and~\ref{fig:fid_ccicgan_rn152}.\looseness-1

In this analysis, we shed some light on the problematic \allicgan modeling of certain classes. 
We believe that computing stratified results for generative models might be a good practice to be adopted by the community, as also supported by \cite{ravuri_classification_2019}. Nevertheless, the observed positive correlation between high classification accuracy and \allicgan's generation quality --studied through the lens of per-class FID and NN corruption-- constitutes a promising result to improve the effectiveness of \ours. To this end, we ran an additional experiment where we avoid applying \ours on classes having very high FID (>= 150), i.e., where \allicgan has very low generation quality. We report the results in Appendix~\ref{suppl:results}. Notably, the impact of leveraging \ours could be potentially improved by increasing the generation quality of the \allicgan's poorly modeled classes. These findings improve upon those of \cite{ravuri_classification_2019}, where a pre-trained BigGAN showed little to no correlation between FID and classification accuracy in a similar setting, strengthening the position of instance-conditioned models such as \allicgan. We observe similar trends for DeiT-B (see Appendix~\ref{suppl:results}).

\subsubsection{Sensitivity and ablation studies}
\label{sssec:ablation}

\begin{figure}
\centering
\begin{subfigure}[b]{\textwidth}
    \centering
    \includegraphics[width=\textwidth]{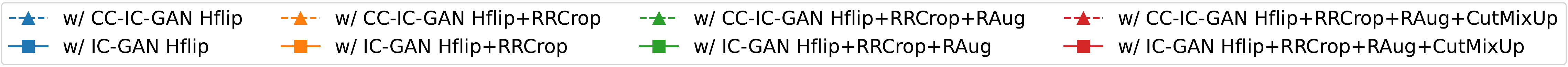}
\end{subfigure}
\hfill
\begin{subfigure}[b]{.32\textwidth}
    \centering
    \includegraphics[height=4.5cm]{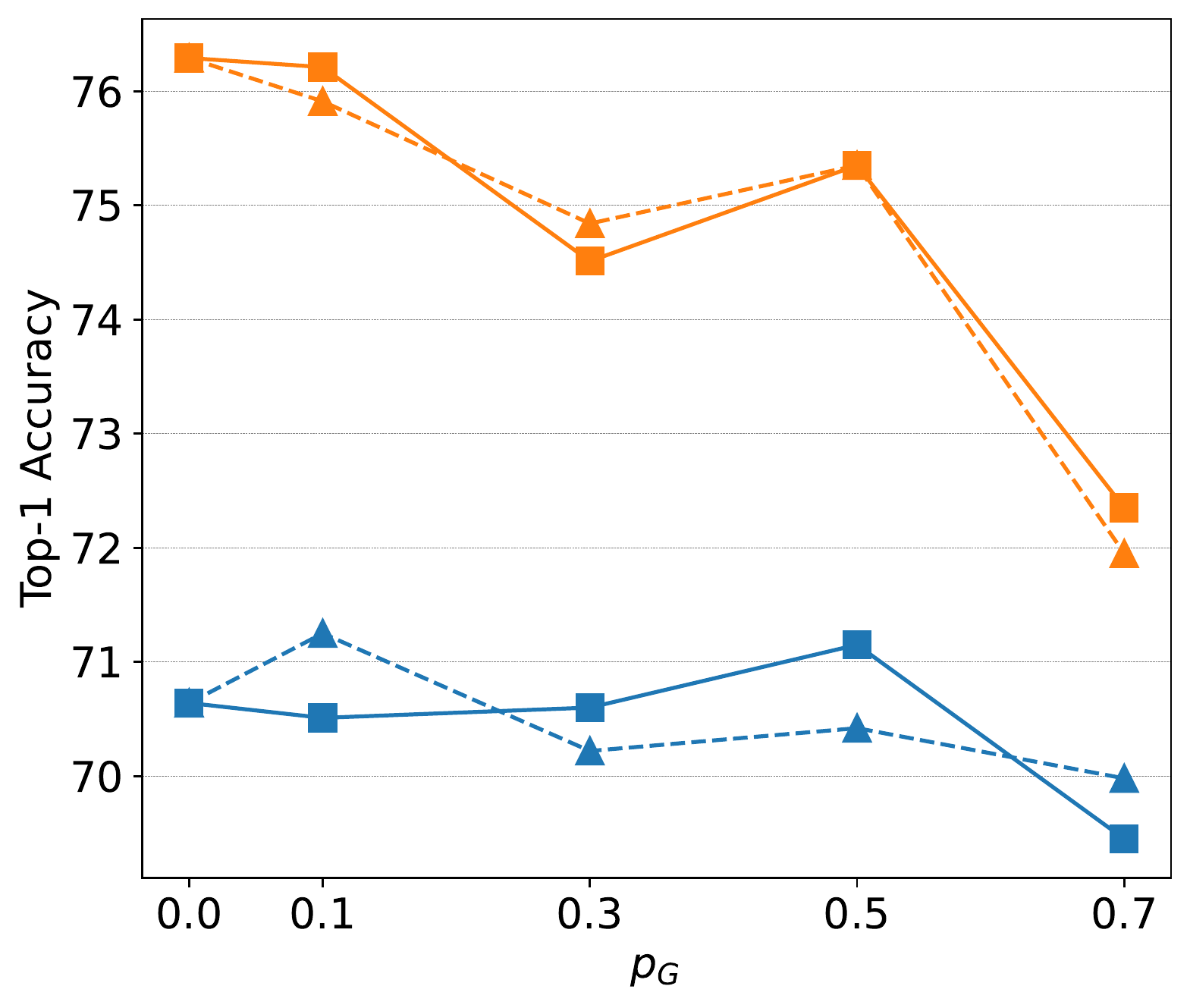}
    \caption{ResNet-50}
    \label{fig:rn50_top1_prob_in1k}
    \end{subfigure}
    \hfill
\begin{subfigure}[b]{.32\textwidth}
    \centering
    \includegraphics[height=4.5cm]{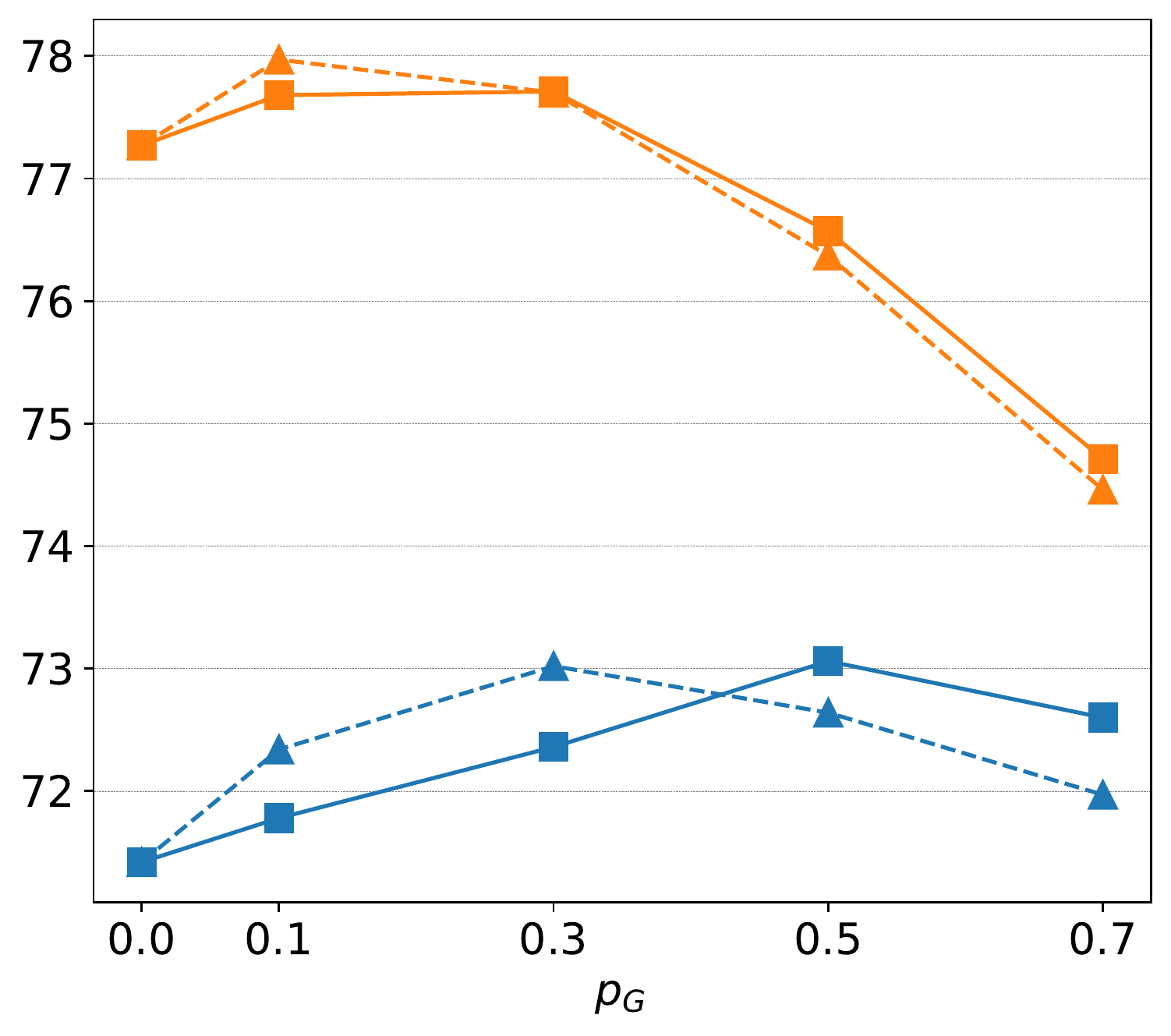}
    \caption{ResNet-152}
    \label{fig:rn152_top1_prob_in1k}
\end{subfigure}
\hfill
\begin{subfigure}[b]{.32\textwidth}
    \centering
    \includegraphics[height=4.5cm]{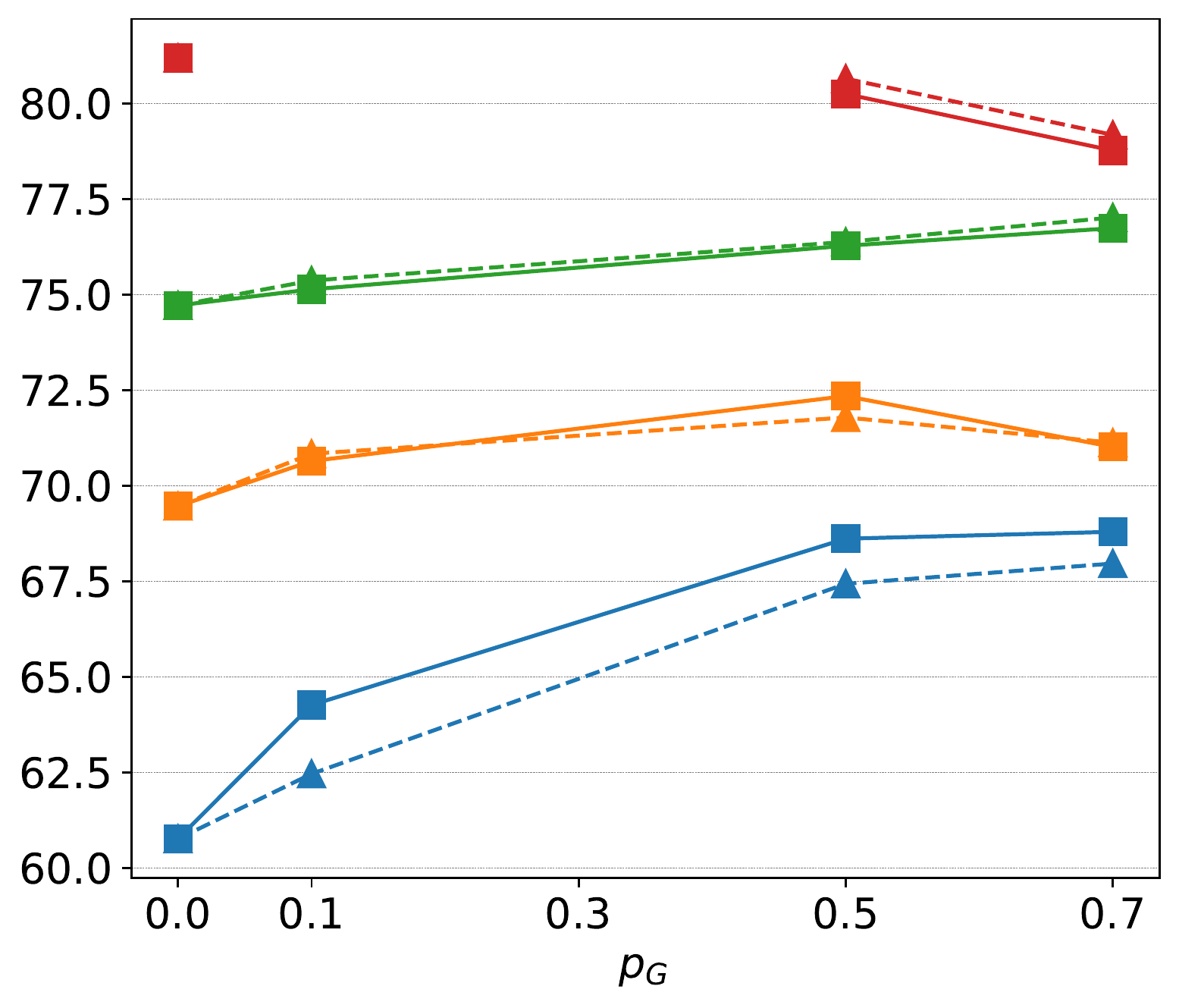}
    \caption{DeiT-B}
    \label{fig:deit_top1_prob_in1k}
\end{subfigure}
\caption{Sensitivity study for the probability $p_G$ of applying \ours. The missing datapoint in red curves of panel (c) are due to trainings not converging.}
\vspace{-0.4cm}
\label{fig:prob}
\end{figure}

\paragraph{Probability $p_G$.} We study the impact of the probability of applying \ours, $p_G$, in Figure~\ref{fig:prob}. We consider our best ResNet model as well as DeiT-B. We further include the study on ResNet-50 as a sanity check to validate our previous over-regularization hypothesis. When coupling \ours with horizontal flip to train the ResNet-50 model (Figure~\ref{fig:rn50_top1_prob_in1k}), we observe that a probability of $p_G=0.1$ and $p_G=0.5$ achieve the best results for \ccicgan and \icgan, respectively. However, when adding random crops to the recipe, ResNet-50 no longer benefits from \ours and obtains the best results for $p_G=0$, highlighting the potential over-regularization suffered by low-capacity models as discussed in section~\ref{sssec:ind}. When it comes to ResNet-152 (Figure~\ref{fig:rn152_top1_prob_in1k}), we observe that the overall accuracy increases until achieving its peak value for some $p_G$ and then starts decreasing. 
More precisely, \ccicgan shows optimal $p_G$ for lower values, 0.1 and 0.3, whereas \icgan benefits from the higher probability values 0.3 and 0.5. Note that in both cases, \ours coupled with random horizontal flips and crops requires lower $p_G$ values than \ours coupled with horizontal flips only, emphasizing the benefit of \ours especially when leveraging soft augmentation strategies. For DeiT-B architecture (Figure~\ref{fig:deit_top1_prob_in1k}), we note that increasing $p_G$ values mostly result in better accuracy when using all DA recipes except the strongest one containing CutMixUp. This trend might be due to the higher capacity of the DeiT-B model that combined with the lower architectural inductive bias -- i.e., no convolution -- requires stronger regularization on IN. 
This shows the benefits of using \ours to regularize  training, especially for architectures prone to overfitting, which require higher $p_G$ values.

\begin{wraptable}{ri}{9cm}
\vspace{-.5cm}
\caption{\small CutMixUp ablation when training DeiT-B with Hflip+RRCrop+RAug and \ours. All models trained with a label smoothing of 0.1. *: Failed to converge.
}
\vspace{-0.2cm}
\label{tab:cmu_ablation}
\centering
\resizebox{.7\linewidth}{!}{%
\begin{tabular}{@{}ccccc@{}}
\toprule
\icgan & \ccicgan & CutMix & MixUp & Top-1 \\ 
\midrule
\xmarkg & \xmarkg & \cmarkg & \cmarkg & 81.2 \\
\xmarkg & \xmarkg & \cmarkg & \xmark & 81.7 \\
\xmarkg & \xmarkg & \xmark & \cmarkg & 80.3 \\
\xmarkg & \xmarkg & \xmark & \xmark & 77.4 \\ 
\midrule
\cmark & \xmarkg & \cmarkg & \cmarkg & 80.2 \\
\cmark & \xmarkg & \cmarkg & \xmark & 78.5 \\
\cmark & \xmarkg & \xmark & \cmarkg & * \\
\cmark & \xmarkg & \xmark & \xmark & 78.2 \\
\midrule
\xmarkg & \cmark & \cmarkg & \cmarkg & 80.7 \\
\xmarkg & \cmark & \cmarkg & \xmark & 78.2 \\
\xmarkg & \cmark & \xmark & \cmarkg & * \\
\xmarkg & \cmark & \xmark & \xmark & 78.1 \\
\bottomrule
\end{tabular}
}
\vspace{-.5cm}
\end{wraptable}
\vspace{-0.2cm}
\paragraph{CutMixUp components.} In Table~\ref{tab:cmu_ablation} we present an ablation of the two components in CutMixUp -- CutMix and MixUp -- when training DeiT-B models with and without \ours. To facilitate convergence when using CutMix and/or MixUp, we follow~\cite{touvron_deit_2022} and perform label smoothing. Moreover, we leverage the {\em soft labels} introduced in section~\ref{ssec:sup_res} for all experiments using \ours. We observe that removing CutMix results in a decrease in accuracy of 0.9\%p, whereas removing MixUP increases the performance by +0.5\%p. Removing both CutMix and MixUp results in the lowest accuracy. When considering \ours, we observe that results are consistently better than those of the baseline model when not using CutMix nor MixUp. Leveraging CutMixUp appears to be beneficial but the induced accuracy gains remain lower than those experienced by the baseline model, suggesting that additional tuning of CutMixUp might be required when coupled with \ours.\looseness-1

\subsection{Self-supervised ImageNet training}
\label{ssec:ssl_res}

Given that \icgan can be trained without labels, and can synthesize images without the need of class categories, we employ it to extend multi-view SSL pre-training on IN \citep{he_momentum_2020, chen_simple_2020, caron_unsupervised_2020} by integrating \ours into the standard hand-crafted DA recipes. In Table \ref{tab:ssl} we report the accuracy scores obtained on the IN validation set when testing the pre-trained SSL methods via linear classifier evaluation. %

\begin{wraptable}{r}{.6\linewidth}
\vspace{-.5cm}
\caption{SwAV accuracy on ImageNet validation using different image sources and DA methods. For each view (view1 and view2) in the multi-view SSL setup, the \textit{image source} can be: \textit{Original} - real image from the dataset, or \textit{NN} - a randomly sampled neighbor (k-NN with k=50). \textit{\ours} is applied on top of the image source. \textit{RRC}:  RandomResizedCrop + ColorDistortion + Gaussian Blurring. RRC can produce a \textit{single}-crop or \textit{multi}-crop. 
 }
\centering
\resizebox{\linewidth}{!}{%
\begin{tabular}{@{}ccc|cc|c@{}}
\toprule
\multicolumn{3}{c|}{View1} & \multicolumn{2}{c|}{View2} & \multirow{2}{*}{Top-1} \\
Image source & \ours & RRC & Image source & RRC &  \\ \midrule
Original & \xmarkg & single & Original & single & 67.96 \\
Original & \cmark & - & Original & single & 68.90 \\
NN & \xmarkg & single & Original & single & \textbf{70.06} \\

\midrule
Original & \xmarkg & single & Original & multi & 73.64 \\
Original & \cmark & single & Original & multi & 71.72 \\
NN & \xmarkg & single & Original & multi & \textbf{73.73} \\
\bottomrule
\end{tabular}
}
\vspace{-.5cm}
\label{tab:ssl}
\end{wraptable}
 
\begin{figure}
\centering
\begin{subfigure}[b]{.3\textwidth}
    \centering
    {\footnotesize Conditioning}\hspace{.8cm}{\footnotesize Gen. sample}
    \includegraphics[width=.48\textwidth]{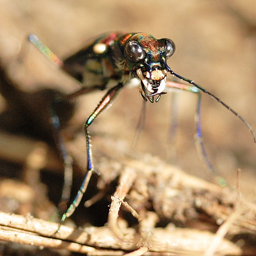}
    \includegraphics[width=.48\textwidth]{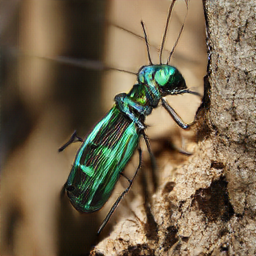}
    \caption{Tiger beetle}
\end{subfigure}
\hfill
\begin{subfigure}[b]{.3\textwidth}
    \centering
    {\footnotesize Conditioning}\hspace{.8cm}{\footnotesize Gen. sample}
    \includegraphics[width=.48\textwidth]{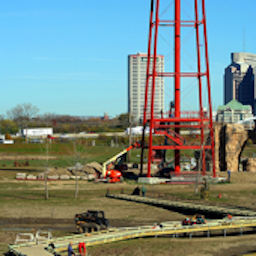}
    \includegraphics[width=.48\textwidth]{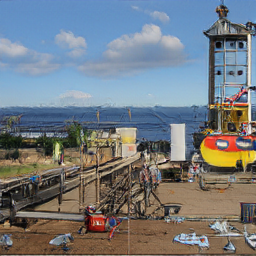}
    \caption{Water tower}
\end{subfigure}
\hfill
\begin{subfigure}[b]{.3\textwidth}
    \centering
    {\footnotesize Conditioning}\hspace{.8cm}{\footnotesize Gen. sample}
    \includegraphics[width=.48\textwidth]{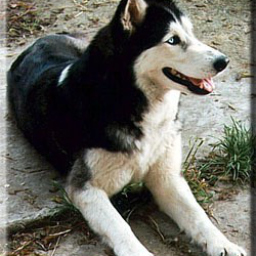}
    \includegraphics[width=.48\textwidth]{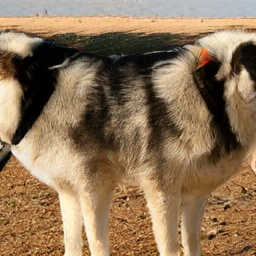}
    \caption{Siberian husky}
\end{subfigure}
    \caption{Example of three \icgan generations, one (a) qualitatively and semantically correct, one (b) semantically incorrect, and one (c) with poor quality. Left image in each pair is the conditioning, where the red square shows the central crop actually used by \icgan.}
    \label{fig:icg_example}
\end{figure}

We observe a clear difference between the two SwAV settings: (i) with single-crop, and (ii) with multi-crop. In the former case, the use of \ours boosts the top-1 classification accuracy of the linear evaluation probe by $\sim$1\%p. On the contrary, when \ours is combined with multi-crop RRC (crops with different sizes and zooms of the original image), we observe a detrimental effect, with roughly a 2\%p accuracy drop. It is worth noting that the multi-crop transformation already results in significant variations from the original image, in some cases even changing its semantics. Hence, combining \ours and multi-crop might result in extreme augmentation diversity (see example reported in Appendix \ref{suppl:visual}).
Moreover, we recall that the SwAV model has a ResNet-50 backbone, whose capacity might be too low to capture such a large image diversity, as discussed in the supervised IN training in Section~\ref{ssec:sup_res}. A further confirmation of this hypothesis may come from the results of SwAV-NN, which are positive for single-crop, with a $\sim$ 2\%p increase, while remaining on-par for multi-crop (+0.1\%); using real image neighbors instead of \icgan-generated ones (last row of Table \ref{tab:ssl}) does not increase the diversity in the training distribution, requiring less model capacity. Moreover, a non-negligible role might be played by the quality of \allicgan generations that for some instances (or classes) might be poor or semantically far from the conditioning, as observed in Section~\ref{sssec:class} and visually shown in Figure \ref{fig:icg_example}.\looseness-1

Overall, this empirical analysis reveals that \ours improves SSL training only when single-crop augmentation is adopted, while the use of stronger hand-crafted DA (multi-crop) on top of \icgan-generated images is detrimental. We point out that in NNCLR~\citep{dwibedi_little_2021-1} no gains from the combination of the multi-crop augmentation with the neighbor-based augmentation were found, and our SwAV-NN experiments confirm this finding.

\section{Conclusions}

We have studied the potential of Instance-Conditioned GAN (\icgan) as a data augmentation technique in state-of-the-art training recipes for visual representation learning. 
Specifically, we have presented \ours, a data augmentation module which leverages the generations of \allicgan and integrates them seamlessly with standard handcrafted data augmentation recipes. 
We have validated \ours in the context of image classification, leveraging supervised learning with ResNets~\citep{he_deep_2016} and DeiT-B~\citep{touvron_training_2021}, as well as self-supervised learning with SwAV~\citep{caron_unsupervised_2020}. The results of this validation have unveiled a beneficial impact of \ours, especially for higher capacity networks and when coupling \allicgan augmentations with soft hand-crafted augmentation strategies, suggesting \ours may act as an implicit regularizer for the models. Additionally, we have found that the representations learned when training models with \ours are more robust when transferred to unseen datasets and more invariant across variations in instance and viewpoint, as a byproduct of augmenting the dataset with generated images obtained with \allicgan. 
Moreover, with a per-class stratification of the results, we have discovered a correlation between per-class performance and generated per-class image quality. These findings hint at two future directions to improve the effect of \ours: increasing generation quality for the classes which are poorly modeled by \allicgan and to tune the \allicgan augmentations per class. 
Finally, in the case of more aggressive data augmentation techniques for which \ours does not provide an improvement over the baselines, such as CutMixUp or multi-crop, we hypothesize that those augmentation recipes already result in strong image variations, and consequently, combining those with \allicgan generations out-of-the-box through \ours may cause an over-regularization of the training.

To conclude, we have shown that current state-of-the-art large capacity models can be improved using instance conditioned generative models such as \icgan in conjunction with hand-crafted data augmentation techniques. 
We further hypothesize that by boosting the quality and diversity of instance conditioned samples, models may eventually stop relying on hand-crafted data augmentation techniques altogether, and instead move towards completely data-driven augmentation schemes to obtain infinitely many realistic augmented samples.\looseness-1

\bibliography{tmlr}

\begin{thebibliography}{75}
\providecommand{\natexlab}[1]{#1}
\providecommand{\url}[1]{\texttt{#1}}
\expandafter\ifx\csname urlstyle\endcsname\relax
  \providecommand{\doi}[1]{doi: #1}\else
  \providecommand{\doi}{doi: \begingroup \urlstyle{rm}\Url}\fi

\bibitem[Alayrac et~al.(2022)Alayrac, Donahue, Luc, Miech, Barr, Hasson, Lenc,
  Mensch, Millican, and Reynolds]{alayrac_flamingo_2022}
Jean-Baptiste Alayrac, Jeff Donahue, Pauline Luc, Antoine Miech, Iain Barr,
  Yana Hasson, Karel Lenc, Arthur Mensch, Katie Millican, and Malcolm Reynolds.
\newblock Flamingo: A visual language model for few-shot learning.
\newblock \emph{arXiv preprint arXiv:2204.14198}, 2022.

\bibitem[Antoniou et~al.(2018)Antoniou, Storkey, and
  Edwards]{antoniou_data_2018}
Antreas Antoniou, Amos Storkey, and Harrison Edwards.
\newblock Data {{Augmentation Generative Adversarial Networks}}.
\newblock \emph{arXiv preprint arXiv:1711.04340}, 2018.

\bibitem[Barbu et~al.(2019)Barbu, Mayo, Alverio, Luo, Wang, Gutfreund,
  Tenenbaum, and Katz]{barbu_objectnet_2019}
Andrei Barbu, David Mayo, Julian Alverio, William Luo, Christopher Wang, Dan
  Gutfreund, Josh Tenenbaum, and Boris Katz.
\newblock {{ObjectNet}}: {{A}} large-scale bias-controlled dataset for pushing
  the limits of object recognition models.
\newblock In \emph{Advances in {{Neural Information Processing Systems}}},
  volume~32. {Curran Associates, Inc.}, 2019.

\bibitem[Besnier et~al.(2020)Besnier, Jain, Bursuc, Cord, and
  P{\'e}rez]{besnier_this_2020}
Victor Besnier, Himalaya Jain, Andrei Bursuc, Matthieu Cord, and Patrick
  P{\'e}rez.
\newblock This {{Dataset Does Not Exist}}: {{Training Models}} from {{Generated
  Images}}.
\newblock In \emph{{{ICASSP}} 2020 - 2020 {{IEEE International Conference}} on
  {{Acoustics}}, {{Speech}} and {{Signal Processing}} ({{ICASSP}})}, pp.\
  1--5, May 2020.
\newblock \doi{10.1109/ICASSP40776.2020.9053146}.

\bibitem[Beyer et~al.(2020)Beyer, H{\'e}naff, Kolesnikov, Zhai, and van~den
  Oord]{beyer_are_2020}
Lucas Beyer, Olivier~J. H{\'e}naff, Alexander Kolesnikov, Xiaohua Zhai, and
  A{\"a}ron van~den Oord.
\newblock Are we done with {{ImageNet}}?, June 2020.

\bibitem[Bowles et~al.(2018)Bowles, Chen, Guerrero, Bentley, Gunn, Hammers,
  Dickie, Hern{\'a}ndez, Wardlaw, and Rueckert]{bowles_gan_2018}
Christopher Bowles, Liang Chen, Ricardo Guerrero, Paul Bentley, Roger Gunn,
  Alexander Hammers, David~Alexander Dickie, Maria~Vald{\'e}s Hern{\'a}ndez,
  Joanna Wardlaw, and Daniel Rueckert.
\newblock Gan augmentation: {{Augmenting}} training data using generative
  adversarial networks.
\newblock \emph{arXiv preprint arXiv:1810.10863}, 2018.

\bibitem[Brock et~al.(2019)Brock, Donahue, and Simonyan]{brock_large_2019}
Andrew Brock, Jeff Donahue, and Karen Simonyan.
\newblock Large {{Scale GAN Training}} for {{High Fidelity Natural Image
  Synthesis}}.
\newblock In \emph{International {{Conference}} on {{Learning
  Representations}}}, 2019.

\bibitem[Caron et~al.(2020)Caron, Misra, Mairal, Goyal, Bojanowski, and
  Joulin]{caron_unsupervised_2020}
Mathilde Caron, Ishan Misra, Julien Mairal, Priya Goyal, Piotr Bojanowski, and
  Armand Joulin.
\newblock Unsupervised learning of visual features by contrasting cluster
  assignments.
\newblock \emph{arXiv preprint arXiv:2006.09882}, 2020.

\bibitem[Caron et~al.(2021)Caron, Touvron, Misra, J{\'e}gou, Mairal,
  Bojanowski, and Joulin]{caron_emerging_2021}
Mathilde Caron, Hugo Touvron, Ishan Misra, Herv{\'e} J{\'e}gou, Julien Mairal,
  Piotr Bojanowski, and Armand Joulin.
\newblock Emerging properties in self-supervised vision transformers.
\newblock \emph{arXiv preprint arXiv:2104.14294}, 2021.

\bibitem[Casanova et~al.(2021)Casanova, Careil, Verbeek, Drozdzal, and
  Romero]{casanova_instance-conditioned_2021}
Arantxa Casanova, Marl{\`e}ne Careil, Jakob Verbeek, Michal Drozdzal, and
  Adriana Romero.
\newblock Instance-{{Conditioned GAN}}.
\newblock In \emph{Thirty-{{Fifth Conference}} on {{Neural Information
  Processing Systems}}}, 2021.

\bibitem[Chen et~al.(2020{\natexlab{a}})Chen, Kornblith, Norouzi, and
  Hinton]{chen_simple_2020}
Ting Chen, Simon Kornblith, Mohammad Norouzi, and Geoffrey Hinton.
\newblock A simple framework for contrastive learning of visual
  representations.
\newblock In \emph{International Conference on Machine Learning}, pp.\
  1597--1607. {PMLR}, 2020{\natexlab{a}}.

\bibitem[Chen et~al.(2020{\natexlab{b}})Chen, Kornblith, Swersky, Norouzi, and
  Hinton]{chen_big_2020}
Ting Chen, Simon Kornblith, Kevin Swersky, Mohammad Norouzi, and Geoffrey~E
  Hinton.
\newblock Big {{Self-Supervised Models}} are {{Strong Semi-Supervised
  Learners}}.
\newblock In \emph{Advances in {{Neural Information Processing Systems}}},
  volume~33, pp.\  22243--22255. {Curran Associates, Inc.}, 2020{\natexlab{b}}.

\bibitem[Chen et~al.(2021)Chen, Xie, and He]{chen_empirical_2021}
Xinlei Chen, Saining Xie, and Kaiming He.
\newblock An {{Empirical Study}} of {{Training Self-Supervised Vision
  Transformers}}.
\newblock In \emph{Proceedings of the {{IEEE}}/{{CVF International Conference}}
  on {{Computer Vision}}}, pp.\  9640--9649, 2021.

\bibitem[Choi et~al.(2019)Choi, Kim, and Kim]{choi_self-ensembling_2019}
Jaehoon Choi, Taekyung Kim, and Changick Kim.
\newblock Self-{{Ensembling With GAN-Based Data Augmentation}} for {{Domain
  Adaptation}} in {{Semantic Segmentation}}.
\newblock In \emph{Proceedings of the {{IEEE}}/{{CVF International Conference}}
  on {{Computer Vision}}}, pp.\  6830--6840, 2019.

\bibitem[Cubuk et~al.(2019)Cubuk, Zoph, Mane, Vasudevan, and
  Le]{cubuk_autoaugment_2019}
Ekin~D. Cubuk, Barret Zoph, Dandelion Mane, Vijay Vasudevan, and Quoc~V. Le.
\newblock {{AutoAugment}}: {{Learning Augmentation Strategies From Data}}.
\newblock In \emph{Proceedings of the {{IEEE}}/{{CVF Conference}} on {{Computer
  Vision}} and {{Pattern Recognition}}}, pp.\  113--123, 2019.

\bibitem[Cubuk et~al.(2020)Cubuk, Zoph, Shlens, and
  Le]{cubuk_randaugment_2020-2}
Ekin~D. Cubuk, Barret Zoph, Jonathon Shlens, and Quoc~V. Le.
\newblock Randaugment: {{Practical Automated Data Augmentation With}} a
  {{Reduced Search Space}}.
\newblock In \emph{Proceedings of the {{IEEE}}/{{CVF Conference}} on {{Computer
  Vision}} and {{Pattern Recognition Workshops}}}, pp.\  702--703, 2020.

\bibitem[Deng et~al.(2009)Deng, Dong, Socher, Li, Li, and Fei-Fei]{deng09cvpr}
J.~Deng, W.~Dong, R.~Socher, L.-J. Li, K.~Li, and L.~Fei-Fei.
\newblock {ImageNet}: A large-scale hierarchical image database.
\newblock In \emph{CVPR}, 2009.

\bibitem[DeVries \& Taylor(2017{\natexlab{a}})DeVries and
  Taylor]{devries_dataset_2017}
Terrance DeVries and Graham~W. Taylor.
\newblock Dataset augmentation in feature space.
\newblock \emph{arXiv preprint arXiv:1702.05538}, 2017{\natexlab{a}}.

\bibitem[DeVries \& Taylor(2017{\natexlab{b}})DeVries and
  Taylor]{devries_improved_2017}
Terrance DeVries and Graham~W. Taylor.
\newblock Improved {{Regularization}} of {{Convolutional Neural Networks}} with
  {{Cutout}}.
\newblock \emph{arXiv:1708.04552 [cs]}, November 2017{\natexlab{b}}.

\bibitem[Dosovitskiy et~al.(2021)Dosovitskiy, Beyer, Kolesnikov, Weissenborn,
  Zhai, Unterthiner, Dehghani, Minderer, Heigold, Gelly, Uszkoreit, and
  Houlsby]{dosovitskiy_image_2021}
Alexey Dosovitskiy, Lucas Beyer, Alexander Kolesnikov, Dirk Weissenborn,
  Xiaohua Zhai, Thomas Unterthiner, Mostafa Dehghani, Matthias Minderer, Georg
  Heigold, Sylvain Gelly, Jakob Uszkoreit, and Neil Houlsby.
\newblock An {{Image}} is {{Worth}} 16x16 {{Words}}: {{Transformers}} for
  {{Image Recognition}} at {{Scale}}.
\newblock In \emph{International {{Conference}} on {{Learning
  Representations}}}, 2021.

\bibitem[Dwibedi et~al.(2021)Dwibedi, Aytar, Tompson, Sermanet, and
  Zisserman]{dwibedi_little_2021-1}
Debidatta Dwibedi, Yusuf Aytar, Jonathan Tompson, Pierre Sermanet, and Andrew
  Zisserman.
\newblock With a {{Little Help From My Friends}}: {{Nearest-Neighbor
  Contrastive Learning}} of {{Visual Representations}}.
\newblock In \emph{Proceedings of the {{IEEE}}/{{CVF International Conference}}
  on {{Computer Vision}}}, pp.\  9588--9597, 2021.

\bibitem[{Frid-Adar} et~al.(2018){Frid-Adar}, Diamant, Klang, Amitai,
  Goldberger, and Greenspan]{frid-adar_gan-based_2018}
Maayan {Frid-Adar}, Idit Diamant, Eyal Klang, Michal Amitai, Jacob Goldberger,
  and Hayit Greenspan.
\newblock {{GAN-based}} synthetic medical image augmentation for increased
  {{CNN}} performance in liver lesion classification.
\newblock \emph{Neurocomputing}, 321:\penalty0 321--331, December 2018.
\newblock ISSN 0925-2312.
\newblock \doi{10.1016/j.neucom.2018.09.013}.

\bibitem[Gafni et~al.(2022)Gafni, Polyak, Ashual, Sheynin, Parikh, and
  Taigman]{gafni22arxiv}
Oran Gafni, Adam Polyak, Oron Ashual, Shelly Sheynin, Devi Parikh, and Yaniv
  Taigman.
\newblock Make-a-scene: Scene-based text-to-image generation with human priors.
\newblock \emph{arXiv preprint}, 2022.

\bibitem[Gao et~al.(2018)Gao, Shou, Zareian, Zhang, and
  Chang]{gao_low-shot_2018}
Hang Gao, Zheng Shou, Alireza Zareian, Hanwang Zhang, and Shih-Fu Chang.
\newblock Low-shot {{Learning}} via {{Covariance-Preserving Adversarial
  Augmentation Networks}}.
\newblock In \emph{Advances in {{Neural Information Processing Systems}}},
  volume~31. {Curran Associates, Inc.}, 2018.

\bibitem[Geusebroek et~al.(2005)Geusebroek, Burghouts, and
  Smeulders]{geusebroek_amsterdam_2005}
Jan-Mark Geusebroek, Gertjan~J. Burghouts, and Arnold~WM Smeulders.
\newblock The {{Amsterdam}} library of object images.
\newblock \emph{International Journal of Computer Vision}, 61\penalty0
  (1):\penalty0 103--112, 2005.

\bibitem[Goodfellow et~al.(2014)Goodfellow, {Pouget-Abadie}, Mirza, Xu,
  {Warde-Farley}, Ozair, Courville, and Bengio]{goodfellow_generative_2014}
Ian Goodfellow, Jean {Pouget-Abadie}, Mehdi Mirza, Bing Xu, David
  {Warde-Farley}, Sherjil Ozair, Aaron Courville, and Yoshua Bengio.
\newblock Generative {{Adversarial Nets}}.
\newblock In \emph{Advances in {{Neural Information Processing Systems}}},
  volume~27. {Curran Associates, Inc.}, 2014.

\bibitem[Goyal et~al.(2021)Goyal, Duval, Reizenstein, Leavitt, Xu, Lefaudeux,
  Singh, Reis, Caron, Bojanowski, Joulin, and Misra]{goyal2021vissl}
Priya Goyal, Quentin Duval, Jeremy Reizenstein, Matthew Leavitt, Min Xu,
  Benjamin Lefaudeux, Mannat Singh, Vinicius Reis, Mathilde Caron, Piotr
  Bojanowski, Armand Joulin, and Ishan Misra.
\newblock Vissl.
\newblock \url{https://github.com/facebookresearch/vissl}, 2021.

\bibitem[Goyal et~al.(2022)Goyal, Duval, Seessel, Caron, Misra, Sagun, Joulin,
  and Bojanowski]{goyal_vision_2022}
Priya Goyal, Quentin Duval, Isaac Seessel, Mathilde Caron, Ishan Misra, Levent
  Sagun, Armand Joulin, and Piotr Bojanowski.
\newblock Vision {{Models Are More Robust And Fair When Pretrained On Uncurated
  Images Without Supervision}}, February 2022.

\bibitem[Grill et~al.(2020)Grill, Strub, Altch{\'e}, Tallec, Richemond,
  Buchatskaya, Doersch, Avila~Pires, Guo, Gheshlaghi~Azar, Piot, {kavukcuoglu},
  Munos, and Valko]{grill_bootstrap_2020-1}
Jean-Bastien Grill, Florian Strub, Florent Altch{\'e}, Corentin Tallec, Pierre
  Richemond, Elena Buchatskaya, Carl Doersch, Bernardo Avila~Pires, Zhaohan
  Guo, Mohammad Gheshlaghi~Azar, Bilal Piot, koray {kavukcuoglu}, Remi Munos,
  and Michal Valko.
\newblock Bootstrap {{Your Own Latent}} - {{A New Approach}} to
  {{Self-Supervised Learning}}.
\newblock In \emph{Advances in {{Neural Information Processing Systems}}},
  volume~33, pp.\  21271--21284. {Curran Associates, Inc.}, 2020.

\bibitem[H{\"a}rk{\"o}nen et~al.(2020)H{\"a}rk{\"o}nen, Hertzmann, Lehtinen,
  and Paris]{harkonen_ganspace_2020}
Erik H{\"a}rk{\"o}nen, Aaron Hertzmann, Jaakko Lehtinen, and Sylvain Paris.
\newblock {{GANSpace}}: {{Discovering Interpretable GAN Controls}}.
\newblock In \emph{Advances in {{Neural Information Processing Systems}}},
  volume~33, pp.\  9841--9850. {Curran Associates, Inc.}, 2020.

\bibitem[He et~al.(2016)He, Zhang, Ren, and Sun]{he_deep_2016}
Kaiming He, Xiangyu Zhang, Shaoqing Ren, and Jian Sun.
\newblock Deep {{Residual Learning}} for {{Image Recognition}}.
\newblock In \emph{Proceedings of the {{IEEE Conference}} on {{Computer
  Vision}} and {{Pattern Recognition}}}, pp.\  770--778, 2016.

\bibitem[He et~al.(2020)He, Fan, Wu, Xie, and Girshick]{he_momentum_2020}
Kaiming He, Haoqi Fan, Yuxin Wu, Saining Xie, and Ross Girshick.
\newblock Momentum contrast for unsupervised visual representation learning.
\newblock In \emph{Proceedings of the {{IEEE}}/{{CVF Conference}} on {{Computer
  Vision}} and {{Pattern Recognition}}}, pp.\  9729--9738, 2020.

\bibitem[Hendrycks et~al.(2021{\natexlab{a}})Hendrycks, Basart, Mu, Kadavath,
  Wang, Dorundo, Desai, Zhu, Parajuli, Guo, Song, Steinhardt, and
  Gilmer]{hendrycks_many_2021}
Dan Hendrycks, Steven Basart, Norman Mu, Saurav Kadavath, Frank Wang, Evan
  Dorundo, Rahul Desai, Tyler Zhu, Samyak Parajuli, Mike Guo, Dawn Song, Jacob
  Steinhardt, and Justin Gilmer.
\newblock The {{Many Faces}} of {{Robustness}}: {{A Critical Analysis}} of
  {{Out-of-Distribution Generalization}}.
\newblock In \emph{Proceedings of the {{IEEE}}/{{CVF International Conference}}
  on {{Computer Vision}}}, pp.\  8340--8349, 2021{\natexlab{a}}.

\bibitem[Hendrycks et~al.(2021{\natexlab{b}})Hendrycks, Zhao, Basart,
  Steinhardt, and Song]{hendrycks_natural_2021}
Dan Hendrycks, Kevin Zhao, Steven Basart, Jacob Steinhardt, and Dawn Song.
\newblock Natural adversarial examples.
\newblock In \emph{Proceedings of the {{IEEE}}/{{CVF Conference}} on {{Computer
  Vision}} and {{Pattern Recognition}}}, pp.\  15262--15271,
  2021{\natexlab{b}}.

\bibitem[Heusel et~al.(2017)Heusel, Ramsauer, Unterthiner, Nessler, and
  Hochreiter]{heusel17nips}
Martin Heusel, Hubert Ramsauer, Thomas Unterthiner, Bernhard Nessler, and Sepp
  Hochreiter.
\newblock {GAN}s trained by a two time-scale update rule converge to a local
  {N}ash equilibrium.
\newblock In \emph{NeurIPS}, 2017.

\bibitem[Huang et~al.(2019)Huang, Zhao, and Huang]{huang_got-10k_2019}
Lianghua Huang, Xin Zhao, and Kaiqi Huang.
\newblock Got-10k: {{A}} large high-diversity benchmark for generic object
  tracking in the wild.
\newblock \emph{IEEE Transactions on Pattern Analysis and Machine
  Intelligence}, 43\penalty0 (5):\penalty0 1562--1577, 2019.

\bibitem[Huang et~al.(2018)Huang, Lin, Chen, Wu, Hsu, and
  Lai]{huang_auggan_2018}
Sheng-Wei Huang, Che-Tsung Lin, Shu-Ping Chen, Yen-Yi Wu, Po-Hao Hsu, and
  Shang-Hong Lai.
\newblock {{AugGAN}}: {{Cross Domain Adaptation}} with {{GAN-based Data
  Augmentation}}.
\newblock In \emph{Proceedings of the {{European Conference}} on {{Computer
  Vision}} ({{ECCV}})}, pp.\  718--731, 2018.

\bibitem[Huh et~al.(2020)Huh, Zhang, Zhu, Paris, and
  Hertzmann]{huh_transforming_2020}
Minyoung Huh, Richard Zhang, Jun-Yan Zhu, Sylvain Paris, and Aaron Hertzmann.
\newblock Transforming and {{Projecting Images}} into {{Class-Conditional
  Generative Networks}}.
\newblock In Andrea Vedaldi, Horst Bischof, Thomas Brox, and Jan-Michael Frahm
  (eds.), \emph{Computer {{Vision}} \textendash{} {{ECCV}} 2020}, Lecture
  {{Notes}} in {{Computer Science}}, pp.\  17--34, {Cham}, 2020. {Springer
  International Publishing}.
\newblock ISBN 978-3-030-58536-5.
\newblock \doi{10.1007/978-3-030-58536-5_2}.

\bibitem[Jahanian et~al.(2020)Jahanian, Chai, and
  Isola]{jahanian_steerability_2020}
Ali Jahanian, Lucy Chai, and Phillip Isola.
\newblock On the "steerability" of generative adversarial networks.
\newblock In \emph{International {{Conference}} on {{Learning
  Representations}}}, 2020.

\bibitem[Jahanian et~al.(2022)Jahanian, Puig, Tian, and
  Isola]{jahanian_generative_2022}
Ali Jahanian, Xavier Puig, Yonglong Tian, and Phillip Isola.
\newblock Generative {{Models}} as a {{Data Source}} for {{Multiview
  Representation Learning}}.
\newblock In \emph{International {{Conference}} on {{Learning
  Representations}}}, 2022.

\bibitem[Kolesnikov et~al.(2020)Kolesnikov, Beyer, Zhai, Puigcerver, Yung,
  Gelly, and Houlsby]{kolesnikov_big_2020}
Alexander Kolesnikov, Lucas Beyer, Xiaohua Zhai, Joan Puigcerver, Jessica Yung,
  Sylvain Gelly, and Neil Houlsby.
\newblock Big {{Transfer}} ({{BiT}}): {{General Visual Representation
  Learning}}.
\newblock In Andrea Vedaldi, Horst Bischof, Thomas Brox, and Jan-Michael Frahm
  (eds.), \emph{Computer {{Vision}} \textendash{} {{ECCV}} 2020}, Lecture
  {{Notes}} in {{Computer Science}}, pp.\  491--507, {Cham}, 2020. {Springer
  International Publishing}.
\newblock ISBN 978-3-030-58558-7.
\newblock \doi{10.1007/978-3-030-58558-7_29}.

\bibitem[Krizhevsky et~al.(2012)Krizhevsky, Sutskever, and
  Hinton]{krizhevsky_imagenet_2012}
Alex Krizhevsky, Ilya Sutskever, and Geoffrey~E Hinton.
\newblock {{ImageNet Classification}} with {{Deep Convolutional Neural
  Networks}}.
\newblock In \emph{Advances in {{Neural Information Processing Systems}}},
  volume~25. {Curran Associates, Inc.}, 2012.

\bibitem[Lemley et~al.(2017)Lemley, Bazrafkan, and Corcoran]{lemley_smart_2017}
Joseph Lemley, Shabab Bazrafkan, and Peter Corcoran.
\newblock Smart {{Augmentation Learning}} an {{Optimal Data Augmentation
  Strategy}}.
\newblock \emph{IEEE Access}, 5:\penalty0 5858--5869, 2017.
\newblock ISSN 2169-3536.
\newblock \doi{10.1109/ACCESS.2017.2696121}.

\bibitem[Li et~al.(2021)Li, Yang, Kreis, Torralba, and
  Fidler]{li_semantic_2021}
Daiqing Li, Junlin Yang, Karsten Kreis, Antonio Torralba, and Sanja Fidler.
\newblock Semantic {{Segmentation With Generative Models}}: {{Semi-Supervised
  Learning}} and {{Strong Out-of-Domain Generalization}}.
\newblock In \emph{Proceedings of the {{IEEE}}/{{CVF Conference}} on {{Computer
  Vision}} and {{Pattern Recognition}}}, pp.\  8300--8311, 2021.

\bibitem[Li et~al.(2022)Li, Ling, Kim, Kreis, Barriuso, Fidler, and
  Torralba]{li_bigdatasetgan_2022}
Daiqing Li, Huan Ling, Seung~Wook Kim, Karsten Kreis, Adela Barriuso, Sanja
  Fidler, and Antonio Torralba.
\newblock {{BigDatasetGAN}}: {{Synthesizing ImageNet}} with {{Pixel-wise
  Annotations}}.
\newblock \emph{arXiv:2201.04684 [cs]}, January 2022.

\bibitem[Liu et~al.(2018)Liu, Zou, Kong, Diao, Yan, Wang, Li, Jia, and
  You]{liu_data_2018}
Xiaofeng Liu, Yang Zou, Lingsheng Kong, Zhihui Diao, Junliang Yan, Jun Wang,
  Site Li, Ping Jia, and Jane You.
\newblock Data {{Augmentation}} via {{Latent Space Interpolation}} for {{Image
  Classification}}.
\newblock In \emph{2018 24th {{International Conference}} on {{Pattern
  Recognition}} ({{ICPR}})}, pp.\  728--733, August 2018.
\newblock \doi{10.1109/ICPR.2018.8545506}.

\bibitem[Mao et~al.(2021)Mao, Cha, Gupta, Wang, Yang, and
  Vondrick]{mao_generative_2021}
Chengzhi Mao, Augustine Cha, Amogh Gupta, Hao Wang, Junfeng Yang, and Carl
  Vondrick.
\newblock Generative {{Interventions}} for {{Causal Learning}}.
\newblock In \emph{Proceedings of the {{IEEE}}/{{CVF Conference}} on {{Computer
  Vision}} and {{Pattern Recognition}}}, pp.\  3947--3956, 2021.

\bibitem[Marchesi(2017)]{marchesi_megapixel_2017}
Marco Marchesi.
\newblock Megapixel size image creation using generative adversarial networks.
\newblock \emph{arXiv preprint arXiv:1706.00082}, 2017.

\bibitem[Perez \& Wang(2017)Perez and Wang]{perez_effectiveness_2017}
Luis Perez and Jason Wang.
\newblock The {{Effectiveness}} of {{Data Augmentation}} in {{Image
  Classification}} using {{Deep Learning}}.
\newblock \emph{arXiv preprint arXiv:1712.04621}, 2017.

\bibitem[Pesteie et~al.(2019)Pesteie, Abolmaesumi, and
  Rohling]{pesteie_adaptive_2019}
Mehran Pesteie, Purang Abolmaesumi, and Robert~N. Rohling.
\newblock Adaptive {{Augmentation}} of {{Medical Data Using Independently
  Conditional Variational Auto-Encoders}}.
\newblock \emph{IEEE Transactions on Medical Imaging}, 38\penalty0
  (12):\penalty0 2807--2820, December 2019.
\newblock ISSN 1558-254X.
\newblock \doi{10.1109/TMI.2019.2914656}.

\bibitem[Purushwalkam \& Gupta(2020)Purushwalkam and
  Gupta]{purushwalkam_demystifying_2020-1}
Senthil Purushwalkam and Abhinav Gupta.
\newblock Demystifying contrastive self-supervised learning: {{Invariances}},
  augmentations and dataset biases.
\newblock In H.~Larochelle, M.~Ranzato, R.~Hadsell, M.F. Balcan, and H.~Lin
  (eds.), \emph{Advances in Neural Information Processing Systems}, volume~33,
  pp.\  3407--3418. {Curran Associates, Inc.}, 2020.

\bibitem[Radford et~al.(2021)Radford, Kim, Hallacy, Ramesh, Goh, Agarwal,
  Sastry, Askell, Mishkin, Clark, Krueger, and
  Sutskever]{radford_learning_2021}
Alec Radford, Jong~Wook Kim, Chris Hallacy, Aditya Ramesh, Gabriel Goh,
  Sandhini Agarwal, Girish Sastry, Amanda Askell, Pamela Mishkin, Jack Clark,
  Gretchen Krueger, and Ilya Sutskever.
\newblock Learning {{Transferable Visual Models From Natural Language
  Supervision}}.
\newblock In \emph{Proceedings of the 38th {{International Conference}} on
  {{Machine Learning}}}, pp.\  8748--8763. {PMLR}, July 2021.

\bibitem[Ramesh et~al.(2022)Ramesh, Dhariwal, Nichol, Chu, and
  Chen]{ramesh_hierarchical_2022}
Aditya Ramesh, Prafulla Dhariwal, Alex Nichol, Casey Chu, and Mark Chen.
\newblock Hierarchical {{Text-Conditional Image Generation}} with {{CLIP
  Latents}}, April 2022.

\bibitem[Ravuri \& Vinyals(2019)Ravuri and Vinyals]{ravuri_classification_2019}
Suman Ravuri and Oriol Vinyals.
\newblock Classification {{Accuracy Score}} for {{Conditional Generative
  Models}}.
\newblock In \emph{Advances in {{Neural Information Processing Systems}}},
  volume~32. {Curran Associates, Inc.}, 2019.

\bibitem[Saharia et~al.(2022)Saharia, Chan, Saxena, Li, Whang, Denton,
  Ghasemipour, Ayan, Mahdavi, Lopes, Salimans, Ho, Fleet, and
  Norouzi]{saharia_photorealistic_2022}
Chitwan Saharia, William Chan, Saurabh Saxena, Lala Li, Jay Whang, Emily
  Denton, Seyed Kamyar~Seyed Ghasemipour, Burcu~Karagol Ayan, S.~Sara Mahdavi,
  Rapha~Gontijo Lopes, Tim Salimans, Jonathan Ho, David~J. Fleet, and Mohammad
  Norouzi.
\newblock Photorealistic {{Text-to-Image Diffusion Models}} with {{Deep
  Language Understanding}}, May 2022.

\bibitem[Sandfort et~al.(2019)Sandfort, Yan, Pickhardt, and
  Summers]{sandfort_data_2019}
Veit Sandfort, Ke~Yan, Perry~J. Pickhardt, and Ronald~M. Summers.
\newblock Data augmentation using generative adversarial networks
  ({{CycleGAN}}) to improve generalizability in {{CT}} segmentation tasks.
\newblock \emph{Scientific Reports}, 9\penalty0 (1):\penalty0 16884, November
  2019.
\newblock ISSN 2045-2322.
\newblock \doi{10.1038/s41598-019-52737-x}.

\bibitem[Schwartz et~al.(2018)Schwartz, Karlinsky, Shtok, Harary, Marder,
  Kumar, Feris, Giryes, and Bronstein]{schwartz_delta-encoder_2018}
Eli Schwartz, Leonid Karlinsky, Joseph Shtok, Sivan Harary, Mattias Marder,
  Abhishek Kumar, Rogerio Feris, Raja Giryes, and Alex Bronstein.
\newblock Delta-encoder: An effective sample synthesis method for few-shot
  object recognition.
\newblock In \emph{Advances in {{Neural Information Processing Systems}}},
  volume~31. {Curran Associates, Inc.}, 2018.

\bibitem[Shorten \& Khoshgoftaar(2019)Shorten and
  Khoshgoftaar]{shorten_survey_2019}
Connor Shorten and Taghi~M. Khoshgoftaar.
\newblock A survey on {{Image Data Augmentation}} for {{Deep Learning}}.
\newblock \emph{Journal of Big Data}, 6\penalty0 (1):\penalty0 60, July 2019.
\newblock ISSN 2196-1115.
\newblock \doi{10.1186/s40537-019-0197-0}.

\bibitem[Shrivastava et~al.(2017)Shrivastava, Pfister, Tuzel, Susskind, Wang,
  and Webb]{shrivastava_learning_2017}
Ashish Shrivastava, Tomas Pfister, Oncel Tuzel, Joshua Susskind, Wenda Wang,
  and Russell Webb.
\newblock Learning {{From Simulated}} and {{Unsupervised Images Through
  Adversarial Training}}.
\newblock In \emph{Proceedings of the {{IEEE Conference}} on {{Computer
  Vision}} and {{Pattern Recognition}}}, pp.\  2107--2116, 2017.

\bibitem[Steiner et~al.(2021)Steiner, Kolesnikov, Zhai, Wightman, Uszkoreit,
  and Beyer]{steiner_how_2021}
Andreas Steiner, Alexander Kolesnikov, Xiaohua Zhai, Ross Wightman, Jakob
  Uszkoreit, and Lucas Beyer.
\newblock How to train your vit? data, augmentation, and regularization in
  vision transformers.
\newblock \emph{arXiv preprint arXiv:2106.10270}, 2021.

\bibitem[Touvron et~al.(2021)Touvron, Cord, Douze, Massa, Sablayrolles, and
  Jegou]{touvron_training_2021}
Hugo Touvron, Matthieu Cord, Matthijs Douze, Francisco Massa, Alexandre
  Sablayrolles, and Herve Jegou.
\newblock Training data-efficient image transformers {$\&$} distillation
  through attention.
\newblock In \emph{Proceedings of the 38th {{International Conference}} on
  {{Machine Learning}}}, pp.\  10347--10357. {PMLR}, July 2021.

\bibitem[Touvron et~al.(2022)Touvron, Cord, and J{\'e}gou]{touvron_deit_2022}
Hugo Touvron, Matthieu Cord, and Herv{\'e} J{\'e}gou.
\newblock {{DeiT III}}: {{Revenge}} of the {{ViT}}, April 2022.

\bibitem[Tritrong et~al.(2021)Tritrong, Rewatbowornwong, and
  Suwajanakorn]{tritrong_repurposing_2021}
Nontawat Tritrong, Pitchaporn Rewatbowornwong, and Supasorn Suwajanakorn.
\newblock Repurposing {{GANs}} for {{One-Shot Semantic Part Segmentation}}.
\newblock In \emph{Proceedings of the {{IEEE}}/{{CVF Conference}} on {{Computer
  Vision}} and {{Pattern Recognition}}}, pp.\  4475--4485, 2021.

\bibitem[Wang et~al.(2019)Wang, Pan, Song, Zhang, Huang, and
  Wu]{wang_implicit_2019}
Yulin Wang, Xuran Pan, Shiji Song, Hong Zhang, Gao Huang, and Cheng Wu.
\newblock Implicit {{Semantic Data Augmentation}} for {{Deep Networks}}.
\newblock In \emph{Advances in {{Neural Information Processing Systems}}},
  volume~32. {Curran Associates, Inc.}, 2019.

\bibitem[Wang et~al.(2021)Wang, Huang, Song, Pan, Xia, and
  Wu]{wang_regularizing_2021}
Yulin Wang, Gao Huang, Shiji Song, Xuran Pan, Yitong Xia, and Cheng Wu.
\newblock Regularizing {{Deep Networks}} with {{Semantic Data Augmentation}}.
\newblock \emph{IEEE Transactions on Pattern Analysis and Machine
  Intelligence}, pp.\  1--1, 2021.
\newblock ISSN 1939-3539.
\newblock \doi{10.1109/TPAMI.2021.3052951}.

\bibitem[Xia et~al.(2022)Xia, Zhang, Yang, Xue, Zhou, and Yang]{xia_gan_2022}
Weihao Xia, Yulun Zhang, Yujiu Yang, Jing-Hao Xue, Bolei Zhou, and Ming-Hsuan
  Yang.
\newblock {{GAN Inversion}}: {{A Survey}}, March 2022.

\bibitem[Xiang et~al.(2014)Xiang, Mottaghi, and Savarese]{xiang_beyond_2014}
Yu~Xiang, Roozbeh Mottaghi, and Silvio Savarese.
\newblock Beyond pascal: {{A}} benchmark for 3d object detection in the wild.
\newblock In \emph{{{IEEE}} Winter Conference on Applications of Computer
  Vision}, pp.\  75--82. {IEEE}, 2014.

\bibitem[Yun et~al.(2019)Yun, Han, Oh, Chun, Choe, and Yoo]{yun_cutmix_2019}
Sangdoo Yun, Dongyoon Han, Seong~Joon Oh, Sanghyuk Chun, Junsuk Choe, and
  Youngjoon Yoo.
\newblock {{CutMix}}: {{Regularization Strategy}} to {{Train Strong Classifiers
  With Localizable Features}}.
\newblock In \emph{Proceedings of the {{IEEE}}/{{CVF International Conference}}
  on {{Computer Vision}}}, pp.\  6023--6032, 2019.

\bibitem[Zagoruyko \& Komodakis(2016)Zagoruyko and
  Komodakis]{zagoruyko_wide_2016-1}
Sergey Zagoruyko and Nikos Komodakis.
\newblock Wide {{Residual Networks}}.
\newblock In \emph{Procedings of the {{British Machine Vision Conference}}
  2016}, pp.\  87.1--87.12, {York, UK}, 2016. {British Machine Vision
  Association}.
\newblock ISBN 978-1-901725-59-9.
\newblock \doi{10.5244/C.30.87}.

\bibitem[Zhang et~al.(2017)Zhang, Cisse, Dauphin, and
  {Lopez-Paz}]{zhang_mixup_2017}
Hongyi Zhang, Moustapha Cisse, Yann~N. Dauphin, and David {Lopez-Paz}.
\newblock Mixup: {{Beyond Empirical Risk Minimization}}.
\newblock In \emph{International {{Conference}} on {{Learning
  Representations}}}, 2017.

\bibitem[Zhang et~al.(2022)Zhang, Roller, Goyal, Artetxe, Chen, Chen, Dewan,
  Diab, Li, Lin, Mihaylov, Ott, Shleifer, Shuster, Simig, Koura, Sridhar, Wang,
  and Zettlemoyer]{zhang_opt_2022}
Susan Zhang, Stephen Roller, Naman Goyal, Mikel Artetxe, Moya Chen, Shuohui
  Chen, Christopher Dewan, Mona Diab, Xian Li, Xi~Victoria Lin, Todor Mihaylov,
  Myle Ott, Sam Shleifer, Kurt Shuster, Daniel Simig, Punit~Singh Koura, Anjali
  Sridhar, Tianlu Wang, and Luke Zettlemoyer.
\newblock {{OPT}}: {{Open Pre-trained Transformer Language Models}}, May 2022.

\bibitem[Zhang et~al.(2021)Zhang, Ling, Gao, Yin, Lafleche, Barriuso, Torralba,
  and Fidler]{zhang_datasetgan_2021-2}
Yuxuan Zhang, Huan Ling, Jun Gao, Kangxue Yin, Jean-Francois Lafleche, Adela
  Barriuso, Antonio Torralba, and Sanja Fidler.
\newblock {{DatasetGAN}}: {{Efficient Labeled Data Factory With Minimal Human
  Effort}}.
\newblock In \emph{Proceedings of the {{IEEE}}/{{CVF Conference}} on {{Computer
  Vision}} and {{Pattern Recognition}}}, pp.\  10145--10155, 2021.

\bibitem[Zhao \& Bilen(2022)Zhao and Bilen]{zhao_synthesizing_2022}
Bo~Zhao and Hakan Bilen.
\newblock Synthesizing {{Informative Training Samples}} with {{GAN}}, April
  2022.

\bibitem[Zhong et~al.(2020)Zhong, Zheng, Kang, Li, and Yang]{zhong_random_2020}
Zhun Zhong, Liang Zheng, Guoliang Kang, Shaozi Li, and Yi~Yang.
\newblock Random {{Erasing Data Augmentation}}.
\newblock \emph{Proceedings of the AAAI Conference on Artificial Intelligence},
  34\penalty0 (07):\penalty0 13001--13008, April 2020.
\newblock ISSN 2374-3468.
\newblock \doi{10.1609/aaai.v34i07.7000}.

\bibitem[Zhu et~al.(2016)Zhu, Kr{\"a}henb{\"u}hl, Shechtman, and
  Efros]{zhu_generative_2016}
Jun-Yan Zhu, Philipp Kr{\"a}henb{\"u}hl, Eli Shechtman, and Alexei~A. Efros.
\newblock Generative {{Visual Manipulation}} on the {{Natural Image Manifold}}.
\newblock In Bastian Leibe, Jiri Matas, Nicu Sebe, and Max Welling (eds.),
  \emph{Computer {{Vision}} \textendash{} {{ECCV}} 2016}, Lecture {{Notes}} in
  {{Computer Science}}, pp.\  597--613, {Cham}, 2016. {Springer International
  Publishing}.
\newblock ISBN 978-3-319-46454-1.
\newblock \doi{10.1007/978-3-319-46454-1_36}.

\end{thebibliography}
\bibliographystyle{tmlr}

\appendix
\section{Implementation and Practical Tips}
\label{suppl:details}

\subsection{Data augmentation and optimization hyper-parameters}
\paragraph{Handcrafted data augmentations.} We used the VISSL library~\citep{goyal2021vissl} which relies on Torchvision transformations\footnote{\url{https://pytorch.org/vision/stable/transforms.html}}, and considered the following data augmentation strategies:
\begin{itemize}
    \item \underline{Hflip}: Random horizontal flipping applied with 50\% probability.
    \item \underline{RRCrop}: Random resized cropping. The location of the crop is sampled uniformly based on the sampled crop size, which is in the range (0.08, 1). When adopted in our experiments, we apply it to all images in a batch.
    \item \underline{RAug} = RandAugment + Color Jittering (CJ) + Random erasing (RE): When enabled,  RandAugment~\citep{cubuk_randaugment_2020-2} is applied to all images with magnitude $9.0\pm0.5$ and increasing distortion severity for higher magnitude values. At each iteration RandAugment randomly chooses two types of distortions. CJ distorts brightness, contrast, saturation, and hue, each with probability 0.4. RE is applied with probability 0.25, erasing a rectangle of size sampled from (0.02, 0.33).
    \item \underline{CutMixUp}. CutMix~\citep{yun_cutmix_2019} and MixUp~\citep{zhang_mixup_2017} are never applied simultaneously; there is a 0.5 probability of choosing one or the other at each iteration. Note that Mixup is applied 80\% of the time when selected. As previously mentioned, we adopt label smoothing of 0.1 to ease convergence when using CutMixUp.  
\end{itemize}

\paragraph{Optimization hyper-parameters.} In Table~\ref{suppl:tab:in1k_training}, we list the optimization hyper-parameters explored for supervised training on ImageNet. Note that for all supervised experiments, we optimized the multi-class cross-entropy loss.

\begin{table}[ht]
    \caption{Training hyper-parameters for supervised training on ImageNet.}
    \label{suppl:tab:in1k_training}
    \resizebox{\linewidth}{!}{%
    \begin{tabular}{@{}lcccccc@{}}
    \toprule
    Model & Optimizer & Epochs & Learning rate (LR) & LR scheduler & LR scaling & Weight decay \\
    \midrule
    ResNet-50 & Mom. SGD%
    & 105 & \{(5, 2, 1)e-1, (5, 1)e-2, 5e-3\} & Step(30,60,90,100) & Lin-256 & \{(1, 5)e-5, (1, 5)e-4, 1e-3\} \\
     ResNet(-101, -152, 50W2) & Mom. SGD%
    & 105 & 1e-1 & Step(30,60,90,100) & Lin-256 & 1e-4 \\
    DeiT-B & AdamW%
   & 100/300 & 1/5e-4 & Lin + Cosine & Lin-512 & 1e-1/5e-2 \\
    \bottomrule
    \end{tabular}
    }
\end{table}

\subsection{Implementation details}

\paragraph{Fixing the number of IC-GAN-augmented datapoints.} Variable batch size can cause unexpected breaks in GPU-accelerated computations, mostly due to GPU memory pre-allocation. To avoid this phenomenon, we fix the number of \icgan-augmented images in a batch to be \texttt{ceil(batch\_size * p\_G)}.

\paragraph{Computational overhead of \ours.} Adding \ours to the training recipe requires some additional space and time for the \icgan generation. For instance, in terms of space, $\sim$11GB of a single GPU memory are required to generate a batch of 64 images at $256\times 256$ resolution. In terms of time, we noticed that training ResNets with \ours and $p_G=1.0$ doubles the training time, whereas for $p_G=0.5$ the time requirement increases by roughly 50\%. However, we did not take advantage of half-precision computations nor of any other inference-only trick like \texttt{jit} scripting in PyTorch. We hypothesize that exploring such optimizations might significantly reduce the computational overhead of \ours.

\paragraph{Pre-computing dataset embeddings.} The \icgan generation step requires feature representation of the conditioning images. In order to reduce the computation needed during the training, we compute the embeddings of the entire training dataset in advance, and store them in an array which is loaded into memory at the beginning of the training. 

\paragraph{Hardware used for experiments.} For most of the experiments, we performed distributed training using cluster nodes with 8 Nvidia V100 GPUs with 32GB memory. We changed the number of nodes based on the training model and the desired batch size -- e.g., 1 node for ResNets, 4 for DeiT-B, and 8 for SwAV.

\section{Additional Results}
\label{suppl:results}

\subsection{DeiT-B per-class analysis}

\begin{figure}[ht]
\centering
\begin{subfigure}[b]{.43\textwidth}
    \centering
    \includegraphics[width=\textwidth]{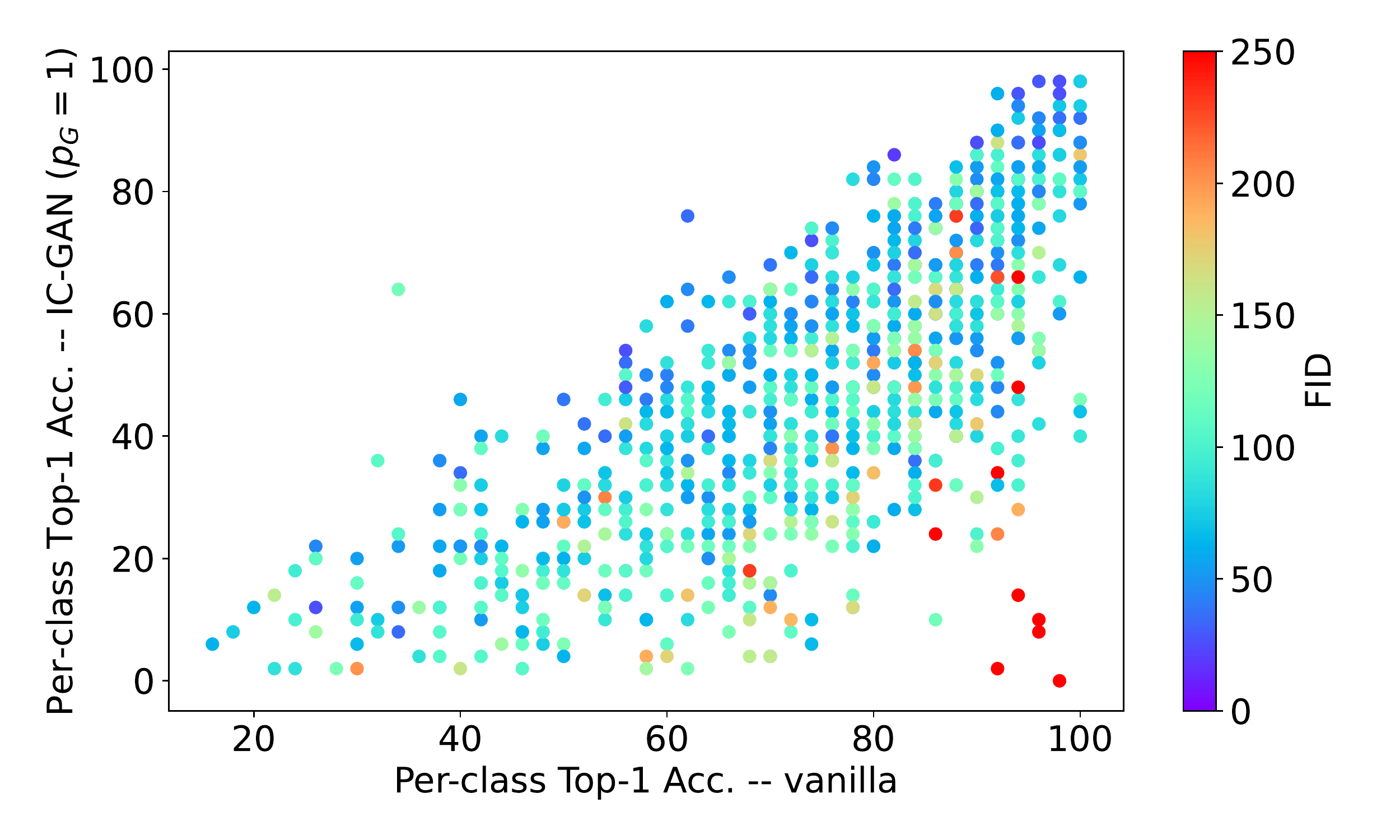}
    \caption{FID -- DeiT-B, \icgan}
\end{subfigure}
\begin{subfigure}[b]{.43\textwidth}
    \centering
    \includegraphics[width=\textwidth]{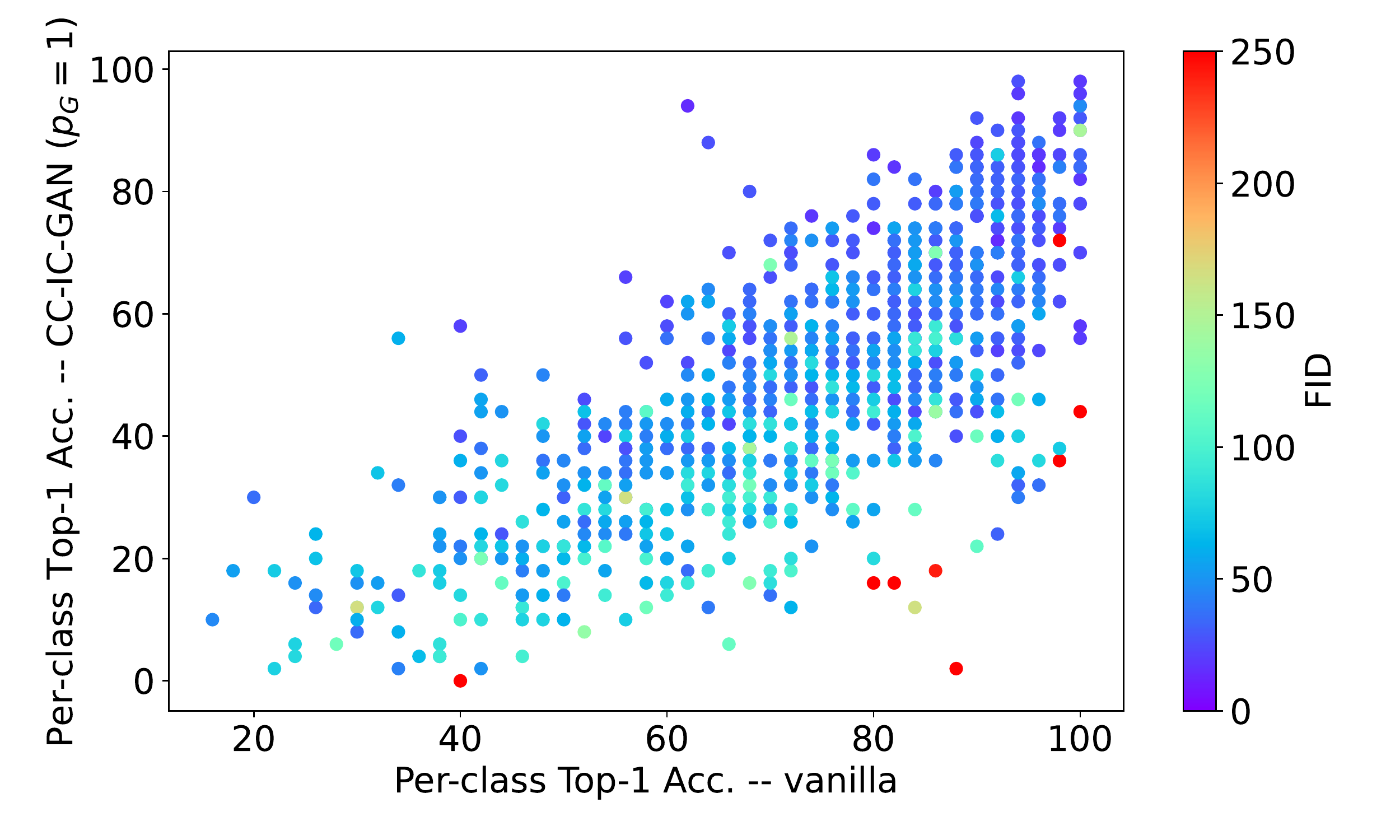}
    \caption{FID -- DeiT-B, \ccicgan}
\end{subfigure}
\begin{subfigure}[b]{.43\textwidth}
    \centering
    \includegraphics[width=\textwidth]{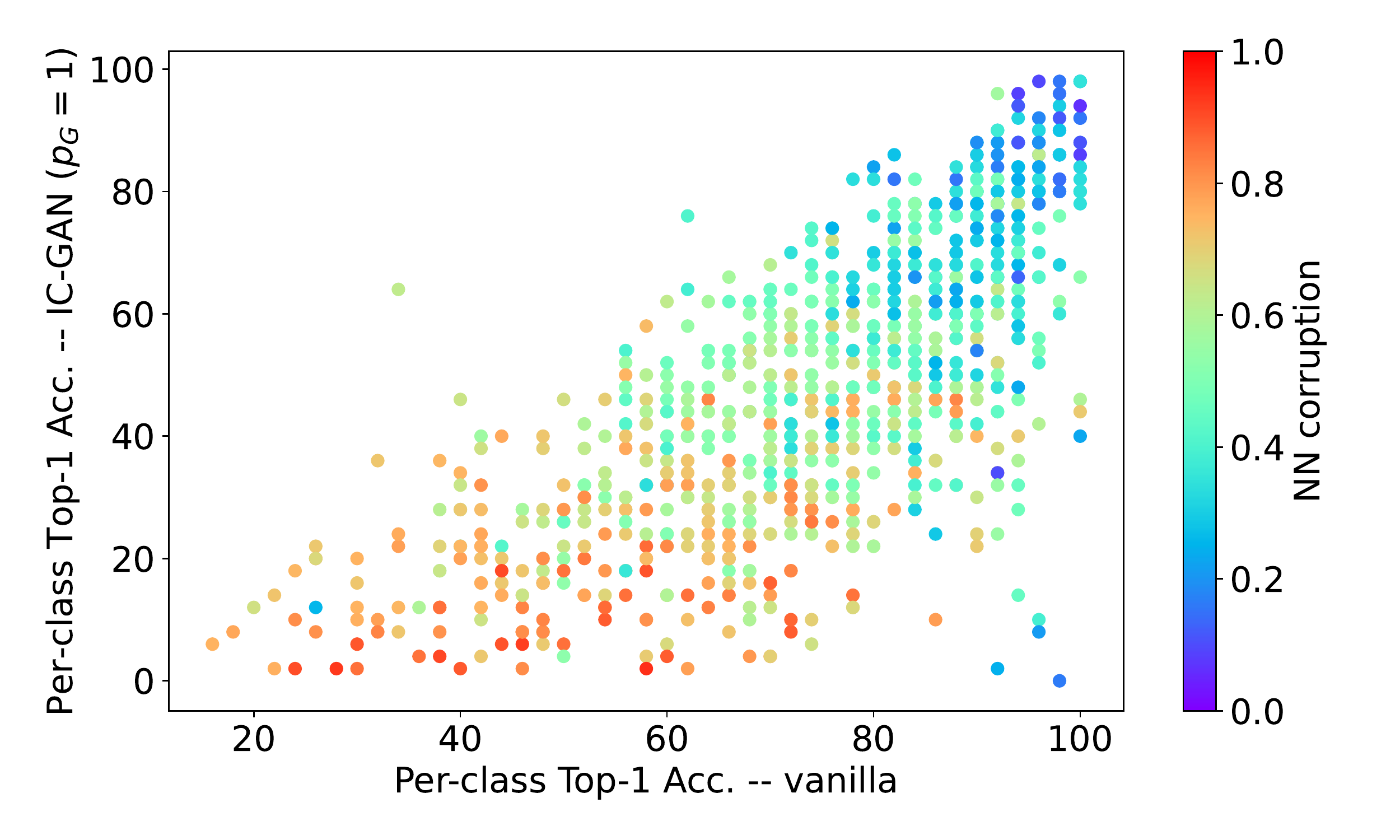}
    \caption{NN corruption -- DeiT-B, \icgan}
\end{subfigure}
\begin{subfigure}[b]{.43\textwidth}
    \centering
    \includegraphics[width=\textwidth]{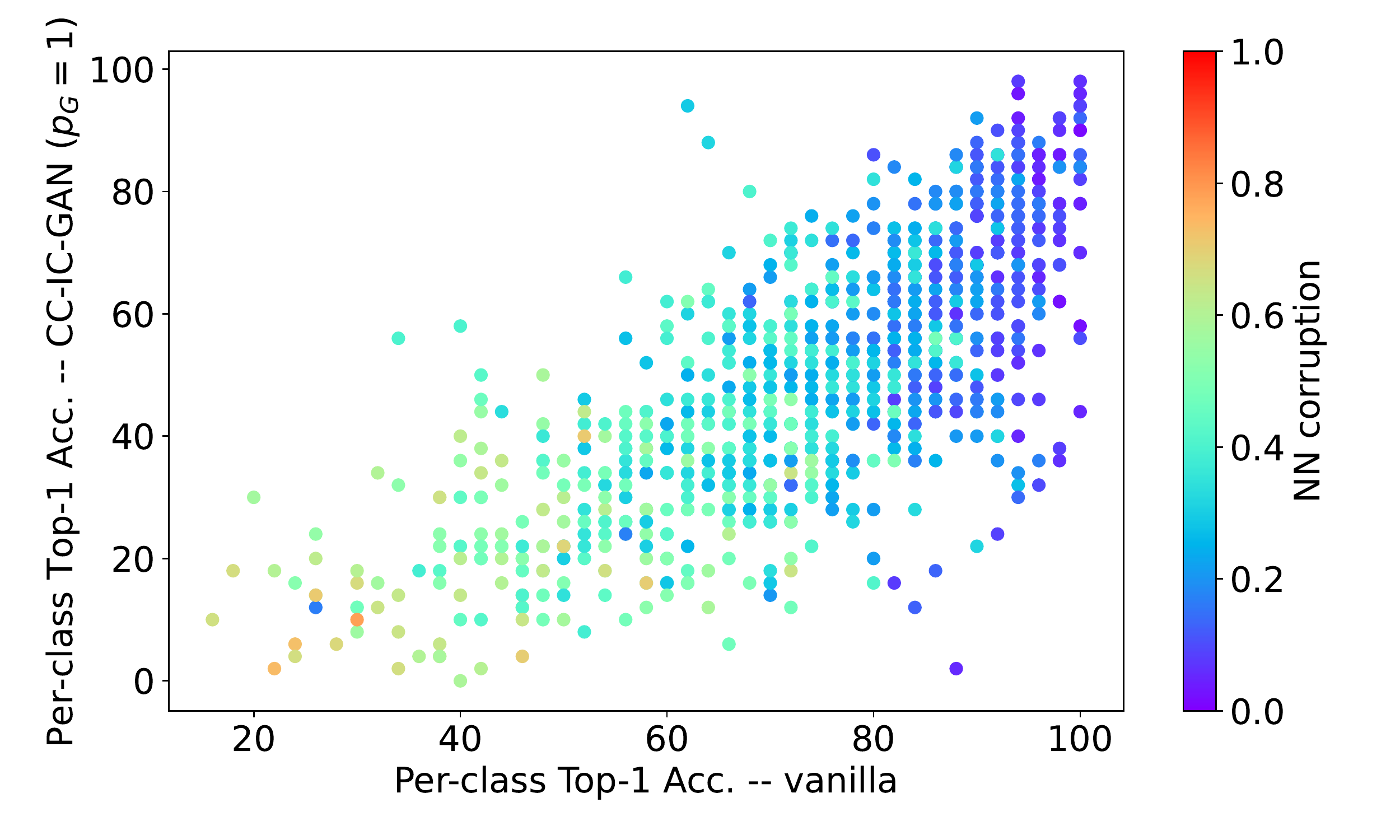}
    \caption{NN corruption -- DeiT-B, \ccicgan}
\end{subfigure}
\caption{
Impact of \allicgan's generation quality on per-class performance. (a-b) Per-class FID as a function of per-class top-1 accuracy of the vanilla and \ours models. We observe that higher quality \allicgan generations tend to result in improved performances. (c-d) Per-class NN corruption as a function of per-class top-1 accuracy of the vanilla and \ours models. We observe that less corrupted classes tend to result in improved performances. ImageNet validation results are shown for the DeiT-B model trained with horizontal flips, random crops, and RandAugment. We limited the FID colormap interval to 250 to aid interpretability, while we observed FID values up to 500 for certain classes.}
\label{suppl:fig:per_class_deit}
\end{figure}

Figure~\ref{suppl:fig:per_class_deit} assesses the impact of \allicgan's generation quality on the per class performance of the DeiT-B model. The exclusive use of generated samples to train DeiT-B leads to rather a low top-1 accuracy of $\sim$48\% and $\sim$51\%, when using \icgan and \ccicgan respectively.\looseness-1

Following section~\ref{sssec:class}, Figures~\ref{suppl:fig:per_class_deit} (a--b) show the per-class FID of \allicgan as a function of per-class top-1 accuracy of the vanilla baseline and the \ours models. We observe similar trends as for the ResNet-152 models, -- i.e., \ours tends to exhibit higher accuracy for classes with lower (better) FID values, and lower accuracy for classes with higher FID values, suggesting that using image generations of poorly modeled classes hurts the performance of DeiT-B. Figures~\ref{suppl:fig:per_class_deit} (c--d) highlight that the low accuracies of the model trained with generated data can be partially explained by the NN corruption.

\subsection{Avoiding \ours on low quality classes}
In this analysis, we try to exploit the observed correlation between per-class accuracy and per-class FID, i.e., samples quality (see Section~\ref{sssec:class}), by restricting the use of \ours only to classes that have an acceptable quality. We set an FID threshold of 150, under which we consider a class to have acceptable quality as the visual inspection of classes with FID >= 150 reveals either very poor image quality or mode-collapse (as shown in Figure~\ref{fig:mode_collapse}); this threshold value is distant around 1.5$\sigma$ and 3$\sigma$ from the average per-class FID computed on \icgan and \ccicgan samples, respectively. For this experiment, we train ResNet-152 with an augmentation recipe composed by HFlip and RRCrop applied to all classes, and \ours applied to FID-filtered classes. We report the results in Table~\ref{suppl:tab:fid_filtered_rn152}.

\begin{table}[ht]
\centering
\caption{ImageNet classification accuracy of ResNet-152 when using \ours indistinctly on all classes vs. augmenting only classes with FID < 150. For each column of results we report the mean top-1 accuracy computed over the indicated set of classes.}
\label{suppl:tab:fid_filtered_rn152}
\resizebox{.9\textwidth}{!}{%
\begin{tabular}{@{}lllccc@{}}
\toprule
\multirow{2}{*}{Method} & \multirow{2}{*}{DA base} & \multirow{2}{*}{\ours} & \multicolumn{3}{c}{Top-1 accuracy} \\
 & & & all classes & classes w/ FID < 150 & classes w/ FID >= 150 \\ \midrule
\multirow{2}{*}{ResNet-152} & \multirow{2}{*}{HFlip + RRCrop} & w/ \icgan & 77.71 & 77.44 & 80.44 \\
 & & w/ FID-filtered \icgan & \textbf{77.94} & \textbf{77.56} & \textbf{81.67} \\ \midrule
\multirow{2}{*}{ResNet-152} & \multirow{2}{*}{HFlip + RRCrop} & w/ \ccicgan & 77.96 & 77.92 & 80.29 \\
 & & w/ FID-filtered \ccicgan & 77.96 & 77.92 & \textbf{80.71} \\ \bottomrule
\end{tabular}%
}
\end{table}

Overall, we obtain, on average, a slightly better top-1 accuracy, +0.2\%p, which can be stratified into +1.2\%p considering the classes with FID >= 150 and +0.1\%p on the remaining classes. From these results we can observe that skipping the use of \ours on poorly modeled classes increases the performances on such classes, while not harming the performance on the others.

\section{Additional Visualizations}
\label{suppl:visual}

\begin{figure}[ht]
\centering
\begin{subfigure}[b]{.2\textwidth}
    \centering
    \includegraphics[width=.9\textwidth]{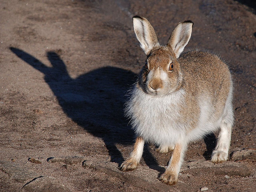}
    \caption{Original}
    \vspace{.13cm}
    \includegraphics[width=.9\textwidth]{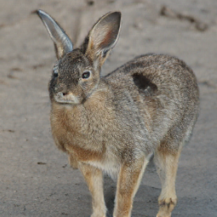}
    \caption{\icgan}
\end{subfigure}
\hfill
\begin{subfigure}[b]{.18\textwidth}
    \centering
    \includegraphics[width=\textwidth]{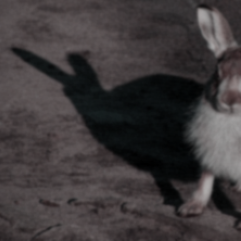}\\
    \includegraphics[width=\textwidth]{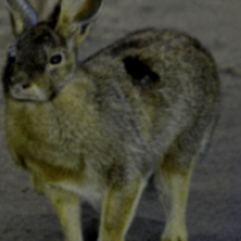}
    \caption{Main crops $224^2$}
\end{subfigure}
\hfill
\begin{subfigure}[b]{.55\textwidth}
    \centering
    \includegraphics[width=.3\textwidth]{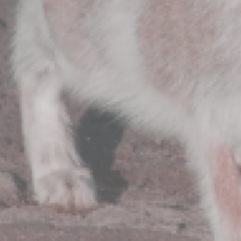}
    \includegraphics[width=.3\textwidth]{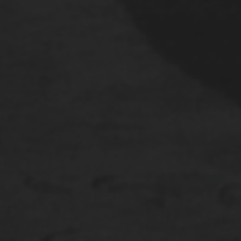}
    \includegraphics[width=.3\textwidth]{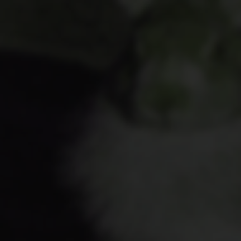}\\
    \includegraphics[width=.3\textwidth]{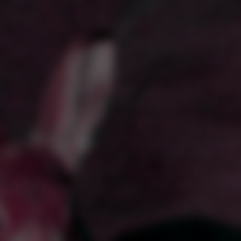}
    \includegraphics[width=.3\textwidth]{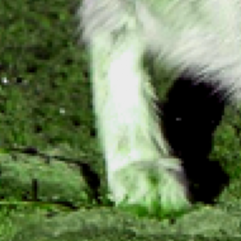}
    \includegraphics[width=.3\textwidth]{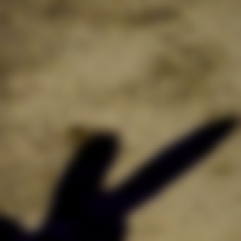}
    \caption{Small crops $96^2$}
\end{subfigure}
    \caption{Example of \ours combined with multi-crop~\citep{caron_unsupervised_2020} augmentation with 2 main crops and 6 small crops: (a) depicts the original image, which is used to condition the \icgan generation process; (b) displays an \icgan generation; (c) shows the main crops of both images; and (d) presents the small crops obtained from the original image.}
    \label{suppl:fig:seqC_mc_batch}
\end{figure}

Figure~\ref{suppl:fig:seqC_mc_batch} displays images resulting from the combination of \ours with the multi-crop augmentation used in the SwAV model~\cite{caron_unsupervised_2020}. As shown in the figure, these augmentations result in significant variations of the original image, with small crop images notably differing from the \icgan generations.\looseness-1  

\begin{figure}[ht]
\centering
\begin{subfigure}[b]{\textwidth}
     \raggedright
     \small{``warthog''}
 \end{subfigure}
 \begin{subfigure}[b]{.1\textwidth}
    \centering
    \includegraphics[width=\textwidth]{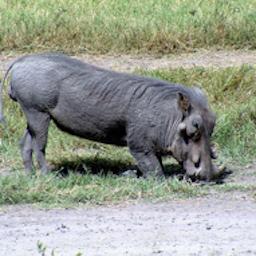}
 \end{subfigure}
\hfill
\begin{subfigure}[b]{.44\textwidth}
    \centering
    \includegraphics[width=.22\textwidth]{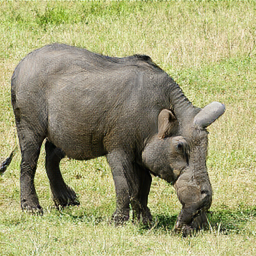}
    \includegraphics[width=.22\textwidth]{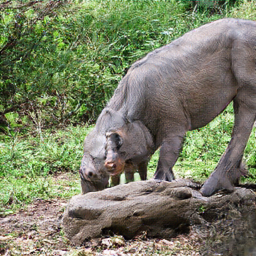}
    \includegraphics[width=.22\textwidth]{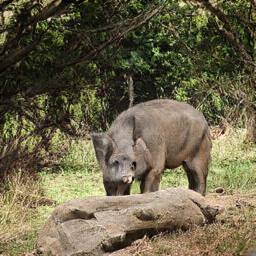}
    \includegraphics[width=.22\textwidth]{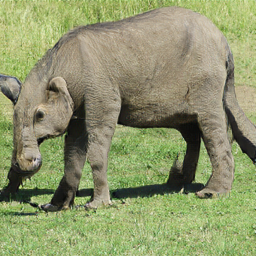}
\end{subfigure}
\hfill
\begin{subfigure}[b]{.44\textwidth}
    \centering
    \includegraphics[width=.22\textwidth]{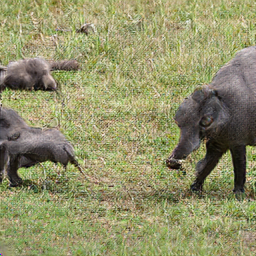}
    \includegraphics[width=.22\textwidth]{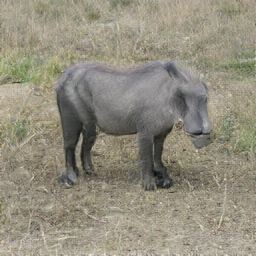}
    \includegraphics[width=.22\textwidth]{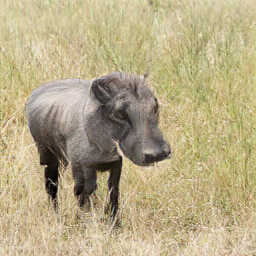}
    \includegraphics[width=.22\textwidth]{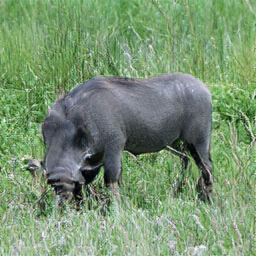}
\end{subfigure}
\\
\begin{subfigure}[b]{\textwidth}
     \raggedright
     \small{``black swan''}
 \end{subfigure}
 \begin{subfigure}[b]{.1\textwidth}
    \centering
    \includegraphics[width=\textwidth]{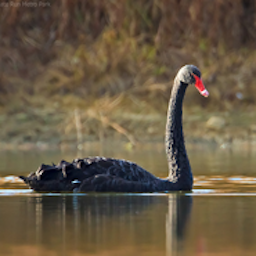}
 \end{subfigure}
\hfill
\begin{subfigure}[b]{.44\textwidth}
    \centering
    \includegraphics[width=.22\textwidth]{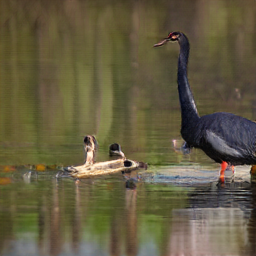}
    \includegraphics[width=.22\textwidth]{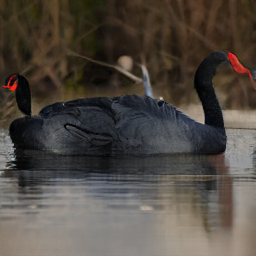}
    \includegraphics[width=.22\textwidth]{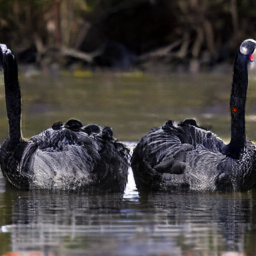}
    \includegraphics[width=.22\textwidth]{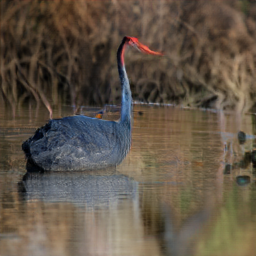}
\end{subfigure}
\hfill
\begin{subfigure}[b]{.44\textwidth}
    \centering
    \includegraphics[width=.22\textwidth]{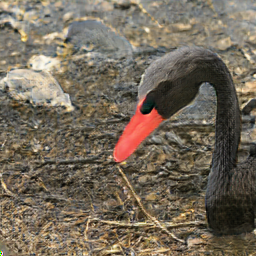}
    \includegraphics[width=.22\textwidth]{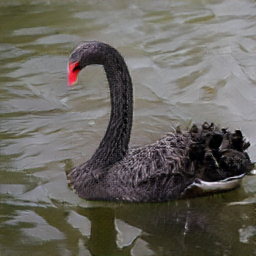}
    \includegraphics[width=.22\textwidth]{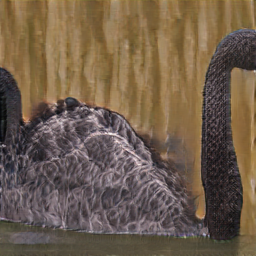}
    \includegraphics[width=.22\textwidth]{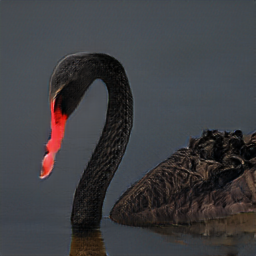}
\end{subfigure}
\\
\begin{subfigure}[b]{\textwidth}
     \raggedright
     \small{``siberian husky''}
 \end{subfigure}
 \begin{subfigure}[b]{.1\textwidth}
    \centering
    \includegraphics[width=\textwidth]{figures/icgan_examples/img_25_class_Siberian_husky}
 \end{subfigure}
\hfill
\begin{subfigure}[b]{.44\textwidth}
    \centering
    \includegraphics[width=.22\textwidth]{figures/icgan_examples/img_26_class_Siberian_husky}
    \includegraphics[width=.22\textwidth]{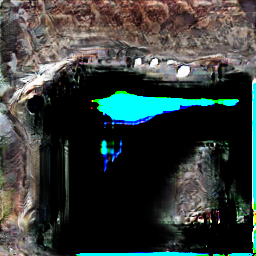}
    \includegraphics[width=.22\textwidth]{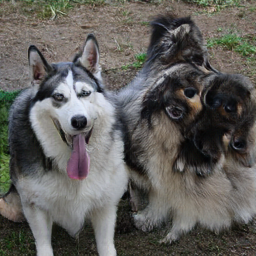}
    \includegraphics[width=.22\textwidth]{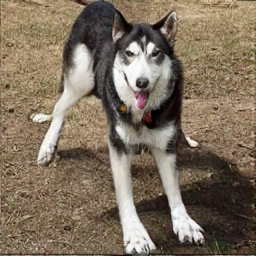}
\end{subfigure}
\hfill
\begin{subfigure}[b]{.44\textwidth}
    \centering
    \includegraphics[width=.22\textwidth]{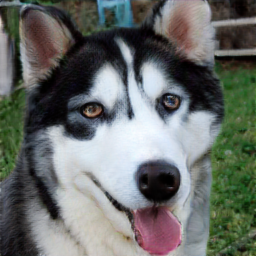}
    \includegraphics[width=.22\textwidth]{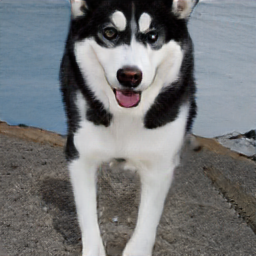}
    \includegraphics[width=.22\textwidth]{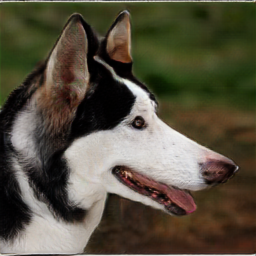}
    \includegraphics[width=.22\textwidth]{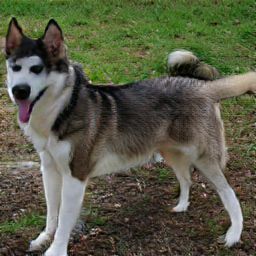}
\end{subfigure}
\\
\begin{subfigure}[b]{\textwidth}
     \raggedright
     \small{``tiger beetle''}
 \end{subfigure}
 \begin{subfigure}[b]{.1\textwidth}
    \centering
    \includegraphics[width=\textwidth]{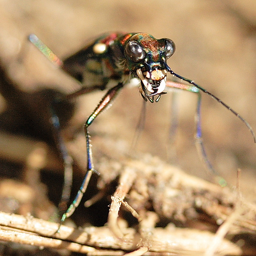}
 \end{subfigure}
\hfill
\begin{subfigure}[b]{.44\textwidth}
    \centering
    \includegraphics[width=.22\textwidth]{figures/icgan_examples/img_16_class_tiger_beetle}
    \includegraphics[width=.22\textwidth]{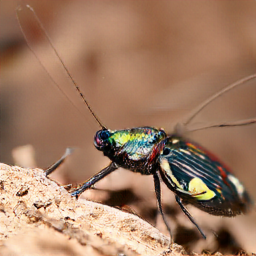}
    \includegraphics[width=.22\textwidth]{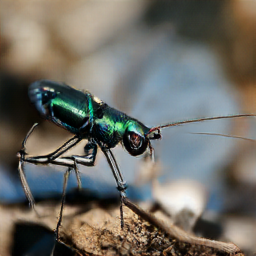}
    \includegraphics[width=.22\textwidth]{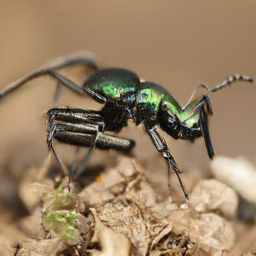}
\end{subfigure}
\hfill
\begin{subfigure}[b]{.44\textwidth}
    \centering
    \includegraphics[width=.22\textwidth]{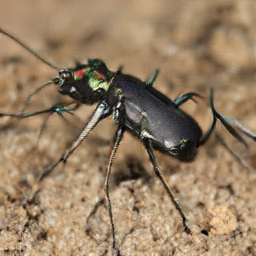}
    \includegraphics[width=.22\textwidth]{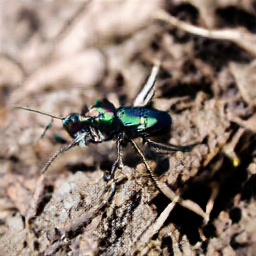}
    \includegraphics[width=.22\textwidth]{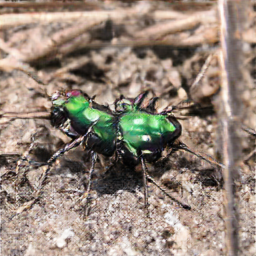}
    \includegraphics[width=.22\textwidth]{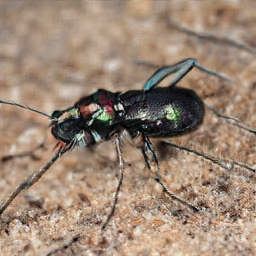}
\end{subfigure}
\\
\begin{subfigure}[b]{\textwidth}
     \raggedright
     \small{``beer glass''}
 \end{subfigure}
 \begin{subfigure}[b]{.1\textwidth}
    \centering
    \includegraphics[width=\textwidth]{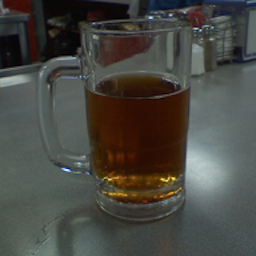}
 \end{subfigure}
\hfill
\begin{subfigure}[b]{.44\textwidth}
    \centering
    \includegraphics[width=.22\textwidth]{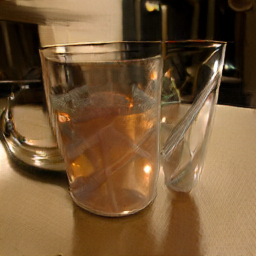}
    \includegraphics[width=.22\textwidth]{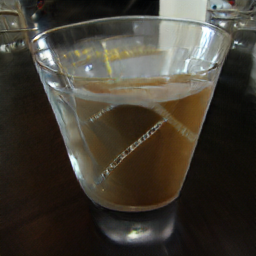}
    \includegraphics[width=.22\textwidth]{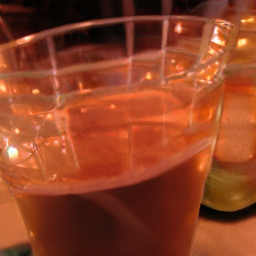}
    \includegraphics[width=.22\textwidth]{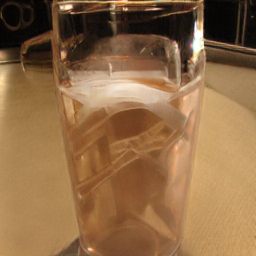}
\end{subfigure}
\hfill
\begin{subfigure}[b]{.44\textwidth}
    \centering
    \includegraphics[width=.22\textwidth]{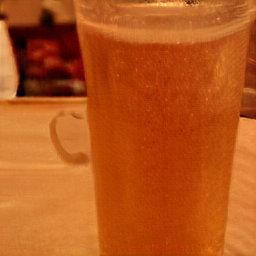}
    \includegraphics[width=.22\textwidth]{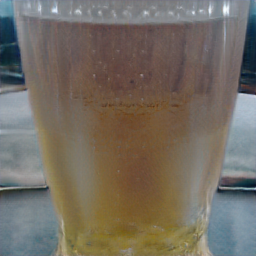}
    \includegraphics[width=.22\textwidth]{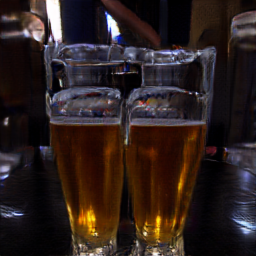}
    \includegraphics[width=.22\textwidth]{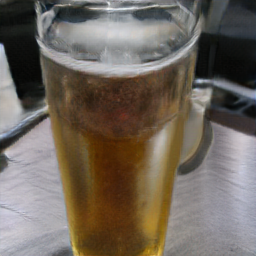}
\end{subfigure}
\\
\begin{subfigure}[b]{\textwidth}
     \raggedright
     \small{``cliff dwelling''}
 \end{subfigure}
 \begin{subfigure}[b]{.1\textwidth}
    \centering
    \includegraphics[width=\textwidth]{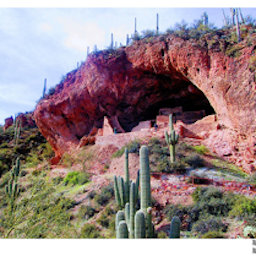}
 \end{subfigure}
\hfill
\begin{subfigure}[b]{.44\textwidth}
    \centering
    \includegraphics[width=.22\textwidth]{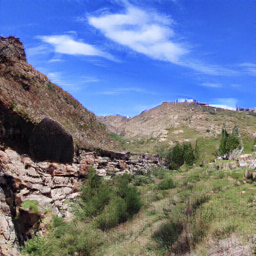}
    \includegraphics[width=.22\textwidth]{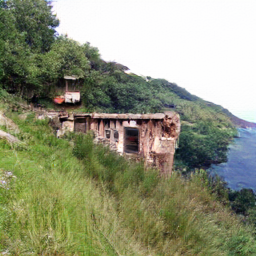}
    \includegraphics[width=.22\textwidth]{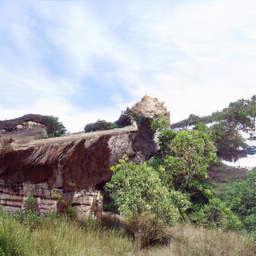}
    \includegraphics[width=.22\textwidth]{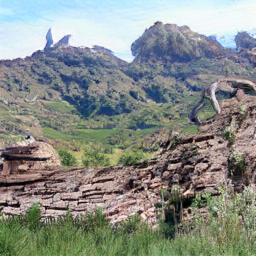}
\end{subfigure}
\hfill
\begin{subfigure}[b]{.44\textwidth}
    \centering
    \includegraphics[width=.22\textwidth]{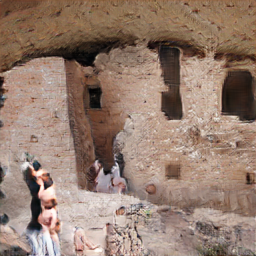}
    \includegraphics[width=.22\textwidth]{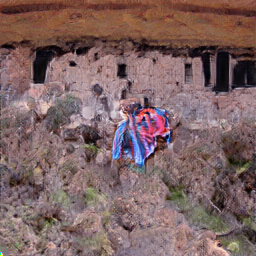}
    \includegraphics[width=.22\textwidth]{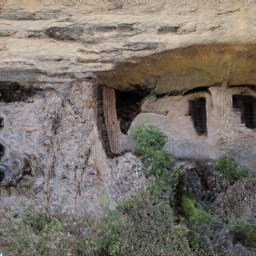}
    \includegraphics[width=.22\textwidth]{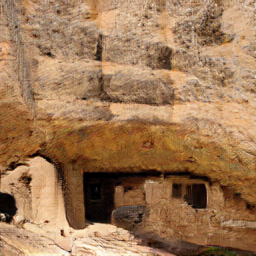}
\end{subfigure}
\\
\begin{subfigure}[b]{\textwidth}
     \raggedright
     \small{``hook'}
 \end{subfigure}
 \begin{subfigure}[b]{.1\textwidth}
    \centering
    \includegraphics[width=\textwidth]{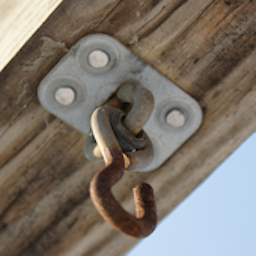}
 \end{subfigure}
\hfill
\begin{subfigure}[b]{.44\textwidth}
    \centering
    \includegraphics[width=.22\textwidth]{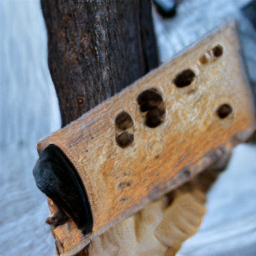}
    \includegraphics[width=.22\textwidth]{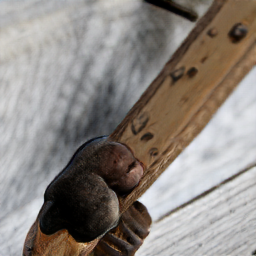}
    \includegraphics[width=.22\textwidth]{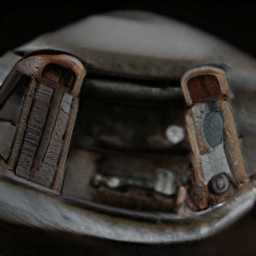}
    \includegraphics[width=.22\textwidth]{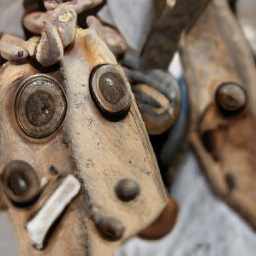}
\end{subfigure}
\hfill
\begin{subfigure}[b]{.44\textwidth}
    \centering
    \includegraphics[width=.22\textwidth]{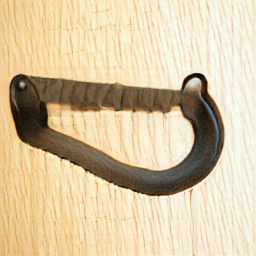}
    \includegraphics[width=.22\textwidth]{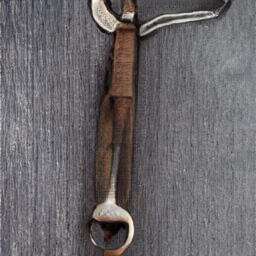}
    \includegraphics[width=.22\textwidth]{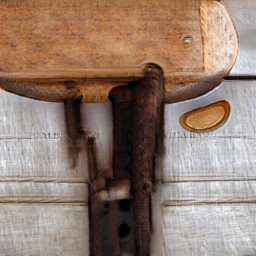}
    \includegraphics[width=.22\textwidth]{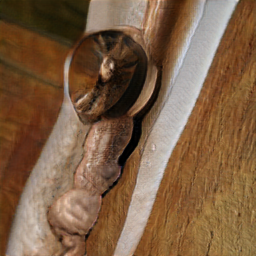}
\end{subfigure}
\\
\begin{subfigure}[b]{\textwidth}
     \raggedright
     \small{``slot''}
 \end{subfigure}
 \begin{subfigure}[b]{.1\textwidth}
    \centering
    \includegraphics[width=\textwidth]{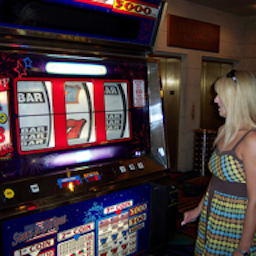}
 \end{subfigure}
\hfill
\begin{subfigure}[b]{.44\textwidth}
    \centering
    \includegraphics[width=.22\textwidth]{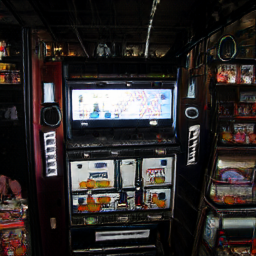}
    \includegraphics[width=.22\textwidth]{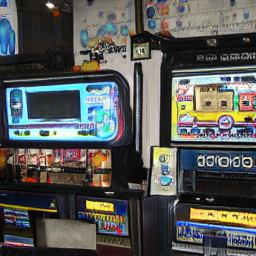}
    \includegraphics[width=.22\textwidth]{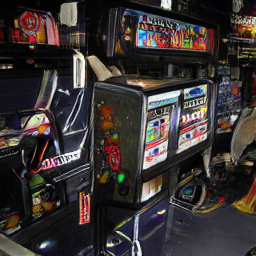}
    \includegraphics[width=.22\textwidth]{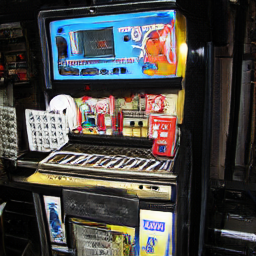}
\end{subfigure}
\hfill
\begin{subfigure}[b]{.44\textwidth}
    \centering
    \includegraphics[width=.22\textwidth]{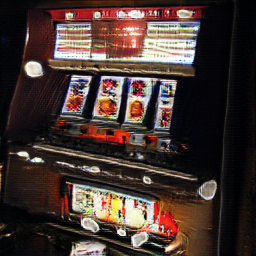}
    \includegraphics[width=.22\textwidth]{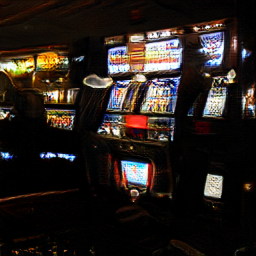}
    \includegraphics[width=.22\textwidth]{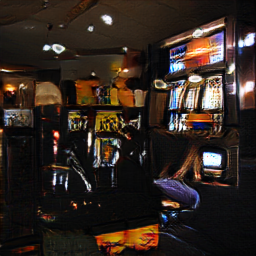}
    \includegraphics[width=.22\textwidth]{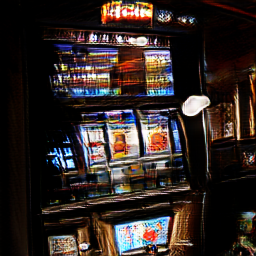}
\end{subfigure}
\\
\begin{subfigure}[b]{\textwidth}
     \raggedright
     \small{``water tower''}
 \end{subfigure}
 \begin{subfigure}[b]{.1\textwidth}
    \centering
    \includegraphics[width=\textwidth]{figures/icgan_examples/img_45_class_water_tower_crop}
 \end{subfigure}
\hfill
\begin{subfigure}[b]{.44\textwidth}
    \centering
    \includegraphics[width=.22\textwidth]{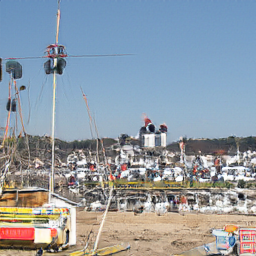}
    \includegraphics[width=.22\textwidth]{figures/icgan_examples/img_47_class_water_tower}
    \includegraphics[width=.22\textwidth]{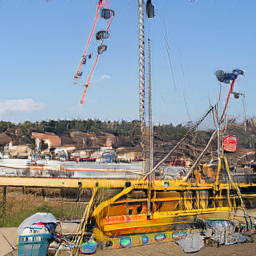}
    \includegraphics[width=.22\textwidth]{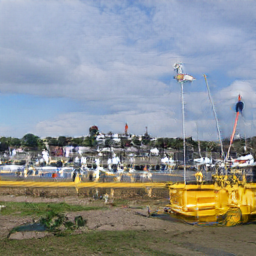}
\end{subfigure}
\hfill
\begin{subfigure}[b]{.44\textwidth}
    \centering
    \includegraphics[width=.22\textwidth]{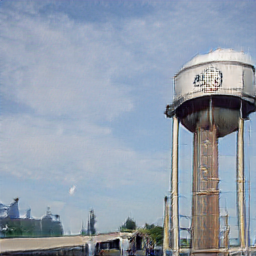}
    \includegraphics[width=.22\textwidth]{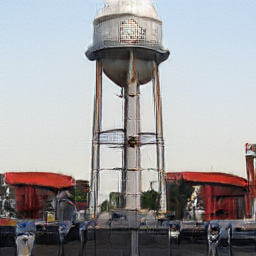}
    \includegraphics[width=.22\textwidth]{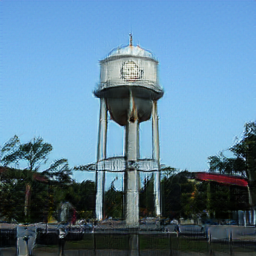}
    \includegraphics[width=.22\textwidth]{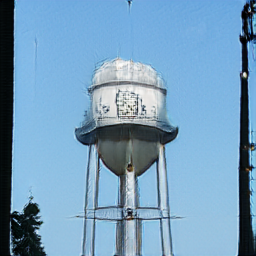}
\end{subfigure}
\\
\vspace{.3cm}
\begin{subfigure}[b]{.1\textwidth}
     \centering
     \caption*{Condit.}
\end{subfigure}
\hfill
\begin{subfigure}[b]{.44\textwidth}
     \centering
     \caption{\icgan samples}
 \end{subfigure}
 \hfill
\begin{subfigure}[b]{.44\textwidth}
     \centering
     \caption{\ccicgan samples}
 \end{subfigure}
    \caption{Visual examples of \allicgan generations. Each row shows, from left to right, the conditioning image -- i.e., central crop of ImageNet image --, followed by \icgan (a) and \ccicgan (b) generated samples. Generetad samples were obtained using the depicted image conditioning and different noise vectors.}
    \label{suppl:fig:icgan_ex}
\end{figure}

Figure~\ref{suppl:fig:icgan_ex} displays \allicgan generations. Note that \icgan and \ccicgan generations (a--b) tend to show slightly different viewpoints and instances of the object present in the conditioning image (left-most column).

\end{document}